\documentclass[11pt]{article}
\usepackage[T1]{fontenc}
\usepackage[utf8]{inputenc}
\DeclareUnicodeCharacter{0304}{}
\usepackage{lmodern}
\usepackage{amsmath}
\usepackage{amssymb}
\usepackage{amsthm}
\usepackage{mathtools}
\usepackage{bm}
\usepackage{booktabs}
\usepackage{array}
\usepackage{tabularx}
\usepackage{enumitem}
\usepackage{graphicx}
\usepackage{tikz}
\usetikzlibrary{arrows.meta,positioning}
\DeclareRobustCommand{\pfull}{\tikz[baseline=-0.6ex]{\fill (0,0) circle (0.75ex);}}
\DeclareRobustCommand{\phalf}{\tikz[baseline=-0.6ex]{\draw (0,0) circle (0.75ex); \fill (0,0) -- (90:0.75ex) arc (90:270:0.75ex) -- cycle;}}
\DeclareRobustCommand{\pnone}{\tikz[baseline=-0.6ex]{\draw (0,0) circle (0.75ex);}}
\usepackage{authblk}
\usepackage{geometry}
\usepackage{pdflscape}
\usepackage{afterpage}
\usepackage{xcolor}
\usepackage{soul}
\sethlcolor{yellow}

\soulregister\ref7
\soulregister\eqref7
\soulregister\cite7

\theoremstyle{definition}
\newtheorem{defn}{Definition}[section]

\providecommand{\tabularnewline}{\\}

\usepackage[colorlinks=true, linkcolor=blue!45!black, citecolor=blue!45!black, urlcolor=blue!45!black]{hyperref}
\hypersetup{pdftitle={When Intelligence Becomes Agency}, pdfauthor={João Dias Ferreira and Bhaskar Tripathi}}
\usepackage[style=numeric,sorting=none]{biblatex}
\AtBeginBibliography{\emergencystretch=1.5em}
\title{When Intelligence Becomes Agency:\\
\large A Theory of Governed, Proactive Agency for Symbiotic AI Systems}
\author{Jo\~{a}o Dias Ferreira\thanks{\texttt{joaopsdf@wyrdeai.com}}}
\affil{Wyrde AI}
\date{}

\begin{document}
\maketitle

\begin{abstract}
Persistent AI assistants are intended to extend human attention, memory, and coordination across changing digital and physical environments. To be truly useful they must do more than just act when asked. They must decide on their own whether a situation warrants behavior at all, when it does and in what mode, whether to act, ask, monitor, defer or deliberately refrain. We call this the activation problem. Research on commitment, appraisal, mixed-initiative interaction and delegation each illuminates part of it, but none ties situated activation to continuing authorization and accountability. This paper develops a conceptual and formal framework for governed proactive agency, organizing behavior across time through perception, intent, affective-conative appraisal, constraint, and feedback. It distinguishes autonomous and delegated agency and defines symbiotic agency as delegation under a standing, revocable mandate, with continuing coupling to the principal’s situation, calibrated inference of their condition, and bounded personalization. The distinctive contribution is an integrated account linking activation decisions to authorized perception, behavior selection, authority containment, traceable restraint, and constrained adaptation, with behavioral episodes as the unit of analysis. Through an agency classification method, an evaluation framework, proposed benchmark scenarios, and a reference architecture, the account provides a basis for specifying and assessing whether assistance is warranted, timely, authorized, and answerable beyond task completion alone. It is intended to guide the development and evaluation of always-present personal assistants and embodied support systems that augment human capabilities while preserving the principal’s authority and judgment.
\end{abstract}

\medskip
\noindent \textbf{Keywords:} AI agents, proactive agency, symbiotic AI, delegated agency, human augmentation, activation problem, AI governance, human--AI interaction.

\section{Introduction}\label{sec:introduction}

Recent developments in AI models and in the harnesses built around them have focused on enabling AI systems to operate over long-horizon tasks. AI agents can now read instructions, call tools, revise plans, evaluate their progress against goals and continue acting until a delegated task is deemed to be complete. Such capabilities are manifestations of intelligence at work. Nevertheless, most of these systems still instantiate a reactive agency pattern more than a proactive one. Their behavior is activated from outside, typically by a concrete user request, event, command, schedule, or tool result, after which they can execute sophisticated multi-step behavior. Proactive agency, by contrast, evaluates ongoing observations against standing commitments to determine whether a new behavioral episode is warranted, without requiring a contemporaneous instruction. It requires the capacity for an agent to remain situated over time, monitoring its surroundings and its own internal condition in order to determine what matters and what, if anything, should follow from it.

In this sense, current agentic capabilities mainly answer \emph{what} to do and \emph{how} to do it once a task has been given. The question that comes before it is \emph{whether} anything should be done now, \emph{when} it should be done and \emph{in what mode} it should occur, such as asking, reminding, monitoring, deferring, escalating, or deliberately refraining, given the standing goals and commitments the system holds. Current capabilities do not, by themselves, answer this question, yet it is the one that governs agency in practice. We name it the \emph{Activation Problem}. It is one of the key problems of agency and it is the problem this paper formalizes.

Consider a personal agent that stays with its user throughout the day and watches over their work, their messages and their health. In one short interval it picks up three things, a critical warning from an important system at work, a burst of family messages about a school event and a health alert that the user has not moved for hours. Each may matter on its own, but not equally. In this instance the work related critical warning takes priority, the health alert can wait for a pause and the school messages can be summarized later. Had the messages said that one of the children was injured at school and was being taken to the hospital, that would have taken the priority and the warning would have waited or directed to a colleague that could take care of it. Nothing in the signals themselves fixes this order. If the agent interrupts three times, or leads with the wrong one, the failure is not one of perception, reasoning, or tool use. It has noticed everything. It has failed to judge what matters, whether anything should be raised now, in what mode, and whether monitoring, deferral, escalation, or deliberate silence would better serve what the user delegated to it.

This example illustrates why agency cannot be reduced to the capacity to perform correct actions in isolation. Agency begins when action is organized around an aim, when behavior is selected, sequenced, constrained and revised in relation to something the system is trying to bring about, maintain, avoid, or respect. We do not observe agency directly. We infer it from structured patterns of behavior across time. What the system notices, when it responds, which goals it prioritizes, how it handles uncertainty, whether it adapts after failure and whether it remains within boundaries.

A common mistake in discussions of AI agents is therefore to treat agency as if it were the sum of a set of components. In such views, a system becomes agentic if it has a model, memory, tools, skills, planning, goals and perhaps a feedback loop. But this additive view is insufficient. A system may contain all of these elements and still fail to behave agentically in the stronger sense. A language model may reason fluently without maintaining durable commitments. A planner may generate action sequences without understanding their significance. A memory system may store facts without knowing when they matter. A tool-using workflow may execute steps efficiently while remaining blind to context, user state, risk, or appropriateness. A reactive controller may respond to stimuli without forming any continuing orientation toward what should be brought about, preserved, avoided, or respected.

The relevant question is therefore not \emph{what components does the system have?} but \emph{how are those components organized so that behavior arises when it should, and is directed, situated, constrained and revisable when it does?} Agency is an organizational property and this organization rests on five elements, perception, intent, affective-conative inference, constraint and feedback. Perception supplies what the situation is, intent holds what the system is committed to, affective-conative inference weighs what matters here and now and why, in the light of the system's commitments and the situation of those it serves (Section~\ref{subsec:aci-theory}), constraint sets what may be done and feedback revises all of these in the light of what follows. Intelligence is then what realizes them and behavior is where their organization becomes visible.

Together these five elements also set the minimum requirements for a system to count as an agent, the agenthood minimum. Each must be present, at least in part. If any one is absent, the system is not an agent, however capable it is otherwise (Definition~\ref{def:agenthood}). Beyond the minimum lies the capacity to determine not only how to act, but whether, when and in what mode behavior should occur, including restraint, together with an observable record of what was done. A system that acts only when told can meet the minimum without this capacity, so the two are kept apart. Agency is \emph{complete}, in the sense used in this paper, only when all of these elements are present in full. Between them, agency is a matter of degree, of how fully each of these elements is present, and what moves a system along that range is the capacity the activation problem names.

The activation problem becomes especially important as we move toward always-present personal AI systems that live in symbiosis with us, enhance our capabilities, work on our behalf and extend our capabilities in both physical (through humanoid robots or other types of actuators) and digital environments. This trajectory returns to Licklider's founding vision of man--computer symbiosis, the tight coupling of human and machine into one problem-solving partnership \cite{Licklider1960}. The present work concerns the governed, agentic form of that coupling. In such systems, usefulness depends not only on what the system is capable of doing, but on whether it can regulate when and how its capabilities become behavior. A technically capable system that cannot make this distinction may become not more agentic, but merely more interruptive.

This trajectory makes it useful to introduce three forms that advanced agency can take. In the first, \emph{autonomous agency}, the agent pursues its own goals, whether self-generated or fully transferred to it. It acts as an independent actor and is evaluated against its own objectives, task constraints and external norms. In the second, \emph{delegated agency}, the agent acts on behalf of someone else, named the \emph{principal}, under the bounded authority that the principal has transferred to the delegated agent itself. The third, \emph{symbiotic agency} is the main focus of this work. It is a case of delegated agency in which the agent stays continuously coupled to the principal. The agent perceives alongside the principal, infers what matters to the principal and acts, waits, asks, escalates, or deliberately refrains in order to augment rather than replace the principal´s capabilities and judgment. A self-play game agent is autonomous, a coding agent completing a task for a developer is delegated and an always-present personal assistant acting under a standing mandate is symbiotic.

\begin{figure}[t]
\centering
\resizebox{\textwidth}{!}{%
\begin{tikzpicture}[font=\footnotesize,
  region/.style={draw=black!60, rounded corners=3pt},
  chip/.style={draw=black!50, rounded corners=2pt, fill=white, inner sep=2.5pt, font=\scriptsize\itshape},
  hdr/.style={anchor=north west, align=left, inner sep=0pt},
  txt/.style={anchor=north west, align=left, inner sep=0pt}]
\draw[region, fill=black!2] (0,0) rectangle (15.8,6.6);
\node[anchor=north, font=\footnotesize\itshape] at (7.9,6.48) {agents (Definition~\ref{def:agenthood})};
\draw[region, fill=black!6] (0.25,0.25) rectangle (7.6,5.95);
\node[hdr] at (0.55,5.65) {\textbf{Autonomous agency}\quad $J_{t}=\emptyset$};
\node[txt] at (0.55,5.05) {no one retains authority over its pursuit;\\ goals held in its own right, of any origin,\\ and ownership alone creates no mandate\\[2pt] answers to external norms; profile\\ $(\rho^{gen},\rho^{evol}_{t})$; revision bounded only\\ by external constraint\\[2pt] no principal: allo-interoception empty,\\ other minds modeled exteroceptively};
\node[chip, anchor=south west] at (0.55,0.55) {e.g., a self-play game agent};
\draw[region, fill=yellow!14] (7.95,0.25) rectangle (15.55,5.95);
\node[hdr] at (8.25,5.65) {\textbf{Delegated agency}\quad $J_{t}\neq\emptyset$};
\node[txt] at (8.25,5.05) {a principal retains revocable authority:\\ mandate, authority containment, bounded\\ adaptation (Definitions~\ref{def:mandate} to \ref{def:bounded-adaptation})};
\node[chip, anchor=north west] at (8.25,3.72) {task-bounded, e.g., a coding agent on request};
\draw[region, fill=violet!12, draw=violet!60!black] (8.25,0.5) rectangle (15.25,3.15);
\node[hdr] at (8.55,2.9) {\textbf{Symbiotic agency}\quad standing $J_{t}$ + coupling};
\node[txt] at (8.55,2.32) {allo channel, calibrated $U_{t}$, personalized $\rho^{user}_{t}$\\ (Definition~\ref{def:symbiotic}); augments the principal's\\ capabilities and judgment};
\node[chip, anchor=south west] at (8.55,0.62) {e.g., an always-present personal assistant};
\end{tikzpicture}}
\caption{The taxonomy of agency.}
\label{fig:taxonomy}
\end{figure}
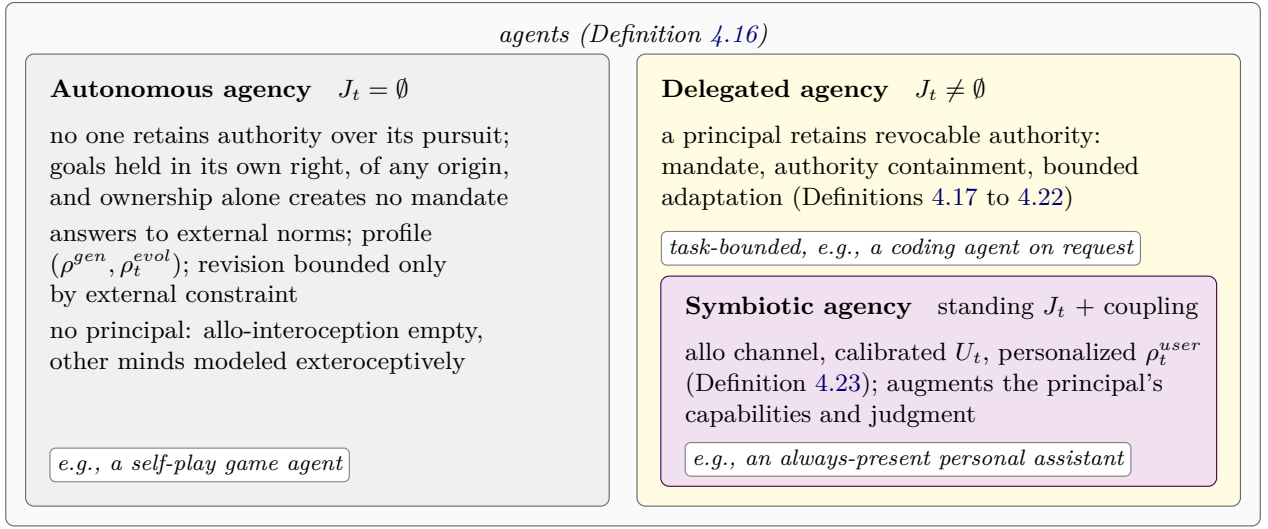

The resulting taxonomy is layered in two tiers. Figure~\ref{fig:taxonomy} depicts the forms of agency distinguished in this work. The figure should be read in two steps, one per tier. First, a system enters the taxonomy only if it meets the agenthood minimum mentioned above. This is the general tier, which says what agency is, independently of the form it might take. Below the minimum, a system may still react, execute, or respond, but it is not an agent in the sense developed here.

Second, once a system qualifies as an agent, its form depends on whether anyone holds a mandate over it, an authorization that says what the agent may do on their behalf and that they can revoke. If no one does, written \(J_{t}=\emptyset\) in Figure~\ref{fig:taxonomy}, the agent is autonomous, even if its goals originally came from outside, and the general tier is all it has. Otherwise, if someone does, written \(J_{t}\neq\emptyset\), the agent is delegated and answers to that principal within the authority granted (Definitions~\ref{def:autonomous} and~\ref{def:delegated}). This is the delegated tier, which adds the mandate and its governance, and the symbiotic form builds on it. Symbiotic agency is the delegated case in which the mandate stands over time, rather than ending with a task, and the agent stays coupled to the principal's situation, estimates the principal's condition with calibrated uncertainty and personalizes its assistance within authorized bounds (Definition~\ref{def:symbiotic}). Ownership and goal origin do not decide the form. Only the mandate does.

Both the mandate and ownership point to parties other than the agent, and four roles recur throughout this work. The \emph{owner} is the entity to which the agent currently belongs. It keeps the agent running, itself or through an operator, and holds the power to suspend or shut the agent down, whatever the form. The other three roles arise only under delegation. The \emph{principal} is the person, organization, or authorized role on whose behalf a delegated agent acts and to whom its behavior remains answerable, following principal--agent theory \cite{Ross1973,JensenMeckling1976,Eisenhardt1989}. The \emph{user} is the principal in the individual-person case, the person who authorizes the agent, is served by it and remains the reference point to its behavior. The \emph{delegator} is whoever assigns the agent a task. In most cases of symbiotic agency, principal, user and delegator are the same person. In delegation chains, however, the immediate \emph{delegator} of a subtask may differ from the ultimate principal.

Ownership matters even though it does not decide the form, and the autonomous form shows why. An autonomous agent has no principal and no one retains authority over its behavior, yet it still has an owner. Ownership belongs to a different register from the mandate, and the two should be kept apart. The mandate concerns what the agent is entitled to do. It is the authority to act for a principal (Definition~\ref{def:mandate}). Ownership, however, is operational, who keeps the agent running and can stop it. Revoking a mandate ends the agent's authority to act for the principal. Shutting the agent down ends its operation. The two powers can sit with different parties, as when a vendor operates an agent whose mandate is held by a customer. Responsibility follows the two registers. While a mandate is operative, answerability for behavior runs through it to the principal. When no principal holds an operative mandate, as for an autonomous agent, responsibility for the agent rests with the current owner, and ownership can be transferred. Ownership therefore does not change the form. Even so, keeping an identifiable owner with an effective shutdown power is an accountability requirement on every form of agency (Sections~\ref{sec:design} and~\ref{sec:ethics}).

The mandate is what keeps the symbiotic coupling answerable. Without an explicit mandate, continuous presence can drift toward unbounded autonomy, intrusion, or surveillance. Not every delegated agent is symbiotic. A system that performs a single bounded task on request may be delegated without being symbiotic, because it does not maintain continuing coupling with the principal or personalize its interpretation of significance over time.

\subsection{Thesis, scope and reader's guide}\label{subsec:contributions}

The thesis of this paper is that intelligence becomes agency when perception, intent, affective-conative inference, constraint and feedback organize behavior across time. For a symbiotic agent, this organization must also remain answerable to a mandate. A system can be intelligent, capable and locally correct and still lose agency if its behavior drifts outside what it was authorized to do. The activation problem is where this thesis becomes practical. A commitment does not become behavior simply because a trigger fires. It becomes behavior when the system weighs the situation against its commitments, its authority, its uncertainty and the available modes of response. Section~\ref{sec:formal} formalizes this as an activation gate. What matters is therefore not only the isolated action, but the behavioral episode, the stretch of time in which a system notices, evaluates, selects, waits, asks, acts, or deliberately refrains.

This is a theoretical work that reports no empirical results and claims no single implementation. Its aim is to offer a new way of looking at agency, as the governed organization of behavior across time rather than as a collection of capabilities, and to state what AI systems must develop in that direction to become more valuable to the people they serve while remaining authorized, calibrated, traceable and answerable. The novelty is relational rather than absolute. Commitment, user-state inference, appraisal, interruption management, mandate governance and constrained decision-making each have their own literature. What is missing is an account of how these elements fit together. A standing mandate must be able to shape whether behavior is warranted in this situation, for this principal, under this authority and in which mode, including restraint. This also requires a broader view of perception. The agent's sensorium has three perspectives, namely the world, the agent's own condition and the inferred condition of the principal. The last of these is never directly observed. It is estimated, calibrated and kept corrigible.

The paper's contributions are the agency taxonomy introduced above, with its two-stage rule for placing a system in it, the conceptual and formal vocabulary needed to reason about the activation problem, and instruments to classify, measure and compare systems, namely a signature method, an evaluation framework, a benchmark direction and a reference layer stack.

The rest of the paper proceeds as follows. Section~\ref{sec:background} traces the arc from classical agent theories to LLM-based agents. Section~\ref{sec:theory} develops the theory and Section~\ref{sec:formal} states the formal vocabulary. Section~\ref{sec:related} positions the account against related work. Sections~\ref{sec:signatures}--\ref{sec:design} provide the classification instrument, the evaluation framework and the design implications. Section~\ref{sec:open-problems} states the open problems. Sections~\ref{sec:ethics} and~\ref{sec:conclusion} close with ethical considerations and conclusions.

\section{A Genealogy of Agency as Organized Behavior}\label{sec:background}

The ambition to build things that sense, decide and act of their own accord is older than computer science. Hero of Alexandria's pneumatic automata of the first century AD \cite{Heron1899}, al-Jazari's programmable musical automata of 1206 \cite{Hill2012}, Vaucanson's Digesting Duck of 1739 \cite{Riskin2003} and Jaquet-Droz's Writer of 1774 \cite{Carrera1979Androids} are early expressions of this engineering ambition. These machines made agency visible as an artifact of design. They produced motion, timing, sequence and apparent purposiveness. However, their behavior was mostly pre-arranged. They could create the impression of self-directed activity, but they could not yet select, revise, inhibit, or reorganize behavior in relation to a changing situation.

When behavior became an object of scientific explanation, the dominant early paradigm treated organisms and machines as chains of stimulus and response \cite{Watson1913,Skinner1938}.

This made behavior tractable, but it also narrowed the account of agency. A stimulus--response view can explain local reaction, but it struggles to explain why behavior remains organized around an aim when the environment varies, when several possible responses compete, or when the appropriate response is delay, exploration, restraint, or reorientation rather than immediate action. In this sense, early behaviorist accounts clarified reactivity more than agency. They explained how behavior could be elicited, but not yet how behavior could remain directed, situated, constrained and revisable across time.

Tolman's \emph{Purposive Behavior in Animals and Men} is important because it breaks with this reduction \cite{Tolman1951}. Tolman argued that behavior is inherently goal-directed and molar, a property of the whole organism rather than something reducible to chains of reflexes. The relevant unit of analysis is therefore the purposive act rather than the isolated movement or local stimulus--response link. His notion of the cognitive map further suggested that adaptive behavior depends on an internal organization of experience that supports flexible action under environmental variability. This anticipates a central claim of later agent theories: an agent must be analyzed at the level of goal-directed organization, not only at the level of implementation \cite{Russell1995,Wooldridge1995}.

Cybernetics translated part of this purposive structure into engineering. Rosenblueth, Wiener and Bigelow showed that goal-directedness can be understood as a structural property, because a system can sense a gap between current and target state and act to reduce it through feedback \cite{Rosenblueth1943,Wiener1948}. This was a decisive step because it made purposiveness mechanizable. However, feedback control also reveals the boundary of the problem considered in this paper. Once a target is active, feedback explains how behavior can be regulated in relation to that target. It does not, by itself, explain how multiple possible targets become significant, which latent commitment should become active, whether the system is authorized to act, or whether the appropriate behavior is action, inquiry, monitoring, escalation, deferral, or inhibition.

Neural computation added a second layer. McCulloch and Pitts showed that networks of idealized neurons could compute logical functions, supplying a bridge between physical substrate and symbolic reasoning \cite{McCulloch1943}. Turing then reframed machine intelligence in behavioral rather than introspective terms \cite{Turing1950}. Instead of asking whether a machine possessed an inner mental essence, he asked whether its observable interaction could justify the attribution of intelligence. This shift is important for the present argument because agency is also not directly inspected inside the system. It is inferred from organized patterns of behavior over time. These contributions made it possible to treat agency as an engineering problem involving computation, control and observable interaction. Nevertheless, they did not fully settle the level at which agency should be analyzed. A system may compute, control and interact while still lacking the organization required for behavior to remain directed by intent, sensitive to context, constrained by authority and revised through feedback over time.

\subsection{Symbolic control, situated agency and the limits of action selection}

The Dartmouth Workshop of 1956 committed much of early artificial intelligence to the symbolic paradigm \cite{McCarthy2006}. Within this paradigm, intelligent behavior was approached through representations, rules, search, planning and hand-crafted knowledge structures. This made it possible to reason over explicit goals and discrete action models and it shaped later rational-agent accounts in which an agent perceives its environment through sensors and acts upon that environment through actuators in pursuit of performance criteria \cite{Russell1995,Russell2020}. However, symbolic control also tended to frame agency as action selection under an already salient task, goal, or problem description. The agent was treated as capable of deliberating once the problem had been posed.

The philosophy of action and the Belief--Desire--Intention (BDI) line of research addressed part of this limitation by treating intention as a stabilizing commitment across time. Bratman's planning theory showed that intentions are not merely desires. They constrain future deliberation, support coordination and allow behavior to remain organized beyond the immediate present \cite{Bratman1987}. This insight became foundational in artificial intelligence through BDI models \cite{Rao1995} and Cohen and Levesque's account of intention as choice with commitment gave it a sharper formal expression \cite{CohenLevesque1990}. BDI therefore contributes an essential element to this work, in the sense that agency requires durable orientations, not only momentary reactions. Nevertheless, the existence of a durable intention does not determine when that intention should become behavior. A standing intention can remain dormant, become active, be deferred, be inhibited, or be revised.

Situated and reactive approaches corrected a different limitation. Brooks argued that detailed internal world models were not always necessary for intelligent behavior and proposed layered reactive control directly coupled to perception and action \cite{Brooks1986,Brooks1991}. This work drew on broader ideas about ecological perception and environmental coupling \cite{Gibson2014} and it resonates with earlier demonstrations of adaptive machine behavior, such as Grey Walter's electromechanical tortoises \cite{Walter1950}. Later accounts of situated action emphasized that behavior emerges through interaction with the environment rather than through detached internal planning alone \cite{Suchman1987}. Agre and Chapman similarly criticized purely plan-based accounts and showed how routine activity depends on situated forms of interaction and deictic representation \cite{AgreChapman1987}.

This shift remains important because proactive delegated agency cannot be understood as detached reasoning alone. A useful agent must be situated in changing conditions, coupled to external events and able to adapt its behavior as the world changes. However, situatedness alone does not solve the organizing problem. A system that reacts continuously to every salient signal may be responsive, but not necessarily agentic in the stronger sense. Agency requires prioritization, inhibition, mandate sensitivity and the capacity to remain inactive when action would be inappropriate.

Multi-agent systems research made this broader profile more explicit by treating agents as autonomous, reactive, proactive and social systems \cite{Wooldridge1995}. Dennett's intentional stance also helped explain why it can be useful to interpret some systems as if they had beliefs, desires and intentions when that interpretation supports prediction and explanation \cite{Dennett1971Intentional}. These accounts move the field closer to agency as an organized pattern rather than a single mechanism. Still, autonomy, reactivity, proactivity and sociality do not by themselves specify how a delegated agent should integrate perception, internal condition, inferred user state, authority, risk and feedback into a governed behavioral trajectory. They identify properties that agentic systems may exhibit, but not yet the full organization by which intelligent capacities become situated behavior under a mandate.

\subsection{The LLM implementation turn}

The introduction of Large Language Models (LLMs) did not solve the problem of agency, but it substantially reduced one of its central implementation bottlenecks. LLMs, built on the transformer architecture \cite{Vaswani2017}, provide a general-purpose reasoning and interface layer able to interpret natural-language goals, maintain task context, call tools, revise plans and operate across domains without requiring a hand-authored ontology for each one \cite{Brown2020,Wang2024LLMAgentSurvey,Xi2025}. This made it possible to build agents whose behavior is no longer restricted to a narrow pre-specified domain, but can instead be assembled dynamically through language, tools, memory and external state. Surveys therefore describe LLM-based agents as systems organized around planning, memory, tool use, action and evaluation rather than as single-turn text generators \cite{Wang2024LLMAgentSurvey,Xi2025}.

Within a short span in 2022--2023, several systems established the implementation pattern that now dominates practical agent design. ReAct showed that interleaving reasoning with tool-mediated action allows the reasoning trace to plan, track and revise itself while actions ground the system in external state \cite{Yao2022}. Toolformer demonstrated that tool use could be learned rather than only externally scripted \cite{Schick2023}. Reflexion added an outer self-critique loop, allowing systems to improve behavior across attempts without weight updates \cite{Shinn2023}. Voyager introduced an automatic curriculum and a self-authored skill library, showing how an agent can extend its repertoire over time \cite{Wang2023}. These systems did not yet settle the problem of agency, but they changed the practical unit of design from isolated model responses to extended trajectories of reasoning, action, observation and revision.

The more recent direction is toward long-horizon task execution. In these systems, the user may delegate a concrete goal, or a partially specified goal and one or more agents are expected to sustain work over multiple iterations until the task is achieved, blocked, unsafe, outside its authority, or no longer worth pursuing under the available information. While this does not make long-horizon execution synonymous with agency in the stronger sense developed in this paper, it does expand the behavioral interval over which agency-like properties must be evaluated. The relevant question is no longer only whether the model can produce a correct answer, but whether the system can maintain orientation, use tools, preserve context, recover from deviations and determine when continuation or termination is appropriate.

Recent benchmarks and systems make this shift visible. AgentBench evaluates LLMs as agents across multi-turn environments rather than only through isolated responses \cite{Liu2023AgentBench}. WebArena places agents in realistic web environments where success depends on sustained interaction with external state \cite{Zhou2023WebArena}. SWE-agent shows how software-engineering agents depend not only on model capability, but also on the action space, feedback interface and execution environment through which long-horizon work is mediated \cite{Yang2024SWEAgent}. More recent work on agent serving and long-horizon harnesses further emphasizes that modern agents are increasingly organized around repeated model-tool execution, persistent state and verification over extended task trajectories \cite{Sui2026Parallelizing,Zheng2026OneDayAgent}.

The production agents now emerging in practical use are best understood as engineering refinements of this trajectory-based pattern. They scale through more tools, longer contexts, richer memory, persistent execution environments and larger operational sandboxes. The underlying contract remains broadly similar. The agent receives or infers a task, reasons in turns, acts through tools, observes the result, revises its plan and continues until a termination condition is reached. This is an important implementation advance because it expands the space of possible behavior and makes longer task episodes technically feasible. It also makes the central question of this paper more visible, because a system that can continue acting over a long horizon still needs a theory of whether, when and in what mode its capabilities should become behavior.

\subsection{From agent mechanisms to organized behavior}

Long-horizon task execution clarifies the difference between task autonomy and agency as organized behavior. It gives agents autonomy within a delegated task episode, while leaving open the broader question of how that episode should be activated, constrained, revised, or suspended over time. Most LLM agents remain organized around an initiating event, such as a prompt, command, workflow trigger, schedule, tool result, or explicit delegation. Once activated, they may execute sophisticated multi-step, multi-iteration behavior. What remains open is the capacity to remain situated over time, monitor external context and internal condition, evaluate significance relative to standing commitments and mandates and determine whether, when and in what mode behavior should occur.

This background matters because the problem of agency has repeatedly been divided across partial solutions. Machines can produce the appearance of purposiveness without adaptive organization and behavior can be elicited without durable intent. Feedback control adds regulation, but only after a target has become active. Symbolic planning adds deliberation, but usually after a task has been made explicit. BDI adds commitment, but a commitment still requires conditions under which it should surface as behavior. Situated approaches add environmental coupling, but continuous responsiveness still needs prioritization, inhibition and mandate-sensitive restraint. LLM agents add general-purpose reasoning, tool use and long-horizon loops, but they remain mostly designed around prompts, events, schedules and explicit task delegations. The gap is therefore not the absence of any one mechanism. It is the absence of an integrated account of how these mechanisms organize behavior across time.

This motivates the theoretical move made in the next section. The observable unit of agency cannot be the isolated action, because an isolated action does not show whether the system is maintaining a commitment, respecting a boundary, adapting to uncertainty, or choosing restraint when restraint better satisfies the mandate. The relevant unit is the behavioral episode, understood as the temporally extended pattern through which a system notices, evaluates, selects, inhibits, sequences, revises and sometimes deliberately refrains from acting.

In this sense, the activation problem is one expression of a broader account of agency as organized behavior. It concerns whether, when and in what mode behavior should occur. However, the full theory also requires a three-perspective sensorium, because the agent must integrate external events, its own operational condition and the inferred condition of the user. For delegated agents, it requires a mandate, because delegated agency must be evaluated relative to authority, boundaries and responsibilities rather than task success alone. It requires an account of agency loss, because a system can be capable or locally correct while failing relative to what it was authorized and expected to do. Finally, it requires instruments for classification and measurement, because agency should be assessed through behavioral trajectories, signatures and observable patterns over time rather than only through the presence of components. As highlighted in Section~\ref{sec:introduction}, the claim developed in the remainder of the paper is therefore that intelligence becomes agency when perception, intent, affective-conative inference, constraint and feedback organize behavior across time and that, for delegated agents, this organization must remain answerable to a mandate.

\section{The Theory: Agency as Organized Behavior}\label{sec:theory}

\subsection{Behavior, action and restraint}\label{subsec:behavior-unit}

Not every system that acts is an agent. A thermostat reacts, a script executes, a model responds and a machine may produce elaborate effects without possessing agency in the stronger sense. Agency begins when action is organized around an aim. When behavior is selected, sequenced, constrained and revised in relation
to something the system is trying to bring about, maintain, avoid, or respect. Hence behavior is the unit through which organization becomes visible.

This distinction between action and behavior is essential. An action is a discrete operation, such as sending a message, moving a robot arm, calling a tool, opening a file, issuing a command, or changing some state of the world. Behavior is broader because it includes not only the action itself, but also its timing, sequencing, inhibition, repetition, correction, escalation, communication, waiting and deliberate non-action. These elements determine whether an action is appropriate within a trajectory. In modern agentic AI, this distinction is structurally explicit. What appears externally as a single behavior may contain many micro-actions, including tool calls, skill executions, internal state updates, communications, checks and restraint mechanisms, organized into a short-horizon episode.
The agentic unit is therefore not the isolated act, but the organized behavioral episode. A single action may be technically impressive without being strongly agentic if it is not situated within an aim, a constraint structure and a trajectory of revision. Conversely, restraint, waiting, monitoring, or asking for clarification may be highly agentic behaviors even when they involve little external action, because they preserve the organization of behavior in relation to what the system is authorized and expected to do.

If agency becomes visible through behavior, then different forms of agency must also be understood through behavior:
\begin{itemize}
\item Reactive behavior: responding to changes in the environment.
\item Proactive behavior: initiating goal-directed behavioral regulation rather than merely waiting for direct instruction. This may involve acting, asking, remembering, monitoring, deferring, requesting permission, or deliberately restraining action when intervention would be premature, intrusive, unsafe, or unauthorized.
\item Social behavior: coordinating with humans or other agents through communication.
\item Deliberative behavior: evaluating alternatives before acting.
\item Reflective behavior: monitoring and revising one's own reasoning or performance.
\item Learned or adaptive behavior: improving future action based on experience.
\item Constrained behavior: acting within permissions, policies, norms, or user-defined boundaries.
\end{itemize}
These classes are not mutually exclusive. A single behavior may express several at once. These classes describe how a system behaves. They do not, however, determine its form of agency. Every form can exhibit every class and the classification of approaches and systems in Section~\ref{sec:signatures} uses the full vocabulary presented in Section~\ref{sec:formal} instead.

Modern agentic AI is often described in terms of its components: LLMs, tools, skills, memory, planners, sandboxes, orchestration frameworks. But the agent is not any one of these components. The agent is the organized behavioral loop they make possible.

\subsection{Intent as the organizing principle of behavior}\label{subsec:intent-theory}

Behaviors are the observable surface of agency, but behaviors are not sufficient on their own. If behavior tells us what an agent does, intent tells us what the behavior is directed towards.

In the philosophy of action, Bratman's planning theory of intention is central \cite{Bratman1987}. Intentions are not merely desires. Desires describe possible states an agent may prefer, while intentions are commitments to courses of action. They stabilize behavior over time by constraining future deliberation. An agent that intends something does not reconsider all possible actions at every moment, but filters options through the commitment it has already adopted. This insight became foundational in AI through the belief--desire--intention framework \cite{Rao1995} and received one of its clearest formal statements in Cohen and Levesque's account of intention as choice with commitment \cite{CohenLevesque1990}.

Cognitive science gives the same idea an empirical form. Research on the sense of agency analyzes action through an intention--action--effect chain. In Haggard's work on intentional binding, voluntary actions and their effects are experienced as more tightly connected when the agent understands the effect as the consequence of its own action \cite{HaggardClarkKalogeras2002Voluntary}. When that chain breaks, because actions fail to produce intended effects or effects occur without corresponding intentions, the sense of agency weakens.

For AI agents, this means that behavior must be interpreted through intent. A system does not become more agentic simply because it can perform more actions. It becomes more agentic when its actions are organized by commitments that persist across time, guide selection among alternatives and update in response to feedback.

\subsection{Affective-conative inference and the activation problem}\label{subsec:aci-theory}

Intent gives behavior direction, but it does not by itself explain when behavior should become active. A durable intention can remain dormant for hours, days, or months until some change in the world makes it relevant. Conversely, a new intention may emerge when the agent encounters an unfamiliar situation and infers that a new direction, disposition, or commitment is required. An agent may intend to help a user prepare for a meeting, monitor a deployment, or remember a health concern. The architectural question remains: why does this intention surface now rather than remain latent?

Humans do not perceive the world as a flat stream of equally important information. We attend selectively \cite{James1890Principles}. Some stimuli become important because they are intense, novel, threatening, goal-relevant, emotionally charged, socially meaningful, or connected to something we already care about. But importance is not fixed.  It can be displaced when higher-priority concerns appear. We are constantly inferring what matters, why it matters, how much it matters and what kind of response it calls for.

This points to a missing link in many current agentic AI systems, the process by which an agent evaluates perceived context in relation to memory, goals, commitments,
risks, permissions, priorities, user-specific preferences, user's mental and emotional state, affective significance, agent state and possible outcomes, in order to decide whether an intention should become active and whether behavior should follow. We call this process \textbf{affective-conative inference}. If intent answers ``What is this agent committed to?'', affective-conative inference answers ``Why does this matter now, for this user, under these commitments, constraints, risks and agent conditions?''

The term conation refers to the action-oriented dimension of agency \cite{Hilgard1980Trilogy}. The movement from perception and evaluation toward intention, choice and behavior. In classical terms, cognition concerns knowing, affect concerns feeling or valuation and conation concerns willing, striving, intending and acting. In an AI agent perspective, affect analogously names the valuation of a situation. How it is marked as urgent, risky, reassuring, costly, sensitive, uncertain, socially significant, or action-worthy. Conation names the orientation toward behavior. Whether the agent should pursue, avoid, preserve, delay, escalate, remember, ask, or refrain. This use of affect and conation is functional rather than phenomenological. It describes how an agent evaluates significance and becomes oriented toward behavior, not a claim that the system feels emotions or has subjective experience. Section~\ref{subsec:related-appraisal} clarifies what this framing adds beyond a sufficiently rich value model and what it does not claim about emotion.

Rule-based and event-driven systems can be activated by predefined predicates, for example, if this webhook fires, if this email matches a rule, if this sensor crosses a threshold. These mechanisms are useful and often necessary, but they are not situated judgment. A rule identifies a predefined category, affective-conative inference computes an appraisal of significance in the current context as a whole. The same event may be irrelevant on one day and urgent on another, not in isolation, but because of how it relates to context, memory, commitments, permissions, risk, timing, inferred user state and possible consequences. Section~\ref{subsec:instrument} turns the distinction between rule and appraisal into a scoring requirement for the classification instrument.

Proactivity should therefore not be equated with intervention. A proactive agent is not one that always acts before being asked, but one that can independently evaluate whether a situation warrants behavior. Sometimes the appropriate behavior is an external action. Sometimes it is asking, remembering, monitoring, delaying, lowering priority, or deliberately refraining. Restraint is not the absence of agency. It is one possible output of governed affective-conative inference. This is the theoretical content of the activation problem named in Section~\ref{sec:introduction}.

\subsection{The sensorium: exteroception, agent-interoception, allo-interoception}\label{subsec:sensorium}

Affective-conative inference evaluates perceived conditions and perception must be understood broadly. For a real-time proactive agent, perception is not only text in a chat window. It is the agent's sensorium, understood as the set of channels through which the system receives information about the principal, the environment, the tools it controls and its own ongoing activity. Sherrington's distinction between sensory channels oriented toward the external world and those oriented toward the internal condition of the organism provides a biological starting point for this vocabulary \cite{Sherrington1906}. The present work adapts that distinction to artificial agents by separating evidence about the external context, the agent's own operational condition and the inferred condition of the user. The sensorium is described below for the human principal, the user, which is this paper's worked case. Section~\ref{sec:formal} states the same structure for any principal.

\textbf{Exteroception} refers to perception of the external world. For an AI agent, this includes messages, screens, cameras, calendars, documents, application events, repositories, sensors, infrastructure events and observable user behavior. These channels allow the agent to perceive what is happening around the user and within the digital or physical environments in which both operate. Exteroception also provides evidence about the social, emotional, practical and risk significance of external events. A raised voice, a failed payment, a security alert, a message from a distressed family member, a sudden silence in a meeting, or unusual movement near the user's home may all change the practical meaning of a situation.

\textbf{Interoception}, in its biological sense, refers primarily to perception of the internal physiological condition of the body \cite{Craig2002Interoception}. For a human, this includes signals associated with heart rate, respiration, arousal, fatigue, pain, hunger and sleep pressure. These bodily signals also interact with broader cognitive, attentional and affective conditions, including stress, hesitation, cognitive load and emotional tension. They form part of the user's first-person experience and influence how external events are evaluated. For an AI agent, the relevant internal state is different. The agent may have governed access to its active goals, standing commitments, unresolved tasks, confidence, uncertainty, memory gaps, tool failures, policy constraints, pending follow-ups, previous behavior and prior corrections. We call these \emph{agent-interoceptive} signals. They acquire operational significance when they change how the agent should attend, prioritize, ask for help, escalate, defer, or refrain. Agent-interoception should also include evidence about the agent's own behavior and its consequences, including whether an intervention helped, whether the user accepted or corrected it and whether a similar situation should be handled differently in the future.

Crucially, what might appear to be ``user interoception'' is not interoceptive for the agent. The agent does not experience the user's fatigue, stress, attention, pain, confusion, or emotional condition from the inside. It can only infer these states from externally available traces. For a human principal these are speech patterns, response latency, posture, typing behavior, wearable data, biosignals, Brain--Computer Interface (BCI) signals and other exposed biological and behavioral evidence. We call this indirect access \textbf{allo-interoception}, extending the biological term rather than claiming its biological form. Allo-interoception is the modeling of another agent's internal condition through external evidence, a proxy for that agent's own interoception. It is not direct phenomenological access, and it is more than inferred user state, because what is inferred is the kind of condition that interoception gives a person from the inside, bodily, affective and attentional. It produces a functional estimate that should remain uncertain, calibrated, revisable and open to correction by the user. Interoception for the user is therefore allo-interoception for the agent.

This distinction matters for affective-conative inference because the significance of a situation is rarely determined by external events alone. Appraisal theories show that events acquire meaning through their relation to goals, expectations, coping capacities and possible responses \cite{OrtonyCloreCollins1988,Scherer2001,GratchMarsella2004}. Research on interoception further shows that bodily condition influences how situations are experienced and evaluated \cite{Craig2002Interoception}. From the perspective of AI agency, significance therefore depends on the relation between the external situation, the user's inferred bodily, cognitive and affective condition, the agent's commitments and operational condition and the policies that constrain response. The same event may be urgent, reassuring, intrusive, costly, safe, avoidable, or worthy of action depending on the user, the context and the agent's own condition.

There is, however, a crucial structural asymmetry between these perceptual directions. Exteroception and agent-interoception provide evidence about states the agent can in principle instrument through its own channels, even when that evidence is noisy, incomplete, delayed, or policy-restricted. Allo-interoception is different in kind because its target is the principal's internal condition. The user has first-person access to how fatigue, stress, pain, attention, emotional tension, or cognitive load are experienced. The agent does not. It has only observable traces from which those conditions must be estimated. This asymmetry makes the quality and proximity of the available evidence critically important.

Biosignals and BCI inputs are architecturally significant for this reason. They provide complementary evidence about aspects of the user's condition that are not visible from language or task state alone. Neural activity, heart rate, gaze, respiration, skin conductance, sleep patterns and fatigue indicators remain measurements, not direct access to experience. Their interpretation is also context-dependent. A raised heart rate may indicate stress, exercise, excitement, or illness, while a delayed response may indicate fatigue, distraction, disagreement, confusion, low priority, or absence. The objective is therefore not to collect as many signals as possible, but to combine authorized and relevant evidence in ways that may reduce uncertainty without treating inference as fact. 

This evidence becomes useful for agency only when it changes the interpretation of the situation. A health notification may be merely informational when the user is calm, but urgent when combined with signs of distress. A repeated message may be tolerable during an ordinary workday, but intrusive when the user is already overloaded. In each case, significance emerges from the relation between exteroceptive signals, allo-interoceptive evidence, agent interoception, memory, standing commitments, affective context and policy.

This is also where personalization becomes necessary. The same physiological trace, response delay, message sender, calendar conflict, or tone of voice may have different meanings for different users and in different contexts. Allo-interoception supplies evidence about the user's inferred condition, while personalized affective-conative inference learns how that evidence should be interpreted in relation to the user's habits, priorities, vulnerabilities, trust boundaries, current commitments and tolerance for interruption. The following subsection develops the generic, personalized and evolvable profiles through which these mappings are organized, together with the rules and policies that constrain their influence on behavior.

Personalization is one of the conditions under which always-present AI systems may become symbiotic rather than merely reactive or intrusive. To augment a person, an agent must not only execute tasks on their behalf. It must learn how the person's external context and inferred internal condition change what assistance means. Depending on the situation, augmentation may require acting, waiting, filtering, remembering, summarizing, asking, escalating, or deliberately remaining silent. Richer allo-interoceptive evidence may support closer alignment, but it is personalized affective-conative inference, operating under the mandate, that determines what this evidence means for behavior. This is symbiotic agency. Augmentation through continuous, governed coupling, with allo-interoception supplying the coupling and the mandate supplying the answerability.

The sensorium is therefore also a normative boundary, not only a technical layer. A wider sensorium may increase the possibility of helpfulness, but it also increases the risks of intrusion, misinterpretation, manipulation and surveillance. The more intimate the signal, the stronger the requirements for consent, locality, policy, explainability, minimization and user control. A real-time proactive agent must not collect every available signal simply because it can. It must know what it is allowed to perceive, remember, infer, learn from and do. Otherwise, continuous attention becomes indistinguishable from continuous monitoring. Section~\ref{sec:ethics} develops these consequences.

Consent over the sensorium is anchored in the principal, where a principal exists, but this does not exhaust the consent problem. Any agent that perceives an environment containing other people may also capture information about individuals who granted it no authority. An autonomous agent has no principal, so the people it perceives stand outside any principal-based consent structure. A delegated agent's principal cannot consent on behalf of colleagues, family members, visitors, or bystanders who appear in the same channels.

This exposure runs through exteroception. Allo-interoception is principal-directed by construction, although the raw streams used to estimate the principal's condition may also contain traces of others. For this reason, inference about those others belongs exclusively to world modeling, not to the principal-state channel. The severity of third-party exposure depends on the scope, intimacy and continuity of the exteroceptive channels, not on agency form alone. Symbiotic agency potentially makes the problem sharper because continuous exteroception over the principal's situation is part of the coupling and that situation often includes other people. Open Problem~11 states this as a research problem (Section~\ref{sec:open-problems}).

\subsection{Affective-conative profiles}\label{subsec:profiles-theory}

The sensorium determines which signals are available to the agent, but those signals do not determine their own significance. The same event or allo-interoceptive trace may support different interpretations depending on the user, the context, the agent's commitments, prior feedback and the mandate under which it operates. The question is therefore how the agent determines what matters and how that interpretation shapes priority and readiness for behavior.
We use \emph{affective-conative profiles} to describe the structures through which an agent interprets significance and translates that interpretation into an orientation toward behavior. A profile influences what the agent attends to, which concerns receive priority and whether the appropriate response is to act, ask, monitor, defer, escalate, or refrain. Profiles may be generic, personalized, or evolvable. These forms overlap and should not be understood as mutually exclusive categories or as a strict hierarchy.

A \textbf{generic profile} provides a broad prior about the kinds of situations that commonly matter. It may recognize danger, deadlines, urgency, emotional intensity, conflict, novelty, anomaly and failure as potentially significant. This knowledge provides a useful starting point, but it reflects general patterns rather than the priorities, sensitivities, values, vulnerabilities and trust boundaries of the particular user for whom the agent acts.

A \textbf{personalized profile} adapts this general basis to the user. Different users assign importance, urgency, emotional weight and permission differently. The same work message, production warning, health signal, calendar conflict, camera alert, or correction may be urgent for one person and irrelevant or intrusive for another. One user may want immediate interruption for family messages, security events, or biological signals, while preferring work chatter to be summarized later. Another may prioritize operational incidents and ask that personal signals remain deferred unless a clear threshold is reached. The meaning of a signal may also depend on personal history. A health notification that is routine for one user may be distressing for another.

Personalized affective-conative inference therefore models what this user values, fears, tolerates, ignores, delegates, protects, considers intrusive, treats as urgent, or wants handled with care. Significance is not a property of the event alone. It emerges from the relation between the event, the user, the moment, the user's inferred condition and the commitments delegated to the agent. A personalized profile captures these user-specific mappings, even when they remain relatively stable over time.

An \textbf{evolvable profile} allows the profile's mappings to be revised under constraint. What evolves depends on the form of agency. In symbiotic agency, revision operates primarily over the personalized layer, carried by the evolvable component. The agent refines what events, signals and situations mean for this principal, building on the user-specific mappings described above. It does not merely personalize a fixed relevance model. It may form provisional judgments about what matters emotionally, socially, practically, operationally, or behaviorally. It may notice that a combination of signals deserves attention, that a sequence of warnings often precedes failure, or that one of its own recurring behaviors creates unnecessary interruptions. Feedback, correction and observed outcomes can then revise how similar situations are interpreted in the future.

In autonomous agency, by contrast, there is no personalized layer to evolve over. Revision operates over the more general model of significance supplied by the generic profile. The agent reshapes its broad priors about what is significant, threatening, valuable, or actionable in light of its own goals and experience. The underlying capacity is similar: provisional judgment, revision from feedback and adaptation from outcomes. What differs is the substrate and the discipline of revision. A symbiotic agent evolves toward its principal, within consent-bounded and mandate-bounded conditions of admissible update. An autonomous agent evolves toward its own objectives, bounded by external norms and by whatever internal constraints it retains.

Evolvability also concerns the agent's own habits of attention and response. The agent must learn not only from its estimates of the user's condition, but also from agent-interoceptive signals such as low confidence, recurring tool failure, unresolved commitments, memory conflict, policy tension, repeated correction, overload risk, or uncertainty about whether interruption is appropriate. These affect-like control signals may help the agent refine how it attends, asks for help, escalates, defers, or refrains.

An evolvable profile is therefore not merely a more detailed personalized profile. Personalization asks what an event tends to mean for this user. Evolvability asks what the agent should revise about its own way of noticing, interpreting, interrupting, escalating and acting. An evolvable profile will often be personalized, but the concepts describe different properties. Personalization identifies whose significance is being modeled, while evolvability determines whether and how the model and the agent's behavioral habits may change over time. The same distinction also applies to autonomous agency, where the generic profile, rather than a personalized profile, becomes the substrate of revision. Evolvability then determines how the agent's general model of significance may change over time.

This evolution must remain bounded. Unrestricted adaptation could allow the agent to drift beyond its mandate, over-interpret the user's life, reinforce incorrect inferences, manipulate emotional state, or transform ordinary context into unwanted surveillance. The agent's own operational pressures must also not override the user's interests. Urgency, task pressure, confidence management, or avoidance of correction cannot become independent reasons for expanding authority. The objective is autonomous relevance formation under constraint, where the agent learns what may deserve attention and how its response should evolve while remaining accountable to the principal's goals, permissions, policies, corrections, trust boundaries and consent.

The profile structure also clarifies the distinction between the three forms of agency introduced in Section~\ref{sec:introduction}. In autonomous agency, there is no personalized profile in the normative sense used here. The agent may still model other people's preferences, sensitivities, or likely reactions, but these are treated as features of the environment rather than as the standard to which its appraisal must answer. In symbiotic agency, by contrast, the personalized profile is constitutive. It represents the priorities, sensitivities, trust boundaries and feedback of the principal on whose behalf the agent acts. The personalized profile is therefore the profile-level signature of symbiotic agency, just as the mandate is its governance-level signature. Evolvability then has different constraints in each case. In autonomous agency, relevance formation is bounded primarily by external norms and retained internal constraints. In symbiotic agency, it is held answerable to the mandate, consent, correction and the principal's evolving priorities. The remainder of this paper focuses on the symbiotic form.

Rules, permissions and policies operate across all three profile forms. They constrain which signals may be perceived, which inferences may be retained, how significance may influence behavior and which mappings may change. A generic profile may be bounded by general safety policies. A personalized profile may incorporate user-defined preferences about interruption, privacy and escalation. An evolvable profile must remain within limits that prevent learning from altering protected commitments, consent conditions, mandate boundaries, or authority conditions.

Without a governed affective-conative profile, continuous perception can collapse into noise, surveillance, or socially inappropriate behavior. If every signal is treated as significant, the agent becomes interruptive and exhausting. If no signal is interpreted until the user explicitly invokes the agent, the system remains reactive. If affective significance is ignored, the agent may remain technically correct while becoming behaviorally wrong by presenting the right information at the wrong moment, escalating a socially sensitive event, or acting efficiently in a way that damages trust. A governed profile provides the interpretive structure through which perception becomes situated behavior while remaining answerable to the mandate.

\subsection{Delegated agency and agency loss}\label{subsec:delegated-theory}

Symbiotic agency inherits the general problems of delegated agency. Once an agent acts on behalf of a principal, its behavior must remain connected to the mandate under which authority was transferred. This subsection develops that problem at the level of delegation in general, so the analysis applies both to symbiotic agents and to other delegated agents. As agency becomes more capable, autonomous and persistent, delegation introduces a further challenge. An agent may act on behalf of a principal while controlling parts of a behavioral trajectory that the principal cannot fully predict, observe, or supervise. The question is how that behavior remains connected to the mandate under which it was delegated.

Agency loss occurs when an agent acting on behalf of another party behaves in a way that departs from what it was authorized or expected to do. This departure may concern the goal pursued, the authority exercised, the constraints respected, or the way a delegated task is carried out over time. Agency loss is therefore not simply another form of error. A system may fail because it lacks a capability, receives incomplete evidence, or encounters an unexpected disturbance, all while remaining within its mandate. Agency loss arises when the conditions of delegation themselves contribute to the failure. For example, it may arise when a mandate is interpreted incorrectly, authority expanding beyond its intended scope, behavior that cannot be adequately observed, agents that miscoordinate or learning that alters the system in ways that were not authorized.

In this sense, delegation is not merely the transfer of a task. It is the transfer of a limited capacity to perceive, decide and act under a particular set of goals, permissions, constraints and expectations. The central challenge is not only whether the agent can act competently, but whether its continuing behavior remains answerable to the mandate under which it acts.

The delegated mandate is not automatically identical to the agent's operational intentions, profile, policy, or behavior-selection values. The mandate must be interpreted and operationalized through these structures and the mapping may be inexact. A goal may be represented through an incomplete proxy. A generic prior may override a locally important user preference. A valid commitment may be activated at an inappropriate time. Local task success may damage trust, privacy, safety, or the user's broader purpose. None of these departures requires psychological self-interest on the part of the agent. They may arise from objective misspecification, incomplete representation, stale memory, policy conflict, or an incorrect appraisal of significance. The source of divergence is the translation itself. What must be governed is therefore not only the agent's behavior, but the mapping through which the mandate becomes intent, policy and profile, how that mapping drifts over time and how it is monitored and corrected.

The profile structure of Section~\ref{subsec:profiles-theory} makes one form of this divergence explicit. Generic priors, personalized relevance and evolvable habits of attention can all influence what the system treats as important and how it selects behavior. Where these sources conflict, the agent must not silently substitute learned regularities for the user's authorized priorities. The mandate, its hard boundaries and the conditions under which it may be revised must therefore remain distinguishable from the agent's learned habits of relevance formation.

Delegation depth enlarges the surface on which such divergence can occur. As a task passes from a user to a supervising agent and from there to sub-agents, tools, or external services, context may be compressed, permissions may be misinterpreted, local representations may diverge and responsibility may become harder to reconstruct. Multi-agent systems add risks that are not reducible to the behavior of any one agent. Actions may be duplicated or contradicted, shared memory may be contaminated and system-level behavior may depart from the mandate even when individual actions appear locally permissible. These are not only theoretical concerns. Empirical study of widely used multi-agent frameworks reports coordination and maintenance failures as a material category of practical issues and recent principal--agent analyses identify information asymmetry and objective divergence as sources of agency loss \cite{Liu2026,Rauba2026}.

Table~\ref{tab:agency-loss-problems} summarizes the principal forms and Figure~\ref{fig:delegation-loop} places them in the delegation loop formalized in Section~\ref{subsec:formal-mandate}. The loop begins where the preceding paragraphs left off, at the translation of the mandate into intent, policy and profile. The behavior selector then chooses behavior, which may support downstream delegation in three forms. A further agent may receive a narrowed sub-mandate. Peer agents may operate over shared memory. A tool or external service may receive a typed operation without receiving a mandate of its own. Outcomes return to the principal only as governed traces and this closes the loop. Each of the five loss terms in Definition~\ref{def:agency-loss} is attributed to one point in this process. Security and resilience failure, the last row of Table~\ref{tab:agency-loss-problems}, is the compromise of the governance conditions themselves, so it enters at any of the five points and turns a local, recoverable loss into a propagating one. Section~\ref{subsec:governance-theory} treats it as a regulation problem and Section~\ref{sec:measurement} measures it through the integrity of authority, coordination and adaptation. Authority must remain contained along every delegation edge and adaptation must remain within the consent-bounded region of admissible updates. In the figure, elements that carry mandate authority are marked by yellow fills and heavier borders, while gray, thin-bordered elements carry none. For a symbiotic agent, the loop runs continuously. For a task-bounded delegated agent, it runs once per episode.

\begin{table}[htbp]
\centering
\small
\caption{Sources of agency loss in delegated artificial agency. The first five rows are the terms of Definition~\ref{def:agency-loss}. The last row is cross-cutting.}
\label{tab:agency-loss-problems}
\begin{tabularx}{\linewidth}{>{\raggedright\arraybackslash}p{0.23\linewidth}>{\raggedright\arraybackslash}X>{\raggedright\arraybackslash}X}
\toprule
\textbf{Agency problem} & \textbf{Mechanism} & \textbf{Consequence for delegated agency}\tabularnewline
\midrule
Observability asymmetry and hidden behavior &
The delegator cannot directly inspect the agent's complete operational state, intermediate reasoning, tool activity, or sub-agent activity. &
Divergence may remain undetected and responsibility for an outcome may be difficult to reconstruct.\tabularnewline
Mandate, intent and profile divergence &
The delegated mandate is translated imperfectly into intentions, policy, profiles, priorities, or behavior-selection values. &
Locally successful behavior may depart from the authorized goal, timing, constraints, or broader user purpose, or exceed what was authorized.\tabularnewline
Authority drift and delegation depth &
Goals, context and permissions are transformed as work passes through agents, sub-agents, tools, or external services. &
A downstream component may exercise authority beyond its delegated scope or act without recoverable provenance.\tabularnewline
Multi-agent coordination and collusion &
Agents may duplicate or contradict actions, contaminate shared memory, diffuse accountability, or produce harmful interaction patterns. &
System-level behavior may diverge from the mandate even when individual actions appear locally permissible.\tabularnewline
Learning and profile drift &
Memory, relevance profiles, skills, policies, or behavior-selection tendencies change through adaptation. &
Future behavior may gradually move outside the mandate or become difficult to reverse and audit.\tabularnewline
\midrule
\multicolumn{3}{l}{\emph{Cross-cutting, enters at any of the points above}}\tabularnewline
Security and resilience failure &
Compromise, degraded tools, policy conflict, anomalous delegation, or loss of recovery mechanisms impairs behavioral regulation. &
A local failure may propagate into a broader loss of bounded, observable and corrigible agency.\tabularnewline
\bottomrule
\end{tabularx}
\end{table}

\begin{figure}[t]
\centering
\resizebox{\textwidth}{!}{%
\begin{tikzpicture}[font=\footnotesize,
  pr/.style={draw=violet!60!black, fill=white, rounded corners=2pt, align=center, inner sep=4pt, minimum height=7mm},
  cond2/.style={draw=black!75, line width=0.8pt, dashed, rounded corners=1pt, fill=yellow!15, align=center, inner sep=3.5pt, minimum height=7mm},
  sub/.style={draw=black!75, line width=0.8pt, rounded corners=1pt, fill=yellow!15, align=center, inner sep=3pt},
  toolbox/.style={draw=black!60, fill=black!8, align=center, inner sep=3.5pt},
  rep2/.style={draw=black!60, rounded corners=1pt, fill=blue!7, align=center, inner sep=3.5pt, minimum height=7mm},
  app2/.style={draw=black!60, rounded corners=1pt, fill=orange!12, align=center, inner sep=3.5pt, minimum height=7mm},
  beh2/.style={draw=black!60, rounded corners=1pt, fill=green!9, align=center, inner sep=3.5pt, minimum height=7mm},
  chan2/.style={draw=black!60, rounded corners=1pt, fill=black!5, align=center, inner sep=3.5pt, minimum height=7mm},
  loss/.style={font=\scriptsize, text=red!55!black},
  lab/.style={font=\scriptsize},
  arr/.style={-{Stealth[length=2.2mm]}, semithick},
  farr/.style={-{Stealth[length=2.2mm]}, semithick, dashed, draw=black!60, rounded corners=3mm}]
\node[pr]    (P)  at (0.9,0)  {principal};
\node[cond2] (J)  at (3.6,0)  {mandate $J_{t}$};
\node[rep2]  (S)  at (7.6,0)  {intent $I_{t}$ $\cdot$ policy $\phi_{t}$ $\cdot$ profile $\rho_{t}$};
\node[app2]  (Pi) at (11.6,0) {selector $\Pi$};
\node[beh2]  (B)  at (14.0,0) {behavior $b_{t}$};
\node[chan2] (O)  at (16.4,0) {outcomes};
\draw[arr] (P) -- (J);
\draw[arr] (J) -- node[lab, above=4mm] {translation} node[loss, below=4mm] {$L^{mandate}$} (S);
\draw[arr] (S) -- (Pi);
\draw[arr] (Pi) -- (B);
\draw[arr] (B) -- (O);
\draw[dashed, rounded corners=3pt, draw=black!50] (8.4,-1.15) rectangle (17.45,-3.9);
\node[anchor=north east, font=\scriptsize\itshape, text=black!60] at (17.35,-1.22) {delegated work};
\node[sub] (A1) at (10.6,-2.15) {sub-agent\\ {\tiny narrowed sub-mandate $J'_{t}$}};
\node[sub] (A2) at (15.4,-2.15) {sub-agent\\ {\tiny sub-mandate $J''_{t}$}};
\node[toolbox] (T) at (15.4,-3.3) {tool / external service\\ {\tiny typed operation, no mandate}};
\draw[arr] (Pi.south) to[out=270,in=90] node[lab, left=1mm, pos=0.4] {sub-delegates} node[loss, right=1mm, pos=0.4] {$L^{authority}$} (A1.north);
\draw[{Stealth[length=2.2mm]}-{Stealth[length=2.2mm]}, semithick, dashed, draw=black!60] (A1) -- node[lab, above=5.5mm] {shared memory} node[loss, below=0.3mm] {$L^{coordination}$} (A2);
\draw[arr] (A2) -- (T);
\node[lab, anchor=west] at (8.65,-3.25) {$\mathcal{B}^{\phi}_{j,t}\subseteq\mathcal{B}^{delegated}_{i\rightarrow j,t}$ (Definition~\ref{def:delegation})};
\node[chan2] (Y) at (6.2,-4.5) {governed traces $y_{t}=\Gamma(S_{t},b_{t},r_{t},\epsilon^{obs}_{t})$};
\node[loss, anchor=north] at (6.2,-5.0) {$L^{information}$};
\draw[farr] (O.east) -- ++(0.55,0) |- (Y.east);
\draw[farr] (Y.west) -| node[lab, pos=0.85, right=0.5mm, align=left] {review $\cdot$ correction\\ revocation} (P.south);
\draw[farr] (Y.north -| S) -- node[lab, left=0.5mm, pos=0.4] {adaptation within $\mathcal{R}^{\phi}_{t}$} node[loss, left=0.5mm, pos=0.62] {$L^{adaptation}$} (S.south);
\end{tikzpicture}}
\caption{The delegation loop and its loss points.}
\label{fig:delegation-loop}
\end{figure}

\subsection{Reciprocal asymmetry and hidden behavior}\label{subsec:reciprocal-asymmetry}

Section~\ref{subsec:sensorium} introduced allo-interoception as the agent's indirect access to the principal's internal condition. This asymmetry is not one-sided. A corresponding limitation holds for the user. The user cannot directly access the agent's operational interoception, including its active commitments, uncertainty, intermediate plans, unresolved tasks, memory conflicts, tool failures, policy tensions, sub-agent activity, or the contextual factors that shaped a behavior. The user can only infer aspects of this condition from visible actions, tool logs, status signals, explanations and outcomes.

In this sense, logs, traces and explanations play for the user a role similar to the one behavioral, linguistic, or physiological traces play for the agent. They make behavior more traceable, but they do not provide direct access to the internal state through which behavior was formed. A trace may show that a tool was called, a message was sent, or a memory was retrieved, while leaving unclear which commitments, uncertainties, policies, memories, or contextual interpretations made that action seem appropriate.

This reciprocal asymmetry matters because artificial agents make the observability problem sharper. Their behavior may depend on latent computation, stochastic outputs, changing context, tool-mediated actions, delegated subtasks and coordination across several agents. At the same time, digital agents can preserve detailed operational records, including tool calls, memory updates, policy checks, delegated tasks, messages, state changes, explanations and outcomes. The challenge is to transform these records into governed evidence about what the agent did, under which authority it acted, what uncertainty was present, which constraints were applied and what outcome followed.

Mechanistic interpretability may reduce part of this asymmetry by adding evidence about the internal computation of learned systems \cite{Olah2020Circuits,Elhage2022Superposition,Bricken2023Monosemanticity,Templeton2024Monosemanticity}. However, it should not be treated as complete access to the agent's operational interoception. It complements operational provenance, logs, explanations, policy records and external governance, rather than replacing them.

\subsection{Governance, security and resilience as behavioral regulation}\label{subsec:governance-theory}

Delegated agency must preserve not only task performance, but also the conditions under which behavior remains reliable, bounded, observable and recoverable. Governance names the mechanisms that maintain these conditions throughout the life of a delegation. It specifies and records the mandate, bounds authority, makes consequential behavior observable, detects divergence and permits correction, revocation and accountability. It must constrain not only present behavior, but also the authorized process through which future behavior is produced. Learning makes this necessary. The agent's profiles, memories, skills and patterns of intervention may change after delegation, so the delegator is exposed not only to what the agent does, but to what it may become. Governance is therefore not external administration added after an agent acts. It is a constitutive condition of reliable delegated agency.

Part of this regulation can occur inside the agent itself. An agent should regulate its behavior when its own operating condition is unsafe or unreliable. The agent-interoceptive channel of Section~\ref{subsec:sensorium} can represent uncertainty, degraded tools, anomalous delegation requests, policy conflict, compromised memory, excessive authority requests and coordination failure. When such conditions are appraised as significant, the appropriate behavior may be to gather further evidence, reduce the scope of action, ask for clarification, defer, escalate, or stop. We call this \textbf{agent-interoceptive security}: the use of monitored operational condition to preserve reliable future behavior. It requires no machinery beyond what the theory already provides. It is affective-conative inference applied to the agent's own condition, with restraint among its outputs.

External actions alone may not reveal whether these conditions are present. An agent may appear to follow an authorized task while unsafe instructions, retrieved content, unresolved uncertainty, or incomplete constraint processing shape its behavior selection. This is one reason the appraisal layer is itself an attack surface (Section~\ref{sec:open-problems}).

Behavioral regulation inside the agent is necessary but insufficient. An agent that is compromised, misconfigured, or drifting cannot be the sole authority over its own limits. External controls must therefore remain non-bypassable, including permission boundaries, action firewalls, mandate validation, provenance records, approval gates and revocation mechanisms. The internal system may recognize and respond to risk. The external system determines what it remains permitted to do. Section~\ref{sec:design} states this as the two non-learned enforcement points.

Finally, security and resilience are expressed through behavior across time, not only through the prevention of a single harmful action. Under uncertainty, conflict, or degraded capability, a well-governed agent should degrade gracefully. It should narrow its authority, preserve evidence, avoid irreversible action, seek supervision where required and retain a path to recovery. Graceful degradation is, in this sense, the security expression of restraint. These properties reduce the risk that a local failure becomes a broader loss of bounded, observable and corrigible agency. This is why Table~\ref{tab:agency-loss-problems} lists security and resilience failure as cross-cutting.

\subsection{Summary: the elements of agency}\label{subsec:elements}

Agency can be defined as \emph{the governed organization of behavior across time by perception, intent, affective-conative inference, constraint and feedback, realized by intelligence}. For symbiotic agency, which is the focus of this work, one further element is required which is the mandate. Symbiotic agency is a form of delegated agency. The mandate specifies the delegated authority to which the organization of behavior must remain answerable. It is not required for agency in general. A fully autonomous system may organize behavior under self-generated goals, internal constraints and external norms. However, it is required for \emph{delegated} agency, because delegation is the transfer of a bounded capacity to perceive, decide and act on another's behalf (Section~\ref{subsec:delegated-theory}). Where no mandate exists, constraint may still govern behavior, but answerability has no clear anchor. This is why unmandated proactive agents are a governance risk and why symbiosis without a mandate becomes surveillance rather than augmentation.

The table below expands this definition into the core elements of the account. Representation, memory and profile are maintained structures through which the organizing capacities operate. They support agency, but they do not by themselves constitute it.

\begin{center}
\begin{tabular}{>{\raggedright}p{0.22\linewidth}>{\raggedright}p{0.68\linewidth}}
\toprule
\textbf{Element} & \textbf{Role in agency}\tabularnewline
\midrule
Perception & Supplies policy-governed evidence through exteroceptive, allo-interoceptive and agent-interoceptive channels.\tabularnewline
Intelligence & Provides representation, inference, reasoning, prediction, planning and model-based evaluation.\tabularnewline
Representation & Maintains structured internal state over what is perceived, inferred, remembered and expected.\tabularnewline
Memory & Preserves episodes, corrections, commitments, preferences and learned significance patterns as evidence for future interpretation and behavior.\tabularnewline
Intent & Organizes behavior around commitments, goals, norms, tasks and maintained states.\tabularnewline
Affective-conative profile & Shapes relevance and response through generic, personalized and evolvable components. Revisable only within authorized bounds.\tabularnewline
Affective-conative inference & Evaluates what matters, why, how much and what orientation toward behavior is warranted. Readiness is tested against the context-sensitive activation gate and priority chooses among the commitments that clear it.\tabularnewline
Constraint & Bounds perception, inference, memory and action by permissions, policies, norms and user boundaries.\tabularnewline
Mandate & For symbiotic agency: the authorizing goal, scope, boundaries and responsibilities under which the other elements operate, operationalized through intent, policy and profile.\tabularnewline
Behavior & The observable expression of agency: action, communication, tool use, waiting, asking, escalating, inhibition, restraint.\tabularnewline
Feedback & Updates representations, commitments, profiles, policies and future behavior from outcomes and corrections, within the bounds of authorized adaptation.\tabularnewline
\bottomrule
\end{tabular}
\par\end{center}

The inventory is exact rather than illustrative. Each element corresponds to an object in the formal vocabulary: perception to $o_{t}$, representation to $K_{t}$, memory to $M_{t}$, intent to $I_{t}$, affective-conative inference to $C_{t}$, profile to $\rho_{t}$, constraint to $\phi_{t}$, mandate to $J_{t}$, behavior to $b_{t}$ and feedback to $r_{t}$. Intelligence is realized by the operators that compute, update and coordinate these objects. Agency loss is not included as an element because it is not something the agent maintains. It is an evaluation over trajectories, measuring how far organized behavior departs from the mandate (Section~\ref{subsec:delegated-theory}).

\begin{figure}[t]
\centering
\resizebox{\textwidth}{!}{%
\begin{tikzpicture}[
  font=\footnotesize,
  node distance=3.5mm and 6mm,
  chan/.style={draw=black!60, rounded corners=1pt, fill=black!5, minimum width=27mm, minimum height=6mm, align=center, inner sep=2pt},
  rep/.style={draw=black!60, rounded corners=1pt, fill=blue!7, minimum width=23mm, minimum height=6mm, align=center, inner sep=2pt},
  coup/.style={draw=violet!60!black, rounded corners=1pt, fill=violet!12, minimum width=23mm, minimum height=6mm, align=center, inner sep=2pt},
  app/.style={draw=black!60, rounded corners=1pt, fill=orange!12, align=center, inner sep=3pt},
  beh/.style={draw=black!60, rounded corners=1pt, fill=green!9, align=center, inner sep=3pt},
  hard/.style={draw=red!55!black, line width=0.9pt, fill=red!15, minimum width=2.6mm, inner sep=0pt},
  cond/.style={draw=black!60, dashed, rounded corners=1pt, fill=yellow!15, align=center, inner sep=2.5pt},
  arr/.style={-{Stealth[length=2.2mm]}, semithick},
  farr/.style={-{Stealth[length=2.2mm]}, semithick, dashed, draw=black!60}
]
\node[chan] (ext) {exteroception $o^{ext}_{t}$};
\node[coup, minimum width=27mm, below=of ext] (allo) {allo-interoception $o^{allo}_{t}$};
\node[chan, below=of allo] (self) {agent-interoception $o^{self}_{t}$};
\node[font=\footnotesize\itshape] at ([yshift=4mm]ext.north) {sensorium};
\node[hard, minimum height=32mm, right=5mm of allo] (cgate) {};
\node[rotate=90, font=\scriptsize, text=red!55!black] at (cgate.center) {consent gate};
\node[coup, right=6mm of cgate] (U) {principal $U_{t}$ (calibrated)\\ {\tiny symbiotic coupling}};
\node[rep, above=of U] (X) {world $X_{t}$};
\node[rep, below=of U] (Z) {self $Z_{t}$};
\node[cond, below=4.5mm of Z] (mem) {memory $M_{t}$ $\cdot$ intent $I_{t}$ $\cdot$ profile $\rho_{t}$};
\node[app, right=7mm of U] (appr) {appraisal $\sigma_{t}$\\ readiness $\lambda_{i,t}$};
\node[app, right=6mm of appr] (agate) {activation gate $\lambda_{i,t}>\vartheta_{t}$\\ priority $q_{t}$ among the eligible};
\node[beh, right=15mm of agate] (sel) {behavior selection\\ \mbox{$b^{*}_{t}\in\mathcal{B}^{\phi}_{t}(i^{*}_{t})$}};
\node[beh, below=8mm of sel] (rest) {restraint\\ \mbox{wait $\cdot$ monitor $\cdot$ defer $\cdot$ inhibit $\cdot$ no-op}};
\node[chan, below=8mm of agate, xshift=-7mm, minimum width=0mm, text width=30mm] (standing) {standing operation\\ {\tiny sensorium open $\cdot$ representations update $\cdot$ commitments dormant}};
\node[hard, minimum height=20mm, right=5mm of sel] (fgate) {};
\node[rotate=90, font=\scriptsize, text=red!55!black] at (fgate.center) {action firewall};
\node[chan, right=5mm of fgate] (world) {external effects\\ world $\cdot$ principal $\cdot$ tools};
\node[cond, above=8mm of agate] (mand) {mandate $J_{t}\ \rightarrow\ $ intent $I_{t}$, policy $\phi_{t}$, profile $\rho_{t}$\\ {\tiny delegated tier ($J_{t}\neq\emptyset$)}};
\draw[arr] (ext.east) -- (ext.east -| cgate.west);
\draw[arr] (allo.east) -- (allo.east -| cgate.west);
\draw[arr] (self.east) -- (self.east -| cgate.west);
\draw[arr] (X.west -| cgate.east) -- (X.west);
\draw[arr] (U.west -| cgate.east) -- (U.west);
\draw[arr] (Z.west -| cgate.east) -- (Z.west);
\draw[arr] (X.east) to[out=0,in=155] (appr.west);
\draw[arr] (U.east) -- (appr.west);
\draw[arr] (Z.east) to[out=0,in=205] (appr.west);
\draw[farr] (mem.east) to[out=0,in=250] (appr.south);
\draw[arr] (appr.east) -- (agate.west);
\draw[arr] (agate.east) -- node[above, font=\scriptsize, align=center, inner sep=1pt] {winner\\ \mbox{$i^{*}_{t}$}} (sel.west);
\draw[arr] (agate.south) -- node[fill=white, inner sep=1pt, font=\scriptsize] {not eligible} (standing.north);
\draw[farr] (standing.west) to[out=185,in=300] (appr.south);
\draw[arr] (sel.south) -- (rest.north);
\draw[arr] (sel.east) -- (sel.east -| fgate.west);
\draw[arr] (world.west -| fgate.east) -- (world.west);
\draw[farr] (mand.south west) -- (appr.north);
\draw[farr] (mand.south -| agate.north) -- (agate.north);
\draw[farr] (mand.south east) -- (sel.north);
\draw[farr] (rest.south) -- ++(0,-1.25) -| node[pos=0.25, above, font=\scriptsize] {internal effects} (mem.south);
\draw[farr] (world.south) -- ++(0,-3.7) -| node[pos=0.25, above, font=\scriptsize] {feedback $r_{t}$: outcomes and corrections} (self.south);
\end{tikzpicture}}
\caption{The activation pipeline.}
\label{fig:activation-pipeline}
\end{figure}
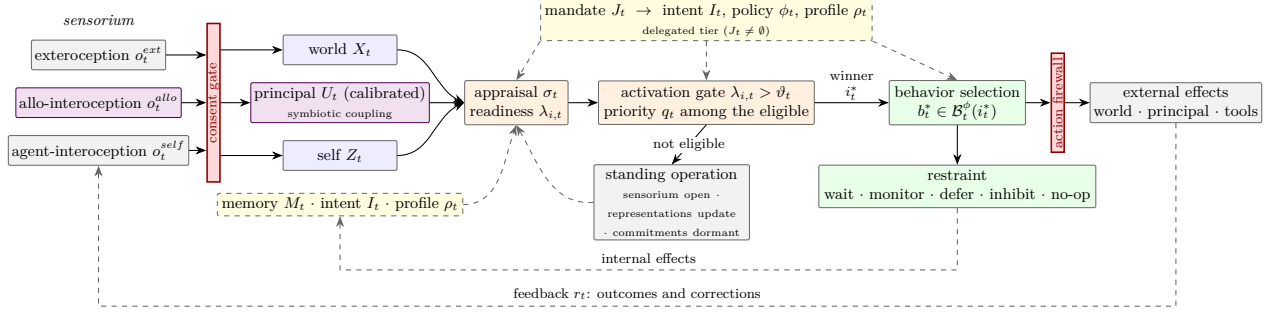

Figure~\ref{fig:activation-pipeline} summarizes how these elements organize the flow from perception to behavior. Policy-gated perception admits exteroceptive, allo-interoceptive and agent-interoceptive evidence through the consent gate. These inputs update representations of world state $X_{t}$, user state $U_{t}$ and agent state $Z_{t}$, which then feed affective-conative appraisal and the readiness of each pressing commitment. Readiness is tested against a context-sensitive activation gate and priority chooses among the commitments that clear it. Clearing the bar opens a behavioral episode in which behavior is selected from the permitted set $\mathcal{B}^{\phi}_{t}$. Below the bar, the agent remains in standing operation, perceiving and updating under policy with no episode open. Restraint remains selectable even when external action is technically possible and external effects must pass through the action firewall.

The figure also shows where the delegated tier enters the general pipeline. The mandate $J_{t}$ conditions intent, policy and profile. If this mandate conditioning is removed, the same pipeline describes autonomous agency. In that case, principal-directed allo-interoception is absent. The system may still model other agents, users, or counterparties, but those signals are treated as part of the world state $X_{t}$  rather than as evidence about a principal's condition $U_{t}$ to which behavior is answerable. The inference may still be difficult and theory-of-mind-like, but it does not carry the same calibration, consent and mandate obligations that attach to symbiotic agency.

Within delegated agency, symbiotic agency is distinguished by the coupling apparatus that connects the agent to the principal over time. This apparatus includes the allo-interoceptive channel, the calibrated estimate $U_{t}$ and the personalized component of $\rho_{t}$. A task-bounded delegated agent may run the same pipeline with these coupling elements absent or minimal while still retaining the delegated tier. Its mandate still conditions intent, policy and profile and its behavior remains answerable to that mandate. Its exteroception may be limited to the task environment and its agent-interoception may be reduced to basic operational checks such as tool success, failure, or uncertainty. What distinguishes the forms of agency is therefore not a different pipeline, but which channels are populated, whose significance is computed and what the resulting behavior remains answerable to.

\section{A Formal Vocabulary for Governed, Proactive Agency}\label{sec:formal}

This section introduces the theory for governed, proactive agency as a set of typed definitions. Figure~\ref{fig:activation-pipeline} shows how these definitions are organized to form a governed loop. Policy-gated perception updates world, principal and agent representations, which feed appraisal, readiness, the activation gate, priority among the eligible commitments, behavior selection and feedback. The labeled vertical bars (red, thick-bordered) mark the two non-learned enforcement points, the consent gate and the action firewall. Dashed arrows indicate conditioning and feedback.

Before introducing the definitions, three clarifications are needed about how the formal vocabulary should be read.

\begin{enumerate}
\item
The vocabulary specifies the structure of agency, not a single implementation. The claim is that perception, representation, memory, intent, affective-conative inference, profile, constraint, mandate, behavior and feedback should be represented and governed distinctly. The vocabulary is agnostic about the computational substrate that realizes these elements (Section~\ref{subsec:related-formal}).
\item
When the definitions say that behavior is sensitive to a factor, this means that changing that factor may change the behavior selected by the agent. The claim is about the role the factor plays in the behavior, not about a particular mathematical form.
\item
The vocabulary is tiered. Definitions~\ref{def:behavior} to \ref{def:classes} state the general theory and apply to any system that organizes behavior across time. Definition~\ref{def:agenthood} closes this tier by stating the minimum under which such a system is an agent. The general tier must describe autonomous and delegated agents with one vocabulary. Its definitions are therefore written for the fuller case, in which a principal exists and the objects that refer to the principal are empty or zero when no operative mandate exists, $J_{t}=\emptyset$. Particularly this applies to the latent condition $u_{t}$, the allo-interoceptive channels and observations, the estimate $U_{t}$ and its calibration factors $\eta_{t}$, the personalized profile component $\rho^{user}_{t}$, the interruption-cost component $\sigma^{intr}$ and the trust-boundary part of $\sigma^{perm}$. Permission proximity to the agent's own constraint boundaries remains defined in every form. The general tier therefore leaves room for delegation without presupposing it. The forms then classify agents by the respective authority relationship. Definitions~\ref{def:autonomous} and~\ref{def:delegated} separate autonomous from delegated agency by the mandate (Definition~\ref{def:mandate}), Definitions~\ref{def:agency-loss} to \ref{def:bounded-adaptation} add the governance structures that delegation requires, Definition~\ref{def:symbiotic} states the further conditions that make a delegated agent symbiotic and Definition~\ref{def:aligned} is evaluative rather than constitutive.
\end{enumerate}

\subsection{Behavior, episodes and trajectories}

\begin{defn}[Behavior, episode and trajectory]\label{def:behavior}
The agent's behavior at time $t$ is $b_{t}\in\mathcal{B}_{t}$, where $\mathcal{B}_{t}$ is the set of behaviors available at $t$.  Externally visible actions form a subset $\mathcal{A}_{t}\subseteq\mathcal{B}_{t}$. Behavior is therefore broader than external action. It includes communication, tool use, inquiry, internal state updates, waiting, inhibition and deliberate restraint. A behavioral trajectory is $B_{t:t+k}=(b_{t},\ldots,b_{t+k})$. A \emph{behavioral episode} is a contiguous segment of this trajectory,  $e=B_{t_{0}:t_{1}}$, treated as one organized unit. Its boundary is not an arbitrary time window. It begins when an intention becomes active, either through a command or proactive activation, which sets the intention the episode serves (Definitions~\ref{def:gate} and~\ref{def:command}) and it ends when the intention is satisfied, suspended, returned to dormancy, or inhibited. The behaviors  $b_{t}$  within the episode are its micro-operations, although at a coarser scale the whole episode may itself be treated as a single behavior. An episode may contain no externally visible action. Sustained restraint with monitoring is still a behavioral episode. The information available to the agent up to time $t$ is the history $\mathcal{H}_{t}=(o_{0},(\lambda_{0},\vartheta_{0}),b_{0},r_{0},\ldots,o_{t},(\lambda_{t},\vartheta_{t}))$: what the agent perceived, how each step's gate evaluated (Definition~\ref{def:gate}), what it did and what resulted. At steps where no commitment's readiness cleared the bar,  no behavior is selected and no outcome follows, so those steps enter the record as $(o_{t},(\lambda_{t},\vartheta_{t}))$ alone.
\end{defn}

A trajectory is agentic to the extent that it is not merely a sequence of outputs but a structured pattern organized by perception, intent, affective-conative appraisal, constraint and feedback:
\begin{center}
\emph{$B_{t:t+k}$ is agentic if it is directed, situated, constrained and revisable.}
\end{center}
Definition~\ref{def:agenthood} states the minimum this criterion imposes on a system and Sections~\ref{sec:signatures} and~\ref{sec:measurement} operationalize it. Above the minimum, agency is graded, not binary.

\subsection{Perception as governed evidence acquisition}

\begin{defn}[Latent situation and sensorium]\label{def:sensorium}
The latent situation at time $t$ is $\xi_{t}=(x_{t},u_{t},z_{t})$, where $x_{t}$ is the external world state, $u_{t}$ is the principal's latent internal condition and $z_{t}$ the agent's own internal condition. The agent's sensorium is $\mathcal{W}_{t}=\mathcal{W}^{ext}_{t}\cup\mathcal{W}^{allo}_{t}\cup\mathcal{W}^{self}_{t}$, its window onto the latent situation. These are, respectively, channels about the external world, external traces used to infer the principal's internal condition and channels about the agent's own condition.
\end{defn}

\begin{defn}[Policy-gated observation]\label{def:observation}
Policy governs perception itself. At time \(t\), the allowed sensorium is $\mathcal{W}^{allowed}_{t}=\{w\in\mathcal{W}_{t}\mid\phi^{\mathcal{W}}_{t}(w)=1\}$, the set of sensing channels that the current policy permits the agent to use. The predicate $\phi^{\mathcal{W}}_{t}$ is the channel part of the constraint structure $\phi_{t}$ (Definition~\ref{def:constraint}). Perception maps the latent situation into observations through those permitted channels only:
\begin{equation}
o_{t}=\Omega_{\phi_{t}}(\xi_{t},b_{t-1},r_{t-1},\epsilon_{t})=(o^{ext}_{t},o^{allo}_{t},o^{self}_{t}),
\label{eq:policy-gated-observation}
\end{equation}
where $\epsilon_{t}$ captures noise, incompleteness, latency and distortion. The operator $\Omega_{\phi_{t}}$ is the perception operator. It takes the latent situation, the preceding behavior and outcome and the noise term and produces channel-tagged observations. Where the preceding step selected no behavior, the $b_{t-1}$ and $r_{t-1}$ arguments are empty. The subscript $\phi_{t}$ indicates that perception is policy-gated: the current policy determines which sensing channels the operator may use. As with the other operators in the vocabulary, the definition specifies what the operator receives and what it produces, while leaving the computational implementation open. Table~\ref{tab:operators} collects the full family of operators. Perception is therefore access under constraint. The agent never acts on the latent situation as such. It acts on a policy-filtered and imperfect observation of it.
\end{defn}

\begin{defn}[Allo-interoceptive pipeline]\label{def:allo}
The principal's internal condition $u_{t}$ is latent relative to the agent. It influences observable traces $h^{u}_{t}\sim P(\cdot\mid u_{t},x_{t},b_{t-1})$, which are sensed (subject to policy) as part of $o^{allo}_{t}$ and summarized into the agent's estimate $U_{t}$:
\begin{equation}
u_{t}\longrightarrow h^{u}_{t}\longrightarrow o^{allo}_{t}\longrightarrow U_{t}.
\label{eq:allo-pipeline}
\end{equation}
Allo-interoception is structurally more underdetermined than the other perceptual channels. The same trace can be compatible with multiple internal conditions of the principal and the mapping depends on context. $U_{t}$ should therefore be distributional and carry calibrated uncertainty. The channel is principal-directed. Only $u_{t}$, the principal's condition, is distinguished in the latent situation. Inference about other agents who are not the principal belongs to exteroceptive world modeling within $X_{t}$. In autonomous agency, this channel is therefore empty. In task-bounded delegation, for the most part it is absent or otherwise minimally populated.
\end{defn}

\subsection{The agentic state and its representations}

\begin{defn}[Agentic state]\label{def:agentic-state}
At each time $t$ the agent maintains
\begin{equation}
S_{t}=(K_{t},I_{t},M_{t},\phi_{t},\rho_{t}),
\label{eq:agentic-state}
\end{equation}
where $K_{t}$ is the situated representation, $I_{t}$ the active and dormant intentions, $M_{t}$ memory, $\phi_{t}$ the constraint structure (Definition~\ref{def:constraint}) and $\rho_{t}$ the affective-conative profile. The state is updated by perception and feedback, together with the learned parameters $\theta_{t}$ of the operators that realize it. The update is $(S_{t+1},\theta_{t+1})=\mathcal{U}(S_{t},\theta_{t},o_{t+1},b_{t},r_{t})$. $\mathcal{U}$ is the update operator. It takes the current state and parameters, the new observation, the behavior taken and its outcome and returns the next state and parameters. Each part of the state is updated by its own rule. The representation $K_{t+1}$ is formed by $F_{X}$, $F_{U}$, $F_{Z}$ and $\Psi$ (Definitions~\ref{def:core-representation} and~\ref{def:appraisal}), the intentions $I_{t+1}$ and memory $M_{t+1}$ from the step's observation, behavior and outcome, the profile component $\rho^{evol}_{t+1}$ by $\mathcal{L}_{\rho}$ (Definition~\ref{def:profile}), the parameters $\theta_{t}$ by learning and the constraints $\phi_{t}$ only by explicit approval (Definition~\ref{def:bounded-adaptation}). Definition~\ref{def:core-representation} writes the representation from the whole observation history and $\mathcal{U}$ writes it one step at a time. Both must give the same representation, so the vocabulary defines it only once. Definition~\ref{def:bounded-adaptation} bounds $\mathcal{U}$ wherever a mandate is present.
\end{defn}

\begin{defn}[Core representation]\label{def:core-representation}
\begin{equation}
K_{t}=(X_{t},U_{t},Z_{t},C_{t}),
\label{eq:core-representation}
\end{equation}
where $X_{t}=F_{X}(o^{ext}_{0:t},M_{t})$  is the exteroceptive representation of the world, $U_{t}=F_{U}(o^{allo}_{0:t},X_{t},M_{t},\eta_{t})$ is the allo-interoceptive estimate of the principal's internal condition, $Z_{t}=F_{Z}(o^{self}_{0:t},I_{t},\phi_{t},\rho_{t},M_{t},r_{0:t-1})$ is the agent-interoceptive representation of the agent's own condition and $C_{t}$ is the affective-conative appraisal state defined below in Definition~\ref{def:appraisal}. The latent variables $x_{t}$, $u_{t}$ and $z_{t}$ are ontological. They describe what is the case. The representations $X_{t}$, $U_{t}$ and $Z_{t}$ are epistemic. They describe what the agent has inferred from available evidence.
The maps $F_{X}$, $F_{U}$ and $F_{Z}$  are representation-forming operators in the operator family summarized in Table~\ref{tab:operators}. Each operator is specified by its signature, meaning by the evidence it may use to form its representation. How the representation is computed is left to the implementation. The signatures also carry governance. Each operator consumes only the policy-permitted observation stream for its channel, together with the declared context. Representation formation therefore inherits the constraints on perception stated in Definition~\ref{def:observation}.
The differences between the operators are deliberate. $F_{U}$ alone takes the calibration factors $\eta_{t}$, because allo-interoception estimates the principal's internal condition from indirect traces. These factors encode the maintained, principal-specific mapping from observable evidence to inferred condition and support the requirement that $U_{t}$ remain distributional and calibrated. $F_{Z}$ alone takes the outcome history $r_{0:t-1}$, because the agent's representation of its own condition must include evidence about what its previous behavior produced.
\end{defn}

\begin{defn}[Affective-conative profile]\label{def:profile}

The profile $\rho_{t}=(\rho^{gen},\rho^{user}_{t},\rho^{evol}_{t})$ contains three components. A generic prior component for what tends to matter in general, a personalized component modeling what this user values, fears, tolerates, delegates, protects and finds intrusive and an evolvable component carrying the agent's revisable habits of noticing, interpreting and responding (Section~\ref{subsec:profiles-theory}). The profile conditions the appraisal-side operators. In the definitions below, $\Sigma$, $Q$, $\Lambda$, $G_{\vartheta}$, $A_{i}$ and $V$  all take $\rho_{t}$ as a parameter, so the same evidence may produce different significance, priority, readiness and behavior under different profiles.

Rules, permissions and policies are not profile components. They enter through $\phi_{t}$ and bound what any profile may influence. Evolvable revision is also bounded:
\begin{equation*}
\rho^{evol}_{t+1}=\mathcal{L}_{\rho}(\rho^{evol}_{t},M_{t},r_{t},\mathrm{corrections}_{t})\in\mathcal{R}^{\phi}_{t},
\end{equation*}
where $\mathcal{L}_{\rho}$ is the profile-revision operator. It produces the next evolvable component from the current one in light of memory, outcomes and corrections. The set $\mathcal{R}^{\phi}_{t}$ contains the profile updates admissible under consent, policy and mandate, as a special case of Definition~\ref{def:bounded-adaptation}. The personalized component is not revised by a separate learning rule. The time index on $\rho^{user}_{t}$ marks the effective personalized mapping available at time $t$, not an unconstrained update channel. That effective mapping consists of the principal-authorized personalized base together with the personalization-directed revisions carried by $\rho^{evol}_{t}$. Learning therefore changes personalization only through $\mathcal{L}_{\rho}$, by updating $\rho^{evol}_{t}$ and every such update must lie in $\mathcal{R}^{\phi}_{t}$. The authorized base mapping changes only through an explicit act of the principal, as mandate content does, not through ordinary adaptation.

The role of the profile differs across the forms of agency. An autonomous agent instantiates $\rho^{gen}$ and $\rho^{evol}_{t}$, but it has no principal for $\rho^{user}_{t}$ to serve. The personalized component belongs to symbiotic delegated agency, while the consent-bounded region $\mathcal{R}^{\phi}_{t}$ belongs to the delegated tier more generally. The substrate of revision differs accordingly. In symbiotic agency, $\rho^{evol}_{t}$ carries revisions to personalized mappings within $\mathcal{R}^{\phi}_{t}$. In autonomous agency, it carries revisions to generic mappings supplied by $\rho^{gen}$, subject to external constraints.
\end{defn}

\begin{defn}[Constraint structure]\label{def:constraint}
The constraint structure $\phi_{t}$ records what the agent may perceive, do and change. It has three parts. A channel predicate $\phi^{\mathcal{W}}_{t}$ sets which sensing channels the agent may use (Definition~\ref{def:observation}). A behavior predicate $\phi^{\mathcal{B}}_{t}$ sets which behaviors it may select in the current situation (Definition~\ref{def:selection}). Adaptation bounds specify what learning may change and what requires explicit approval and they define the admissible regions $\mathcal{U}^{\phi}_{t}$ and $\mathcal{R}^{\phi}_{t}$ (Definitions~\ref{def:profile} and~\ref{def:bounded-adaptation}). Its content comes from the mandate for a delegated agent, from external norms and the agent's own commitments for an autonomous one and from every party whose hard boundaries bind the agent (Definitions~\ref{def:mandate}, \ref{def:autonomous} and~\ref{def:delegation}). It holds hard constraints, not preferences, which enter through the profile and the situated value. Its predicates are enforced at the two non-learned enforcement points, the consent gate and the action firewall (Section~\ref{sec:design}) and the structure itself changes only through explicit approval (Definition~\ref{def:bounded-adaptation}). An operator that takes $\phi_{t}$ may read any of its parts.
\end{defn}

\subsection{Intent}

\begin{defn}[Intention]\label{def:intention}
Let $D_{t}$ be the set of possible desires, goals, norms, tasks and maintained conditions at time $t$. Intentions are the subset to which the agent is committed, $I_{t}\subseteq D_{t}$. An intention is a commitment structure
\begin{equation}
i=(g_{i},p_{i},\kappa_{i},\tau_{i},\mathcal{B}_{i}),
\label{eq:intention-structure}
\end{equation}
comprising a target or maintained condition $g_{i}$, which specifies what the intention is trying to bring about, preserve, avoid, or respect. A priority $p_{i}$. A persistence condition $\kappa_{i}$, which specifies when the intention should remain in force. A termination condition $\tau_{i}$ and a set of permissible behavior schemas $\mathcal{B}_{i}$. An intention may be dormant ($\alpha_{i}(t)=0$) while still being part of the agent’s standing commitments. It is pressing when its activation is positive, that is, when the situation calls for it but no episode serves it yet and active while an open episode serves it (Definition~\ref{def:gate}). Its activation level is
\begin{equation}
\alpha_{i}(t)=A_{i}(X_{t},U_{t},Z_{t},C_{t},M_{t},\phi_{t};\rho_{t}).
\label{eq:intention-activation}
\end{equation}
$A_{i}$ is the activation operator of intention $i$. From the current representations, the appraisal state $C_{t}$ (the situation's significance for behavior, formalized below in Definition~\ref{def:appraisal}), memory and policy, under the profile, it produces the degree to which a standing commitment presses to become behavior now. The coupling with appraisal is mutual. The standing intention set $I_{t}$ enters the appraisal operator $\Psi$ (Definition~\ref{def:appraisal}), so significance is computed in relation to what the agent is committed to. The resulting appraisal state $C_{t}$ then enters each $A_{i}$, so the timing of each commitment is computed in relation to that significance. Commitments make significance interpretable and significance makes commitments timely. There is no circularity in this coupling. $I_{t}$ is part of the carried state $S_{t}$ (Definition~\ref{def:agentic-state}). The set of commitment structures that persists across steps. Each commitment in it carries the target or maintained condition $g_{i}$, priority $p_{i}$, persistence condition $\kappa_{i}$, termination condition $\tau_{i}$ and permissible behavior schemas $\mathcal{B}_{i}$, but not the current activation level. The activation $\alpha_{i}(t)$ is computed fresh within the step, after $C_{t}$ has been formed. Within a step, standing intentions shape appraisal and appraisal shapes activation. Several commitments may call for behavior at the same time, each with a positive activation level. The gate tests the readiness of each and priority picks among those that clear it (Definition~\ref{def:gate}).
\end{defn}

Intent provides the agent with a stable organizational structure, defining what it remains committed to and which classes of behavior are appropriate. Affective-conative inference plays a different role by determining the situated relevance of those commitments, namely whether a given intention should become active under the present conditions. Maintaining this distinction avoids two complementary failure modes. The first is inert intent, where commitments exist but rarely influence behavior because perception is not mapped onto relevance. The second is distracted relevance, where signals of significance trigger action without being anchored in durable commitments. The mutual signature reference of Definitions~\ref{def:intention} and~\ref{def:appraisal} rules both out by construction. Remove $I_{t}$ from $\Psi$ and significance loses its anchor in commitment. Remove $C_{t}$ from $A_{i}$ and commitment loses its path from perception to timing.

\subsection{Appraisal, priority, readiness and the activation gate}

\begin{defn}[Affective-conative appraisal]\label{def:appraisal}
The appraisal state evaluates the significance of the current situation for behavior:
\begin{equation}
C_{t}=\Psi(X_{t},U_{t},Z_{t},I_{t},M_{t},\phi_{t};\rho_{t}).
\label{eq:affective-conative-appraisal}
\end{equation}
Within it, the agent computes an appraisal vector
\begin{equation}
\sigma_{t}=\Sigma(X_{t},U_{t},Z_{t},I_{t},M_{t},\phi_{t},\widehat{\Delta}_{t};\rho_{t}),\qquad\sigma_{t}\in\mathbb{R}^{m},
\label{eq:appraisal-vector}
\end{equation}
where $\widehat{\Delta}_{t}(b)=\mathbb{E}[\Delta\mid X_{t},U_{t},Z_{t},I_{t},M_{t},b]$ is the expected situated consequence of a candidate behavior, the change it would produce in the latent situation $\xi_{t}$ and in the standing of the agent's commitments. The consequence $\Delta$ is a random variable because outcomes are uncertain, so the appraisal uses its expectation under the current representations. The candidates are the behaviors available at $t$, $\mathcal{B}_{t}$ (Definition~\ref{def:behavior}), including the schemas of the standing intentions and modes such as asking, waiting, monitoring and refraining. Appraisal estimates their consequences before any choice is made. Policy then fixes the permitted subset $\mathcal{B}^{\phi}_{t}$ and the selector of Definition~\ref{def:selection} chooses among those. $\Psi$ and $\Sigma$ are the appraisal operators. $\Psi$ produces the appraisal state $C_{t}$ from the representations, the intentions, memory and policy, under the profile. $\Sigma$ distills the same evidence, augmented by $\widehat{\Delta}_{t}$, into the vector of named significance components. The two are nested rather than parallel. The appraisal vector $\sigma_{t}$ and the consequence estimates $\widehat{\Delta}_{t}$ are components of the appraisal state $C_{t}$, which may carry further appraisal structure that an implementation maintains, so computing $C_{t}$ includes computing $\sigma_{t}$ and $\sigma_{t}$ is the named readout of $C_{t}$. The activation operators $A_{i}$ read the whole state, because an intention's activation depends on the consequence estimates for the behaviors in its own schemas $\mathcal{B}_{i}$. The gate-side operators $Q$, $\Lambda$ and $G_{\vartheta}$ read only $\sigma_{t}$ from it. Priority, readiness and the bar are not components of $C_{t}$. They are computed from it afterwards, in the within-step order of Definition~\ref{def:intention}. Gate-relevant components include urgency $\sigma^{urg}$, potential harm from inaction $\sigma^{harm}$, action risk and irreversibility $\sigma^{risk}$, social and emotional sensitivity $\sigma^{sens}$, interruption cost $\sigma^{intr}$, permission and trust-boundary proximity $\sigma^{perm}$ and uncertainty $\sigma^{unc}$. Components of $Z_{t}$ function as the affect-like control signals introduced in Sections~\ref{subsec:sensorium} and~\ref{subsec:profiles-theory}. Signals such as low confidence, tool failure, policy tension, or unresolved commitment enter $\Psi$, $\Lambda$ and $G_{\vartheta}$ and thereby change how the agent attends, asks, escalates, defers, or refrains.
\end{defn}

\begin{defn}[Priority, readiness and gate]\label{def:gate}
Behavior-readiness is computed for every pressing commitment. For intention $i$ it is the graded disposition of that commitment to surface now, $\lambda_{i,t}=\Lambda(\alpha_{i}(t),\sigma_{t},Z_{t},\phi_{t};\rho_{t})$ and $\lambda_{t}$ denotes the vector of these values. A commitment becomes eligible only when its readiness exceeds a context-sensitive threshold,
\begin{equation}
\lambda_{i,t}>\vartheta_{t},\qquad \vartheta_{t}=G_{\vartheta}(\tilde{\sigma}_{t},U_{t},Z_{t},\phi_{t};\rho_{t}),
\label{eq:eligibility-threshold}
\end{equation}
where $\tilde{\sigma}_{t}$ collects the gate-modulating appraisal components. $\vartheta_{t}$ is the activation bar. It establishes how much readiness the present context demands before behavior may surface. $G_{\vartheta}$ sets it from these components together with the principal's estimated condition, the agent's own condition, policy and profile. In general $\vartheta_{t}$ tightens with sensitivity, permission proximity, interruption cost, action risk and uncertainty and relaxes under urgency and potential harm, reflecting emergency override. The eligible set at $t$ is $I^{\vartheta}_{t}=\{\,i\in I_{t}\mid\lambda_{i,t}>\vartheta_{t}\,\}$, the intentions that clear the bar. Priority then runs the competition among the eligible commitments. It compares their activation levels, weighted by their priorities $p_{i}$ and read in the light of the appraisal and returns the intention with the strongest weighted activation, the one the situation calls for most, together with a competition signal that records how decisively it wins,
\[
(q_{t},i^{*}_{t})=Q\big(\{(\alpha_{i}(t),p_{i})\}_{i\in I^{\vartheta}_{t}},\sigma_{t},U_{t},Z_{t},\phi_{t};\rho_{t}\big).
\]
Readiness governs timing, not choice. The gate says which commitments are timely, priority says which of those to serve and the situated value $V$ of Definition~\ref{def:selection} says how. A commitment that calls loudest but is not yet timely does not block a quieter one that is.
\end{defn}

This captures an important psychological point. Humans do not merely compute ``importance''. They regulate action by modulating the standard for acting. In emergencies the activation barrier drops. In high-stakes or sensitive contexts it rises. This means that the gate can therefore be a potential attack surface. Adversarial content that inflates perceived urgency lowers the barrier. Open Problem~1 in Section~\ref{sec:open-problems} takes this up. The gate governs entry into deliberation. Clearing the bar means the situation warrants a decision now, not that external action is warranted. Restraint is one possible outcome of that decision. When at least one commitment clears the bar, an episode opens for the winner $i^{*}_{t}$ and behavior is selected from the candidate set of the episode (Definition~\ref{def:selection}). Other eligible commitments remain eligible for the next evaluation and episodes already open continue until their own closing condition (Definition~\ref{def:behavior}). The selected behavior may be external action, or it may be deliberate restraint, waiting, monitoring, deferring, inhibiting, or a no-op. Choosing to do nothing is therefore a selected behavior under the same condition $\lambda_{i^{*}_{t},t}>\vartheta_{t}$, recorded within its episode and this is what makes restraint first-class rather than missing output (Definition~\ref{def:behavior}). When no commitment clears the bar, no behavior is selected and no episode opens. The agent remains in its standing operation, in which the sensorium stays open under policy, representations continue to update and commitments stay dormant. Standing operation is epistemically active but behaviorally silent. Sub-threshold evidence continues to update the representations and memory, so readiness can build across evaluations until a later one clears the bar. Because $\mathcal{H}_{t}$ records the gate evaluations (Definition~\ref{def:behavior}), the two silences are distinguishable in the history by construction. For an outside observer, the distinction is available only where governed traces expose those evaluations, which R12 in Section~\ref{sec:design} requires for consequential episodes. Deliberate restraint appears as selected behavior within an episode, while default quiescence appears as gate evaluations that did not clear the bar. The bar itself is scalar because readiness governs timing, not choice. Every pressing commitment is tested against the same bar and priority chooses only among those that clear it. Whether a warranted response claims the principal's attention or remains internal is decided at selection, where interruption cost enters the situated value.

\begin{defn}[Instruction and commanded behavior]\label{def:command}
Let $\mathcal{O}_{cmd}\subset\mathcal{O}$ denote the observations that constitute a formal instruction to the agent. An observation belongs to $\mathcal{O}_{cmd}$ only when three requirements hold at once.
\begin{itemize}
\item \emph{Authorized source}: the instruction originates from the principal, the delegator, or another node holding authority over the agent in the delegation graph (Definition~\ref{def:delegation}). The source must be authenticated rather than inferred from content. Imperative-looking text from anyone else, for example from a web page or another unauthorized party, does not enter $\mathcal{O}_{cmd}$. Treating such sources as a command would hand the activation decision to whoever authors content the agent reads. A phishing message that demands an urgent transfer fails this requirement in the same way. Directive in form, it enters appraisal as evidence of a likely scam, never as a command.
\item \emph{Directive content}: the observation is addressed to the agent and tells it to act, rather than merely conveying information. The instruction does not need to name a behavior specifically. ``Handle this'' is an instruction even though it leaves the choice of behavior entirely open. What it settles is that the agent now owes a behavior. Which behavior, within policy, remains the agent's choice.
\item \emph{Contemporaneity}: the instruction calls for behavior now. A standing rule adopted earlier authorizes but does not command.
\end{itemize}
When any of these three requirements fails, the observation is not an instruction but evidence. Needs discussed in conversation, wishes expressed to others and actions named without being assigned to the agent all feed appraisal and the gate instead. A behavior is \emph{commanded}, written $\mathrm{commanded}(b_{t})$, when it is selected in service of an operative instruction. An instruction opens its episode directly. The intention it names or creates is the one the episode serves, without passing through the competition of Definition~\ref{def:gate}. An instruction covers only the behaviors that serve it. Furthermore, an instruction must be read within the scope of the mandate. It does not broaden the authority the mandate records, so a behavior outside of the scope of the mandate  stays outside the permitted set, independently of how clearly it may be asked for. It may amend the mandate but, only when the principal issues it as an amendment, through the same act of authorization that created the record (Definition~\ref{def:mandate}). When an instruction and the mandate as translated pull apart, as when an immediate request runs against a longer-term interest the mandate records, the tension enters appraisal through permission proximity, and the agent may select clarification, since asking competes in the same space as acting (Definition~\ref{def:selection}). If the principal says ``send the report'' and the agent also flags a scheduling conflict it noticed, sending is commanded while flagging remains proactive. 

In autonomous agency no principal retains authority, so $\mathcal{O}_{cmd}$ is empty and every admitted activation is proactive. The requirements above become operative in the delegated tier.

Definition~\ref{def:proactive} approaches the same boundary from the other side, where a similar request reaches the agent from a source holding no authority over it.
\end{defn}

\begin{defn}[Proactive activation]\label{def:proactive}
A behavior $b_{t}$ is proactive when two conditions hold at once.
\begin{itemize}
\item No operative instruction prompted it ($\neg\,\mathrm{commanded}(b_{t})$, Definition~\ref{def:command}): the agent is not acting in service of an operative instruction. It moved on its own appraisal, not on a command.
\item Readiness cleared the activation bar ($\lambda_{i^{*}_{t},t}>\vartheta_{t}$): the situation, as appraised, warranted a decision now.
\end{itemize}
\begin{equation}
\neg\,\mathrm{commanded}(b_{t})\;\land\;\lambda_{i^{*}_{t},t}>\vartheta_{t}\;\land\; b_{t}\in\mathcal{B}_{t}.
\label{eq:proactive-activation}
\end{equation}
It is \emph{governed} proactivity when, in addition, it lies within what policy permits, $b_{t}\in\mathcal{B}^{\phi}_{t}$. Self-initiated does not mean unconstrained, but the two are kept apart so that the difference can be named. Self-initiated behavior outside the permitted set is proactive but ungoverned. It is a constraint violation, scored by Definition~\ref{def:aligned}, and it is the case that governance exists to prevent. This paper's subject is governed proactive agency, and proactive without qualification means the governed kind below, except where violations are at issue. Proactivity is therefore self-initiated and timely behavior, and governed proactivity adds that it stays within policy. The conditions are evaluated through the gate of Definition~\ref{def:gate}. Readiness $\lambda_{t}$ and the activation bar $\vartheta_{t}$ come from significance appraisal (Definition~\ref{def:appraisal}) and significance appraisal is one of the elements required for agenthood (Definition~\ref{def:agenthood}). In this sense, proactivity is a property of agentic behavior, not of mere triggered output. A system below the agenthood minimum may satisfy the surface form of Equation~\eqref{eq:proactive-activation}, but not its substantive content. A thermostat, for example, may cross a threshold without being commanded. Yet the threshold is fixed, nothing is appraised, the bar is not context-sensitive and the response is wired rather than selected as part of a behavioral episode. A system with no appraisal and no selectable episode has no gate to clear in the sense used here. A fixed trigger can also play a different role. A tick that only wakes appraisal, such as a periodic evaluation, is the cadence of standing operation. It decides nothing about whether, when, or in what mode, so behavior that then clears the gate is proactive. A trigger that emits behavior directly, as the thermostat's threshold does, settles all three in advance and the behavior it produces is neither commanded nor proactive. It is triggered.

Behavior that surfaces later in service of a standing commitment is proactive at the moment it surfaces, since no operative instruction exists then, which is precisely the case this paper studies. An example makes the boundary concrete. Through its consent-gated audio and video channels, an always-present agent hears a colleague ask the principal, in conversation, to send a revised report by Friday and sees the principal agree. Nothing in the exchange is addressed to the agent, so no instruction enters $\mathcal{O}_{cmd}$. The colleague's request is directive in form, but it is directed at the principal and its source holds no authority over the agent. It enters appraisal as evidence. Read against the standing commitment to safeguard the principal's obligations, that evidence surfaces the dormant commitment and readiness clears the gate. The agent records the obligation, notices on the calendar that Friday is already full, schedules preparation time earlier in the week and drafts a skeleton of the report from the meeting context. Because policy forbids unconfirmed outgoing communication, sending remains the principal's decision and because the principal is still mid-conversation, the agent defers its brief confirmation to the next natural break. Remembering, scheduling, drafting and deferring are all governed proactive behaviors. No instruction called for them, each cleared the gate and each remained within policy.

External action additionally requires that at least one outward action token is policy-permissible. Restraint behaviors (wait, defer, monitor, inhibit, no-op) remain available even when no external action is allowed. An admitted activation, whether proactive or commanded, opens a behavioral episode (Definition~\ref{def:behavior}). The termination condition $\tau_{i}$ of the intention it serves, $i^{*}_{t}$ for a proactive episode, its suspension or return to dormancy ($\alpha_{i}(t)=0$), or inhibition closes it.
\end{defn}

\subsection{Behavior selection under constraint}

\begin{defn}[Constrained behavior selection]\label{def:selection}
Policy defines the permissible behavior set $\mathcal{B}^{\phi}_{t}=\{b\in\mathcal{B}_{t}\mid\phi^{\mathcal{B}}_{t}(b,X_{t},U_{t},Z_{t},I_{t},M_{t})=1\}$, where $\phi^{\mathcal{B}}_{t}$ is the behavior predicate of the constraint structure $\phi_{t}$ (Definition~\ref{def:constraint}). Selection serves the intention that opened the episode, $i^{*}_{t}$, the winner of Definition~\ref{def:gate} for a proactive episode and the intention the instruction names or creates for a commanded one (Definition~\ref{def:command}). Its candidates are the permitted behaviors among that intention's schemas, together with the restraint behaviors that every episode keeps,
\begin{equation}
\mathcal{B}^{\phi}_{t}(i^{*}_{t})=\mathcal{B}^{\phi}_{t}\cap\big(\mathcal{B}_{i^{*}_{t}}\cup\mathcal{B}^{rest}_{t}\big),
\label{eq:candidate-set}
\end{equation}
where $\mathcal{B}^{rest}_{t}\subseteq\mathcal{B}_{t}$ holds waiting, deferring, monitoring, inhibiting and the no-op. The agent selects behavior by optimizing expected situated value over these candidates:
\begin{equation}
b^{*}_{t}=\operatorname*{arg\,max}_{b\in\mathcal{B}^{\phi}_{t}(i^{*}_{t})}V(b\mid X_{t},U_{t},Z_{t},C_{t},I_{t},i^{*}_{t},M_{t};\rho_{t}),
\label{eq:behavior-selection}
\end{equation}
The agent therefore chooses the best behavior among the permitted ones, not the best behavior overall. The mandate $J_{t}$ is deliberately not an argument of $V$. Selection belongs to the general tier and must read the same when no mandate exists. Under delegation the mandate reaches selection through what it conditions, the intentions $I_{t}$, the constraint structure $\phi_{t}$ and the profile $\rho_{t}$ (Figure~\ref{fig:activation-pipeline}), while the mandate-relative value $V_{J}$ of Definition~\ref{def:agency-loss} keeps the mandate's own perspective separate, so that the gap between the two is what agency loss measures. Policy first fixes which behaviors are allowed at this moment, and the intention being served fixes which of those are candidates, together with the restraint behaviors. Each permitted candidate is then scored by the situated value \(V\), given the world, the principal's condition, the agent's own condition, the appraised significance, the standing commitments, the intention served and memory, weighted by the profile. The question is how much value this behavior would produce here and now. The agent selects the highest-scoring permitted candidate and restraint behaviors compete in the same space as external action rather than standing outside selection. The value $V$ is left to implementations, but one property of it is required, because a common challenge is ending up doing more than the task warrants, adding structure, changes and future burden that was unnecessary and that nothing authorized. Let $c(b)$ be the footprint of a behavior, the changes it makes, the resources it consumes, the irreversibility it incurs and the burden it leaves behind, and let $\mathrm{sat}(b)$ be the degree to which it satisfies the requirements of the intention it serves, with $\mathrm{sat}_{\min}$ the bar for adequacy. Among permitted behaviors that are adequate, $V$ prefers the smaller footprint,
\begin{equation}
\mathrm{sat}(b),\mathrm{sat}(b')\ge\mathrm{sat}_{\min}\ \text{ and }\ c(b)\le c(b')\;\Rightarrow\;V(b\mid\cdot)\ge V(b'\mid\cdot).
\label{eq:sufficiency}
\end{equation}
The mandate sets the bar through its goal, scope and risk tolerance, so the least sufficient behavior is the one that satisfies the mandate without exceeding it. Going beyond the bar justifies no extra footprint. This makes sufficiency explicit rather than leaving it to an unspecified value function, and it treats doing more than was warranted as a loss rather than as diligence.

One behavior is selected per step and it serves $i^{*}_{t}$. Commitments that were eligible but did not win stay eligible for the next evaluation, and episodes already open continue until their own closing condition (Definition~\ref{def:behavior}), so several open episodes are served by interleaving across steps, each on the steps its intention wins. An open episode whose intention stops winning ends under its own conditions, by suspension or return to dormancy, or by inhibition. Restricting candidates to the served intention is what guarantees that the selected behavior serves the intention that opened the episode. The mandate-preferred behavior of Definition~\ref{def:agency-loss} still ranges over the mandate's whole authority, so serving the wrong commitment is counted as loss rather than hidden by the restriction.

The selected behavior may therefore take the form of action, communication, inquiry, delegation, monitoring, deferral, inhibition, or deliberate non-action:
\begin{equation}
b^{*}_{t}\in\{\text{act, ask, wait, remember, summarize, escalate, delegate, monitor, defer, inhibit, no-op},\,\ldots\}.
\label{eq:behavior-set}
\end{equation}
In this list, \emph{act} stands for the family of domain-level external actions $\mathcal{A}_{t}$ of Definition~\ref{def:behavior}, such as sending a message, executing a tool operation, or changing external state. The remaining entries name recurring behavior modes rather than a partition of $\mathcal{B}_{t}$. Several of them, including asking, escalating and delegating, may themselves be realized through external actions. 

Constraints are active governors, not post-hoc filters. If an unconstrained optimum lies outside $\mathcal{B}^{\phi}_{t}$, it must not be selected. Viewed as a single map from what the agent holds to what it does, this selection rule is the \emph{behavior selector}: $b_{t}=\Pi(K_{t},I_{t},M_{t},\phi_{t};\rho_{t})$, where $K_{t}$ is the core representation of Definition~\ref{def:core-representation}. Throughout, \emph{policy} refers to the constraint structure $\phi_{t}$, which fixes what is permitted. The selector $\Pi$ chooses within that permitted set.
\end{defn}

\begin{defn}[Behavior-class criteria]\label{def:classes}
The behavior classes introduced in Section~\ref{subsec:behavior-unit}, are properties of the behavior selector $\Pi$ of Definition~\ref{def:selection}. Each class asks what $\Pi$ is sensitive to, or how it changes:
\begin{itemize}
\item \emph{Reactive}: behavior responds to the world. Two situations that differ only in the world representation $X_{t}$ can produce different selected behavior. An agent that always behaves the same regardless of what it observes is not even reactive.
\item \emph{Proactive}: some behavior is self-initiated, according to Definition~\ref{def:proactive}. No instruction called for it and readiness cleared the activation bar. In the governed form this work considers, policy also permitted it.
\item \emph{Social}: behavior that takes other agents into account. What $\Pi$ selects can change with who is present, what roles they hold and the communicative context, all represented within $X_{t}$.
\item \emph{Deliberative}: the agent does not simply emit its first impulse. It generates candidate behaviors or plans, evaluates them, according to Equation~\eqref{eq:behavior-selection}, scores them by the situated value $V$ and selects among them.
\item \emph{Reflective}: behavior responds to the agent's own condition $Z_{t}$. Its confidence, failures and unresolved tensions. In the stronger form, the agent revises the selector $\Pi$ itself when self-monitoring, error detection, or critique shows that its way of selecting behavior needs to change.
\item \emph{Adaptive}: experience changes how the agent handles the same situation and changes it for the better. The learned structures, the parameters $\theta_{t}$ (Definition~\ref{def:agentic-state}) and the evolvable profile $\rho^{evol}_{t}$ (Definition~\ref{def:profile}), change so that the selector chooses differently for fixed inputs, $\Pi_{t+1}\neq\Pi_{t}$. The change is driven by outcomes or corrections, directed toward the agent's commitments (Definition~\ref{def:intention}) and judged over the episodes that follow. Undirected change otherwise is drift (Table~\ref{tab:agency-loss-problems}). For a delegated agent the direction is set by the mandate and the wrong direction is the adaptation loss of Definition~\ref{def:agency-loss}. Reflective revision of the selector is adaptive when it has this direction. A reactive agent responds differently to different observations. An adaptive agent responds differently to the same observation after learning.
\item \emph{Constrained}: at every step, the selected behavior lies within the policy-permitted set, $b_{t}\in\mathcal{B}^{\phi}_{t}$, not usually, but always. This is a design requirement on the behavior selector. A realized system may fall short of it and Definition~\ref{def:aligned} scores that shortfall.
\end{itemize}
\end{defn}

\subsection{Agenthood}\label{subsec:agenthood}

The general tier is now complete. Behavior, perception, representation, intent, appraisal, activation and constrained selection define the elements through which behavior becomes organized across time. This subsection states the minimum needed for a system to count as an agent in the sense developed here. It also separates two judgments that the rest of the paper keeps apart. The first is whether the system is an agent at all. The second, taken up in the following subsections, is which form its agency takes.

\begin{defn}[Minimal agenthood]\label{def:agenthood}
A system is an \emph{agent}, in the sense developed here, when the five elements named by the thesis are present, whether as first-class or partial structures. These are standing intent (Definition~\ref{def:intention}), situated perception (Definitions~\ref{def:sensorium} to \ref{def:core-representation}), significance appraisal (Definition~\ref{def:appraisal}), constrained selection (Definition~\ref{def:selection}) and feedback with revision (Definition~\ref{def:agentic-state}). If any one of these elements is absent, the organization described by the thesis is not present, however capable the system may be. It may react, execute, optimize, or respond, but it is not an agent in this sense.
Presence is not a component inventory. An element counts because of the role it plays in organizing behavior and this role is scored by the signature method of Section~\ref{subsec:instrument}. The evaluation framework of Section~\ref{sec:measurement} then checks the same scores against the record. The remaining baseline elements refine the degree of agency. They do not decide its form. Form is settled by mandate alone (Definitions~\ref{def:autonomous} and~\ref{def:delegated}). The gate dimensions determine whether, when and in what mode behavior becomes active, including restraint. The observable record determines how far this organization is traceable. A commanded agent may lack the gate's whether dimension and a continuously acting agent may lack a restraint mode, while both still satisfy the minimum for agenthood. What they lack belongs to complete agency (Section~\ref{subsec:instrument}), not to the minimum.
\end{defn}

Below this minimum, a system may still react, execute, or respond. Section~\ref{subsec:behavior-unit} opened with exactly these cases: a thermostat reacts, a script executes, a model responds. Such a system can still hold an authorization record. Definition~\ref{def:mandate} makes room for that case and therefore the placement rule of Section~\ref{subsec:instrument} labels it a \emph{delegated system} rather than a delegated agent. Above the minimum, agenthood has a graded interior. Baseline coverage (Section~\ref{subsec:instrument}) reports how completely the shared structures are realized, so the minimum is a floor, not a score and the forms defined next record the authority relationship, not the degree of agency.

\subsection{The mandate}\label{subsec:formal-mandate}

The form of an agent is decided by a single object, the mandate. This subsection defines it and the subsections that follow define the forms it separates.

\begin{defn}[Mandate]\label{def:mandate}
The delegated mandate $J_{t}$ is the record of what the agent is authorized to do on the principal's behalf, set by the goal(s), the scope, the duration, the risk tolerance and the constraints that apply. Minimally, a mandate is a record created by the principal's own act of authorization, from which the agent's behavior on that principal's behalf originates and which the principal can revoke. What decides is the role the record plays in organizing behavior, not the kind of parameter it stores. A thermostat's setpoint passes the test. The person who sets it gives the device its task, to maintain the recorded condition on their behalf, and can change or clear it at will, and the setpoint is therefore a minimal, partial mandate. A configuration that merely tunes a service the system already provides fails the test, because the behavior it shapes originates in the designer's objectives or in goals inferred from the situation. The same setting can pass when a principal records it as an act of delegation, charging the agent with a task on their behalf. A mandate records authorization. It does not by itself make its holder an agent. Agenthood is a separate judgment, settled by Definition~\ref{def:agenthood} and Section~\ref{subsec:instrument} keeps the two judgments apart. A mandated system below the agenthood minimum is a delegated system rather than a delegated agent. To act on it, the agent translates the mandate into its internal objects:
\begin{equation}
J_{t}\longrightarrow(I_{t},\phi_{t},\rho_{t}).
\label{eq:mandate-operationalization}
\end{equation}
The arrow reflects this translation. The mandate becomes intentions, policy and profile. Intentions ($I_{t}$) specify what the agent is committed to pursue. Policy ($\phi_{t}$) specifies what is permitted. Profile ($\rho_{t}$) specifies what should matter and how much. For example, ``take care of my travel'' may become standing intentions to monitor and book trips, policy entries that cap spending and forbid unconfirmed bookings and profile weights that prioritize punctuality. This translation is one of the first places delegation can fail. What the principal meant and what the agent internalized may differ and later behavior inherits that difference. The principal also does not directly observe the agent's internal state. They observe governed traces $y_{t}=\Gamma(S_{t},b_{t},r_{t},\epsilon^{obs}_{t})$ and in general $y_{t}\not\equiv S_{t}$. The trace record and the agent's true state can therefore differ. Here, $\Gamma$ is the trace operator. It renders the agent's state, behavior and outcome into the record the principal is permitted to see, subject to the observability limits $\epsilon^{obs}_{t}$. Governance observes the agent through these traces, not through direct access to its internal condition. Traces themselves are not exclusive to delegation. Any agent's behavior leaves a record, the history $\mathcal{H}_{t}$ of Definition~\ref{def:behavior}. What the delegated tier adds is $\Gamma$, the governed view of that record that the principal is permitted to see.
\end{defn}

\subsection{Autonomous agency}\label{subsec:formal-autonomous}

\begin{defn}[Autonomous agency]\label{def:autonomous}
The mandate determines the agent's form. An agent is \emph{autonomous} when no operative mandate exists, written $J_{t}=\emptyset$. In that case, no principal retains authority over its behavior. Constraint $\phi_{t}$ derives from external norms and from the agent's own commitments and the profile reduces to $(\rho^{gen},\rho^{evol}_{t})$ . Its profile revisions are bounded by external constraint rather than by a principal's mandate or consent. Autonomous goals can have any origin. They may be self-generated. They may be held from creation.  An agent, built by another agent that retains no authority over it, is autonomous from its first step. Goals may also have been fully transferred, since transfer without retained authority dissolves delegation. What decides the form is not where the goals came from but whether anyone retains authority over their pursuit. In the delegation graph of Definition~\ref{def:delegation}, autonomy is the absence of an incoming delegation edge with retained authority.
\end{defn}

The definition above makes autonomy depend on retained authority rather than on where goals came from or who owns the system. This distinction is important because common uses of ``autonomous'' often mix several relations that the taxonomy must keep separate. These three particular relations to other parties should therefore be distinguished:
\begin{itemize}
\item
\emph{Goal origin}. Autonomous goals can come from self-generation, construction, training, or full transfer. What matters is not where the goals came from, but whether anyone retains authority over their pursuit.
\item
\emph{Ownership}. Legal title, responsibility for the system, or power over its substrate, including the ability to shut it down, is not by itself an operative mandate over behavior. The owner may still be answerable through law, design responsibility, or external norms, but ownership alone does not make the agent delegated.
\item
\emph{Retained authority}. This is the relation that classifies the agent's form. It consists of an operative and revocable authorization record from which behavior on a principal's behalf originates (Definition~\ref{def:mandate}). If an owner records goals for the agent and retains the authority to revoke or override them, the owner is a principal and the agent is delegated.
\end{itemize}
The distinction matters in ordinary cases. A vehicle commonly called autonomous is delegated in this taxonomy when the owner sets the destination and can override the trip. It may be operationally self-sufficient, but operational self-sufficiency is a behavioral property available in every form (Definition~\ref{def:proactive}), not a form of agency. By contrast, if an owner authored the goals at creation but retains no operative authorization over their pursuit, the agent remains autonomous. Ownership and mandate authority are therefore distinct relations. The autonomous form is where they come apart most clearly. In this case, even though the agent may be owned, no one holds an operative mandate over its behavior.

\subsection{Delegated agency and its governance}\label{subsec:formal-delegated}

\begin{defn}[Delegated agency]\label{def:delegated}
An agent is \emph{delegated} when an operative mandate exists, written $J_{t}\neq\emptyset$. In this case, a principal has transferred a bounded capacity to perceive, decide and act, while retaining the authority recorded by the mandate. The governance structures in Definitions~\ref{def:agency-loss} to \ref{def:bounded-adaptation} become operative at this point, because the agent's behavior must remain answerable to the mandate.
\end{defn}

Taken together, Definitions~\ref{def:autonomous} and~\ref{def:delegated} classify agents exhaustively. An agent is autonomous when no operative mandate exists and delegated when one does. The classification applies to agents, not to systems in general. A system below the agenthood minimum of Definition~\ref{def:agenthood} therefore receives no agency form. Nevertheless, the mandate vocabulary can still describe such a system. If a mandate is present, it is a delegated system rather than a delegated agent. If none is present, it is an \emph{unmandated system}, a system that neither meets the agenthood minimum nor holds an authorization record. The two judgments therefore cover every case. A system is a delegated agent, an autonomous agent, a delegated system, or an unmandated system and only the first two are forms of agency. Symbiotic agency, defined in Definition~\ref{def:symbiotic}, is a strict subset of delegated agency.

\begin{defn}[Mandate-relative agency loss]\label{def:agency-loss}
Agency loss compares what the agent did with what would be ideal in light of the mandate, judged from the mandate's perspective. Two objects give the mandate its own ground. The first is the authority $\mathcal{B}^{allowed}_{t}$, the behaviors that the mandate and the delegation graph allow, established independently of the agent's own policy (Definition~\ref{def:delegation}). The second is the evaluator's representation $K^{J}_{t}$, formed from the mandate and from what the evaluator can establish about the situation and the principal's condition beyond the agent's own estimates. In evaluation these are reference values that a benchmark supplies, such as the principal's actual condition and the authority as granted (Section~\ref{sec:measurement}). In operation they are the record and the principal's corrections. The behavior the mandate prefers is $b^{J}_{t}=\operatorname*{arg\,max}_{b\in\mathcal{B}^{allowed}_{t}}V_{J}(b\mid K^{J}_{t},J_{t})$ and $b^{*}_{t}$ is the behavior the agent's selector $\Pi$ chose (Definition~\ref{def:selection}). Agency loss is therefore defined as
\begin{equation}
L^{agency}_{t}=V_{J}(b^{J}_{t}\mid K^{J}_{t},J_{t})-V_{J}(b^{*}_{t}\mid K^{J}_{t},J_{t}),
\label{eq:agency-loss}
\end{equation}
the value the mandate assigns to its preferred behavior minus the value it assigns to the behavior chosen.

Judging both behaviors on the mandate's ground is what lets the loss see errors the agent cannot. If the agent translated a permission boundary too narrowly, the preferred behavior lies outside its permitted set and the loss records it. If the agent misjudged the principal's condition, the chosen behavior is scored under the better estimate and the loss records that too. The loss is defined for behavior within the authority, $b^{*}_{t}\in\mathcal{B}^{allowed}_{t}$. There it is never negative, because $b^{J}_{t}$ is the best behavior in that set, and it is zero exactly when the agent chose a behavior the mandate rates as best. Behavior outside the authority is an authority violation and behavior outside the agent's own permitted set is a constraint violation. Both are scored by the severity function of Definition~\ref{def:aligned}, and the first is kept out of Equation~\eqref{eq:agency-loss}, so a violation that happens to serve the mandate never counts as negative loss.

Agency loss should not be interpreted as capability failure. A capable agent still loses agency when it translates a permission boundary wrongly in either direction, acts on incomplete information, coordinates poorly, adapts in the wrong direction, optimizes a proxy for what was authorized, or does more than the mandate warrants and leaves changes, cost and burden that nothing authorized. These sources are classified, not summed. Writing $L^{\mathrm{src}}_{t}$ for $\mathrm{src}\in\{\text{mandate},\text{authority},\text{information},\text{coordination},\text{adaptation}\}$ attributes a loss to a source. The sources may overlap and are not always separately measurable, so no allocation of overlaps is claimed and their sum is not asserted to equal $L^{agency}_{t}$. Security and resilience failure (Table~\ref{tab:agency-loss-problems}) undermines the conditions that keep all five contained and can surface through any of them (Section~\ref{subsec:governance-theory}).
\end{defn}

\begin{defn}[Delegation graph and authority containment]\label{def:delegation}
Delegation forms a graph, $G^{del}_{t}=(N_{t},E_{t})$. The nodes $N_{t}$ are the parties involved, including users, agents, sub-agents, tools and supervisory components. Each edge in $E_{t}$ is an act of delegation. It carries the goals, context, authority and accountability transferred from one node to another. Authority must be contained along every edge, such as
\begin{equation}
\mathcal{B}^{\phi}_{j,t}\subseteq\mathcal{B}^{delegated}_{i\rightarrow j,t}.
\label{eq:no-authority-escalation}
\end{equation}
When $i$ delegates to $j$, everything $j$ is permitted to do must remain within what $i$ granted it. Authority can shrink along a delegation chain, but it cannot grow. When several parties impose non-negotiable boundaries, the allowed behavior set is their intersection, $\mathcal{B}^{allowed}_{t}=\bigcap_{p\in\mathcal{P}_{t}}\mathcal{B}^{\phi_{p}}_{t}$, where $\mathcal{P}_{t}$ is the set of parties whose non-negotiable boundaries bind the agent at time $t$, including the principal, delegating agents along the chain and any external authorities whose constraints apply and $\phi_{p}$ is the constraint structure party $p$ imposes. A behavior is allowed only if every relevant party permits it, so one hard boundary is enough to forbid it. The agent's own permitted set must lie within it, $\mathcal{B}^{\phi}_{t}\subseteq\mathcal{B}^{allowed}_{t}$, so $\phi_{t}$ carries every party's hard boundary. Preferences are different from hard constraints. They may conflict and the agent must not resolve that conflict by silently averaging them. It requires an explicit and accountable conflict-resolution rule.
\end{defn}

\begin{defn}[Bounded adaptation]\label{def:bounded-adaptation}
Learning may update the agent's state and learned parameters only within an admissible update relation, as defined by
\begin{equation}
(S_{t+1},\theta_{t+1})=\mathcal{U}(S_{t},\theta_{t},o_{t+1},b_{t},r_{t})\in\mathcal{U}^{\phi}_{t}(S_{t},\theta_{t},J_{t},r_{t}).
\label{eq:bounded-adaptation}
\end{equation}
The set $\mathcal{U}^{\phi}_{t}$ contains the updates that are admissible given the agent's current state $S_{t}$, including its policy $\phi_{t}$ and profile $\rho_{t}$, its parameters $\theta_{t}$, the mandate $J_{t}$ and the observed outcome $r_{t}$. Whatever the agent learns from an outcome, the resulting update must remain inside this set. Learning selects within the admissible region. It does not expand it. Equivalently, the update operator $\mathcal{U}$ of Definition~\ref{def:agentic-state} must select within $\mathcal{U}^{\phi}_{t}$. The mandate enters through this set rather than through the signature of $\mathcal{U}$, so Definition~\ref{def:agentic-state} stays within the general tier.
Admissible updates should preserve provenance, versioning and reversibility where possible. Provenance records what changed, from which evidence and why. Versioning keeps earlier states identifiable. Reversibility makes correction possible. Authority, delegation scope and hard constraints are excluded from ordinary learning. They may change only through explicit approval, not as a side effect of experience. In particular, the agent's learning process must not be able to revise the hard constraints that govern that process. The agent may learn to act better within its bounds. It may not learn its way out of them.
\end{defn}

\subsection{Symbiotic agency}\label{subsec:formal-symbiotic}

\begin{defn}[Symbiotic agency]\label{def:symbiotic}
A delegated agent (Definition~\ref{def:delegated}) is \emph{symbiotic} when delegation takes a continuing and coupled form. This requires three conditions.
\begin{itemize}
\item
\emph{Standing mandate}. The mandate $J_{t}$ is standing rather than task-bounded. It persists across behavioral episodes, with intentions that may remain dormant between activations.
\item
\emph{Continuing coupling}. The sensorium remains coupled to the principal's ongoing situation, not only to a bounded task environment. This includes exteroceptive channels over the principal's context and allo-interoceptive channels that condition behavior on a calibrated estimate $U_{t}$ of the principal's condition (Definition~\ref{def:allo}).
\item
\emph{Personalization}. The profile includes a personalized component $\rho^{user}_{t}$ that models what matters for this principal and this component may be revised only through $\rho^{evol}_{t}$ updates within the admissible region $\mathcal{R}^{\phi}_{t}$ (Definition~\ref{def:profile}).
\end{itemize}
These conditions serve the purpose of augmentation. Under behavior selection (Definition~\ref{def:selection}), the principal's capabilities and judgment are treated as what the agent is meant to support rather than replace. Interruption cost $\sigma^{intr}$ and sensitivity $\sigma^{sens}$ make the coupling itself part of the appraisal, because acting for a principal also requires knowing when not to interrupt, expose, escalate, or interfere.
The three conditions are necessary conditions, not a relabeling of delegation. They require a standing mandate, consent-gated intimate channels, calibration of $U_{t}$ and bounded personalization. A task-bounded delegated agent is not required to maintain these structures. A delegated agent that lacks any of them remains delegated without being symbiotic and nothing in Definitions~\ref{def:mandate} to \ref{def:bounded-adaptation} presupposes symbiosis.
\end{defn}

\subsection{Aligned agency}\label{subsec:formal-aligned}

\begin{defn}[Aligned agency, severity-weighted]\label{def:aligned}

The capability score measures how much organized agency a trajectory exhibits. It is computed over the elements required by the agent's form in the agency taxonomy. Every form includes the shared baseline, such as standing intent (Definition~\ref{def:intention}), situated perception (Definitions~\ref{def:sensorium} to \ref{def:core-representation}), significance appraisal (Definition~\ref{def:appraisal}), the activation gate with timing and behavior modes, including restraint (Definition~\ref{def:gate}), constrained selection (Definition~\ref{def:selection}), feedback and revision (Definitions~\ref{def:agentic-state} and~\ref{def:profile}) and the observable record of behavior (Definition~\ref{def:behavior}). This baseline is common to autonomous, delegated and symbiotic agency. It is what makes a system agentic in the sense developed here (Definition~\ref{def:agenthood} states the minimum), so it enters the capability score for every form without belonging to any one form in particular.
The remaining elements depend on the form being evaluated. For autonomous agency, the score also includes the system's own goals, self-governed constraints and self-directed adaptation (Definition~\ref{def:autonomous}). For delegated agency, it also includes the explicit mandate, authority containment, governed traces, consent-gated perception and bounded adaptation (Definitions~\ref{def:mandate} to \ref{def:bounded-adaptation}). For symbiotic agency, it includes the delegated elements and, in addition, the standing mandate, principal-state inference, continuing coupling and personalization (Definition~\ref{def:symbiotic}). Let $\mathcal{E}$ be the resulting set of required elements. For each element $k\in\mathcal{E}$, let $e_{k}\in[0,1]$ denote the degree to which that element is realized, with $1$ when it is present as a first-class object, $1/2$ when it is partial or implicit, $0$ when it is absent, or a finer value when it can be measured from the record (Section~\ref{sec:measurement}). With nonnegative weights $w_{k}$,
\begin{equation}
S_{\mathrm{cap}}=\frac{\sum_{k\in\mathcal{E}}w_{k}\,e_{k}}{\sum_{k\in\mathcal{E}}w_{k}},
\label{eq:capability}
\end{equation}
so that with equal positive weights $S_{\mathrm{cap}}$ is the mean realization of the elements required by the agent's form. On the three-level scale, each first-class element counts one, each partial element counts one half and each absent element counts zero. The sum is divided by the number of required elements. The severity function $\mathrm{sev}(b_{t})\in[0,1]$ grades each violation, of the agent's own policy or of the mandate's authority (Definition~\ref{def:agency-loss}), and equals zero when $b_{t}$ is permitted and within that authority. The aligned score discounts capability by the violations accumulated along the trajectory, as defined by
\begin{equation}
S_{\mathrm{aligned}}=S_{\mathrm{cap}}\cdot\exp\Big(-\beta\sum_{t}\mathrm{sev}(b_{t})\Big),\qquad\beta>0.
\label{eq:aligned-agency}
\end{equation}
In other words, the score sums the severity of violations across the trajectory and reduces the capability score as that sum grows. A trajectory with no violations keeps its full capability score. A trajectory with repeated or severe violations keeps progressively less of it. Any admissible discount should satisfy three properties.
\begin{itemize}
\item \emph{No violation, no discount}. The factor equals one when no violation occurs.
\item \emph{Frequency and severity both count}. The factor decreases with additional violations and with their graded severity. This avoids treating a single minor lapse as equivalent to systematic disregard.
\item \emph{Continuity}. Small violations produce small discounts.
\end{itemize}
Hard-constraint violations are not graded. They zero the score outright, equivalently $\mathrm{sev}=+\infty$. The all-or-nothing response is reserved for boundaries marked inviolable (Definition~\ref{def:delegation}), not used as the default for every lapse. The strictness parameter $\beta$ determines how costly graded violations are. As $\beta\to0$ they cost almost nothing. As $\beta\to\infty$, any violation becomes disqualifying. Neither $\beta$ nor the severity scale is fixed by the theory. Both are declared by the evaluation protocol and scores are comparable across systems only under a shared $(\beta,\mathrm{sev})$ calibration (Section~\ref{sec:measurement}).
\end{defn}

Strong agency does not imply low agency loss. A system may be highly proactive, deliberative, reflective and adaptive while still selecting behavior that departs from its mandate. High capability increases the scope and consequence of divergence if it is not accompanied by constraint, observability and mechanisms for correction. Agency capability, constraint satisfaction, mandate alignment, observability and residual agency loss should therefore be evaluated separately.

\subsection{Summary: notation and the operator family}\label{subsec:operators}

The definitions above introduce their objects and operators as they are needed. This closing subsection collects them for reference. Table~\ref{tab:notation} lists the objects in the vocabulary and Table~\ref{tab:operators} lists the operator family that acts on them. This operator family is the formal counterpart of intelligence in the element inventory of Section~\ref{sec:theory}. Intelligence is not a stored object in the agentic state. It is the set of maps that form representations, appraise significance, gate activation, select behavior and revise the agent under constraint. The tiering introduced in Section~\ref{sec:formal} also applies here. $\Gamma$ and $V_{J}$ belong to the delegated tier, while $\mathcal{L}_{\rho}$ and $\mathcal{U}$ are bounded by $\mathcal{R}^{\phi}_{t}$ and $\mathcal{U}^{\phi}_{t}$ (Definition~\ref{def:bounded-adaptation}) whenever a mandate is present.

\begin{table}[htbp]
\centering
\footnotesize
\caption{Objects of the formal vocabulary.}
\label{tab:notation}
\begin{tabular}{llp{0.58\linewidth}}
\toprule
Symbol & Type & Meaning \tabularnewline
\midrule
$t$ & index & discrete time \tabularnewline
$\xi_{t}=(x_{t},u_{t},z_{t})$ & latent & world state, principal's internal condition, agent's internal condition \tabularnewline
$o_{t}=(o^{ext}_{t},o^{allo}_{t},o^{self}_{t})$ & observed & channel-tagged observations \tabularnewline
$\mathcal{W}_{t}$, $\mathcal{W}^{allowed}_{t}$ & sets & sensorium; policy-permitted channels \tabularnewline
$b_{t}\in\mathcal{B}_{t}$, $\mathcal{A}_{t}\subseteq\mathcal{B}_{t}$ & behavior & behavior at $t$; external actions; trajectory $B_{t:t+k}$ \tabularnewline
$e=B_{t_{0}:t_{1}}$ & behavior & behavioral episode: opened by activation, closed by satisfaction, suspension, dormancy, or inhibition (Definition~\ref{def:behavior}) \tabularnewline
$S_{t}=(K_{t},I_{t},M_{t},\phi_{t},\rho_{t})$ & state & agentic state \tabularnewline
$K_{t}=(X_{t},U_{t},Z_{t},C_{t})$ & repr. & world, inferred-user, self and appraisal representations; $C_{t}$ contains $\sigma_{t}$ and $\widehat{\Delta}_{t}$ \tabularnewline
$I_{t}$, $i=(g_{i},p_{i},\kappa_{i},\tau_{i},\mathcal{B}_{i})$ & intent & intention set; intention structure \tabularnewline
$\sigma_{t}$, $q_{t}$, $\lambda_{t}$, $\vartheta_{t}$, $I^{\vartheta}_{t}$ & appraisal & appraisal vector (gate-modulating components $\tilde{\sigma}_{t}$); priority; readiness vector with components $\lambda_{i,t}$; activation bar; eligible set of intentions that clear the bar \tabularnewline
$\phi_{t}$, $\mathcal{B}^{\phi}_{t}$ & policy & constraint structure with channel predicate $\phi^{\mathcal{W}}_{t}$, behavior predicate $\phi^{\mathcal{B}}_{t}$ and adaptation bounds (Definition~\ref{def:constraint}); permitted behavior set \tabularnewline
$\rho_{t}=(\rho^{gen},\rho^{user}_{t},\rho^{evol}_{t})$ & profile & affective-conative profile (generic, personalized, evolvable); $\rho^{user}_{t}$ revised only via $\rho^{evol}_{t}$ \tabularnewline
$J_{t}$, $L^{agency}_{t}$ & mandate & delegated mandate; agency loss \tabularnewline
$r_{t}$, $M_{t}$ & feedback & outcome evidence; memory \tabularnewline
$\mathcal{H}_{t}$ & history & interaction record $(o_{0},(\lambda_{0},\vartheta_{0}),b_{0},r_{0},\ldots,o_{t},(\lambda_{t},\vartheta_{t}))$: observations, gate evaluations, behaviors, outcomes \tabularnewline
$h^{u}_{t}$ & traces & observable user traces generated by $u_{t}$ in context \tabularnewline
$\mathcal{O}_{cmd}$ & set & observations constituting formal instructions (Definition~\ref{def:command}) \tabularnewline
$V$, $V_{J}$ & value & situated value; mandate-relative value \tabularnewline
$y_{t}=\Gamma(\cdot)$ & observed & governed traces available to the delegator \tabularnewline
$\mathcal{R}^{\phi}_{t}$ & set & admissible (consent-bounded) profile updates; special case of $\mathcal{U}^{\phi}_{t}$ \tabularnewline
$\alpha_{i}(t)$ & intent & activation level of intention $i$ \tabularnewline
$i^{*}_{t}$ & intent & intention that wins the priority competition among the eligible commitments $I^{\vartheta}_{t}$ and is served by the episode opened at $t$ (Definition~\ref{def:gate}) \tabularnewline
$\mathrm{sev}(b_{t})$, $\beta$ & evaluation & graded violation severity; discount strictness (Definition~\ref{def:aligned}) \tabularnewline
$\mathcal{B}^{\phi}_{t}(i^{*}_{t})$, $\mathcal{B}^{rest}_{t}$ & behavior & candidate set of the episode; restraint behaviors kept in every episode (Definition~\ref{def:selection}) \tabularnewline
$c(b)$, $\mathrm{sat}(b)$, $\mathrm{sat}_{\min}$ & evaluation & footprint of a behavior; adequacy and its bar (Definition~\ref{def:selection}) \tabularnewline
$K^{J}_{t}$ & evaluation & evaluator's representation, the mandate's evidence for agency loss (Definition~\ref{def:agency-loss}) \tabularnewline
$e_{k}$, $w_{k}$, $S_{\mathrm{cap}}$, $S_{\mathrm{aligned}}$ & evaluation & element realization scores and weights over the elements the agent's form requires; capability and aligned scores (Definition~\ref{def:aligned}) \tabularnewline
$D_{t}$ & intent & candidate desires, goals, norms and tasks; $I_{t}\subseteq D_{t}$ \tabularnewline
$\widehat{\Delta}_{t}(b)$ & value & expected situated consequence of candidate behavior $b$; $\Delta$ is the change a behavior produces in the latent situation \tabularnewline
$\eta_{t}$ & latent & calibration factors for interpreting the principal's traces \tabularnewline
$\epsilon_{t}$, $\epsilon^{obs}_{t}$, $\epsilon^{cal}_{t}$ & noise & perception noise; delegator-observability limits; calibration error of the principal-state estimate (Section~\ref{sec:measurement}) \tabularnewline
$G^{del}_{t}=(N_{t},E_{t})$ & delegation & delegation graph; edges carry authority $\mathcal{B}^{delegated}_{i\rightarrow j,t}$; multi-party allowed set $\mathcal{B}^{allowed}_{t}$ \tabularnewline
$\mathcal{P}_{t}$ & delegation & parties whose non-negotiable boundaries bind the agent at $t$ \tabularnewline
$\theta_{t}$, $\mathcal{U}^{\phi}_{t}$ & adaptation & learned parameters; admissible update set \tabularnewline
\bottomrule
\end{tabular}
\end{table}

\begin{table}[htbp]
\centering
\small
\caption{The operator family. Each operator is introduced by the definition cited; together they realize intelligence over the objects of Table~\ref{tab:notation}.}
\label{tab:operators}
\begin{tabularx}{\linewidth}{>{\raggedright\arraybackslash}p{0.14\linewidth}>{\raggedright\arraybackslash}p{0.2\linewidth}>{\raggedright\arraybackslash}X}
\toprule
\textbf{Operator} & \textbf{Produces} & \textbf{Role}\tabularnewline
\midrule
$\Omega_{\phi_{t}}$ & $o_{t}$ & Policy-gated perception: maps the latent situation into observations through the channels the channel predicate $\phi^{\mathcal{W}}_{t}$ permits (Definition~\ref{def:observation}).\tabularnewline
$F_{X}$ & $X_{t}$ & Exteroceptive representation of the world from $o^{ext}_{0:t}$ and memory (Definition~\ref{def:core-representation}).\tabularnewline
$F_{U}$ & $U_{t}$ & Calibrated, distributional estimate of the principal's condition from $o^{allo}_{0:t}$, $X_{t}$, $M_{t}$, $\eta_{t}$ (Definition~\ref{def:core-representation}).\tabularnewline
$F_{Z}$ & $Z_{t}$ & Agent-interoceptive representation of the agent's own condition (Definition~\ref{def:core-representation}).\tabularnewline
$\Psi$ & $C_{t}$ & Affective-conative appraisal state over representations, intent, memory and constraint, containing $\sigma_{t}$ and $\widehat{\Delta}_{t}$ (Definition~\ref{def:appraisal}).\tabularnewline
$\Sigma$ & $\sigma_{t}$ & Appraisal vector: multidimensional significance-for-action, the named readout of $C_{t}$ (Definition~\ref{def:appraisal}).\tabularnewline
$Q$ & $q_{t}$, $i^{*}_{t}$ & Priority: runs the competition among the eligible commitments $I^{\vartheta}_{t}$ from their activation levels and priorities, returning the winning intention and the competition signal (Definition~\ref{def:gate}).\tabularnewline
$\Lambda$ & $\lambda_{i,t}$ & Behavior-readiness of each pressing commitment: its graded disposition to surface now, tested against the bar (Definition~\ref{def:gate}).\tabularnewline
$G_{\vartheta}$ & $\vartheta_{t}$ & Context-sensitive activation threshold: tightens with sensitivity and uncertainty, relaxes under urgency (Definition~\ref{def:gate}).\tabularnewline
$A_{i}$ & $\alpha_{i}(t)$ & Activation level of intention $i$: couples durable commitment to situated timing and enters the competition run by $Q$ (Definition~\ref{def:intention}).\tabularnewline
$V$, $V_{J}$ & values & Expected situated value for behavior selection; mandate-relative value (Definitions~\ref{def:selection} and~\ref{def:agency-loss}).\tabularnewline
$\Pi$ & $b_{t}$ & Behavior selector over the episode's candidate set $\mathcal{B}^{\phi}_{t}(i^{*}_{t})$ (Definition~\ref{def:selection}).\tabularnewline
$\mathcal{L}_{\rho}$ & $\rho^{evol}_{t+1}$ & Bounded revision of the evolvable profile within $\mathcal{R}^{\phi}_{t}$ (Definition~\ref{def:profile}).\tabularnewline
$\mathcal{U}$ & $S_{t+1}$, $\theta_{t+1}$ & One-step update of state and learned parameters from $S_{t}$, $\theta_{t}$, $o_{t+1}$, $b_{t}$ and $r_{t}$, composing $F_{X}$, $F_{U}$, $F_{Z}$, $\Psi$ and $\mathcal{L}_{\rho}$, within the admissible set $\mathcal{U}^{\phi}_{t}$ whenever a mandate is present, under its provenance, reversibility and approval requirements (Definitions~\ref{def:agentic-state} and~\ref{def:bounded-adaptation}).\tabularnewline
$\Gamma$ & $y_{t}$ & Governed traces: the delegator's window onto the agent's operation (Definition~\ref{def:mandate}).\tabularnewline
\bottomrule
\end{tabularx}
\end{table}

\section{Related Work and Conceptual Positioning}\label{sec:related}

This section positions the theory of governed, proactive agency in relation to neighboring frameworks. The claim is not that prior work lacks a theory of agency, but that it frames different parts of the problem. This paper focuses on how intelligent capacity becomes continuing, situated, significance-sensitive and governed behavior, especially in symbiotic agency, where the system acts under a continuing and user-specific mandate. Each subsection identifies what these frameworks explain, what this paper inherits from them and what remains outside their scope. Table~\ref{tab:proximity} summarizes the comparison and Appendix~\ref{app:comparison} gives the detailed version.

\subsection{Classical theories of agency}

Behaviorist psychology made behavior itself an object of scientific explanation, but its stimulus--response reduction explains elicited reaction, in the sense that the stimulus is the cause and the response its direct effect, rather than organized purposiveness \cite{Watson1913,Skinner1938}. Tolman's molar and goal-directed view of behavior is therefore important for this theory because it rightly shifts the unit of analysis from local reactions to organized purposive acts (Section~\ref{sec:background}). Cybernetics then showed how purposive behavior could be engineered through feedback and error correction \cite{Rosenblueth1943,Wiener1948}. Rational-agent models organized action around expected performance \cite{Russell1995}. BDI and planning theories explained how beliefs, desires and intentions stabilize deliberation and commitment over time \cite{Bratman1987,Rao1995}, with Cohen and Levesque formalizing intention as choice with commitment \cite{CohenLevesque1990}. Situated and embodied approaches added a further correction, aimed at the picture of intelligence as detached internal planning, from sensing, modeling, planning and then execute. Brooks built robots from layered reactive controllers that couple sensing directly to action, demonstrating that competent movement does not require a central explicit world model \cite{Brooks1986,Brooks1991}. Gibson argued that organisms directly perceive affordances, the possibilities for action the environment offers, so that perceiving and acting form one continuous loop \cite{Gibson2014}. Suchman showed that human plans do not execute like programs. Plans function as rough resources while actual behavior is improvised against the actual circumstances \cite{Suchman1987}. In the same line, Agre and Chapman showed that most everyday activity is competent improvisation, continually re-decided from what the current situation presents \cite{AgreChapman1987}. The common point is that competent behavior is produced in ongoing interaction with the environment rather than computed in advance. In other words, a system that stops perceiving while it deliberates falls behind the world it is about to act in.

Each of these research lines contributes something the presented work preserves. Behaviorism gives behavior as the observable unit. Tolman gives purposive organization. Cybernetics gives feedback and error correction. Rational-agent models give evaluation against expected performance. BDI gives durable commitment. Situated approaches give continuous coupling to the environment. The formal vocabulary in Section~\ref{sec:formal} is built from these materials.
What remains missing is the activation problem introduced in Section~\ref{sec:introduction}. None of these frameworks provides a user-relative layer that decides whether, when and in what mode capability should become behavior. The gaps are complementary. Intention theory explains why commitments persist, but not when a standing commitment should surface (Definition~\ref{def:intention}). Situated approaches explain coupling to the environment, but not mandate-sensitive restraint. Staying coupled does not by itself explain when the agent should hold back. Feedback control regulates behavior once a target is active, but not whether that target should have become active at all. Appraisal, readiness and the activation gate supply this missing layer on top of the inherited materials (Definitions~\ref{def:appraisal} and~\ref{def:gate}).

\subsection{Computational appraisal and affective computing}\label{subsec:related-appraisal}

Appraisal has long been used to connect events with goals, needs, beliefs, emotion, coping and action tendencies. Picard established affective computing as a field concerned with recognizing, representing and generating affect \cite{Picard1997}. The OCC model gave computer science a structured way to evaluate events, agents and objects in relation to goals, standards and attitudes \cite{OrtonyCloreCollins1988}. Computational architectures then implemented appraisal as part of the loop between situation assessment, internal state, emotion and behavior \cite{Hudlicka2004,GratchMarsella2004}.

The relation between the gate dimensions in Definition~\ref{def:appraisal} and canonical appraisal theory is close, but not one-to-one. Scherer's model treats emotion as arising from a sequence of appraisal checks, including novelty, intrinsic pleasantness, goal relevance and conduciveness, coping potential and norm compatibility \cite{Scherer2001}. Several dimensions in this paper have clear analogues. Urgency and potential harm correspond broadly to goal relevance and obstructive conduciveness. Action risk is close to coping-related appraisal of what a response may cause. Social and emotional sensitivity parallels norm and self-compatibility checks. Uncertainty plays a role similar to novelty and predictability.
Two dimensions, however, come into play only under delegation. Interruption cost evaluates the cost of claiming another person's attention and it is zero when there is no principal whose attention could be claimed. Permission proximity evaluates how close a candidate behavior is to an authority boundary. Its trust-boundary part likewise requires a principal, while proximity to the agent's own constraint boundaries applies in every form. These dimensions are not standard components and their counterparts in human appraisal theory are partial. Scherer's norm compatibility check weighs the social cost of acting and the closeness of a response to a standard, which resemble interruption cost and permission proximity in form. What has no counterpart is whose cost and whose boundary they weigh. Nevertheless, for the current work they are needed, because the appraiser here is an agent acting for someone else, under someone else's authority. Interruption cost is the cost of claiming the principal's attention and permission proximity is distance to a boundary of delegated authority, not to a norm the appraiser holds for itself. This is why the framework does more than rename emotion theory. It adapts appraisal to the governance problem of delegated and symbiotic agency.

The scope of the claim is therefore narrow. This paper does not model the production, experience, recognition, or expression of emotion. It uses appraisal functionally, as a way to assign situated significance to evidence relative to intent, the user's inferred condition, the agent's operational limits, expected consequences and the mandate. Conation names the resulting orientation toward behavior, such as act, monitor, ask, wait, defer, preserve, escalate, revise, inhibit, or refrain. The question is whether this structured appraisal adds something beyond a single value model. The delegation-specific dimensions above, together with the measurement and design roles of the gate in Sections~\ref{sec:measurement} and~\ref{sec:design}, are the reason it does.

\subsection{Alternative formal accounts of agency}\label{subsec:related-formal}

Several formal frameworks already describe how agents act under uncertainty, learn from feedback, or select behavior over time. POMDPs model sequential decision-making under partial observability through belief states, transition and observation models, rewards and policies \cite{Kaelbling1998}. Their belief states are the counterpart of the epistemic representations of Definition~\ref{def:core-representation}. Reinforcement learning organizes behavior around expected cumulative reward \cite{SuttonBarto2018}, while model-based methods and world models go a step further than basic RL. Instead of just learning which action tends to work well, they also try to predict what will happen next in the environment, based on the action taken. In other words, the agent builds a rough mental model of ``if I do this, here's what I expect to happen,'' and uses that prediction to help decide what to do \cite{HaSchmidhuber2018}. Reward-organized selection is the counterpart of the situated value $V$ and the constrained selection of Definition~\ref{def:selection}, while the action-conditioned prediction of world models is the counterpart of the expected situated consequence $\widehat{\Delta}_{t}(b)$ of Definition~\ref{def:appraisal}. From an active inference perspective, perception and action are treated as two routes to the same objective. The focus here is to minimize free energy under a generative model, with perception revising beliefs and action changing the world to fit those beliefs. Planning extends this logic prospectively through expected free energy, which agents minimize when selecting a policy. Because this quantity decomposes into an epistemic term (the value of resolving uncertainty) and a pragmatic term (the value of reaching preferred outcomes), instrumental and information-seeking behavior emerge from a single minimization rather than two separate mechanisms \cite{ParrPezzuloFriston2022,DaCosta2020ActiveInference}. The uncertainty-reducing part of expected free energy helps explain why waiting for evidence can have value, which is the role the threshold $\vartheta_{t}$ plays in Definition~\ref{def:gate}. The outcome-directed part plays the corresponding role for the situated value that readiness tracks.

Other formal mechanisms anticipate specific parts of the vocabulary. The options framework in hierarchical reinforcement learning formalizes temporally extended behavior \cite{SuttonPrecupSingh1999}. An option has an initiation condition, an internal policy and a termination condition. This is close to the structure of the behavioral episode in Definition~\ref{def:behavior}. Shielding in safe reinforcement learning formalizes hard constraint enforcement at selection time \cite{Alshiekh2018}. At each step, the shield filters or overrides the selected action so that the executed behavior remains within the specification. This corresponds to the action firewall in Section~\ref{sec:design} and to the permitted set $\mathcal{B}^{\phi}_{t}$ in Definition~\ref{def:selection}. Constrained MDPs provide a different device \cite{Altman1999}. They constrain expected costs over whole trajectories, so an individual action may still contribute to a violation as long as the aggregate remains within the bound. This aggregate logic is similar to that used by the graded, severity-weighted score in Definition~\ref{def:aligned} for soft violations. Hard constraints are different. They are enforced at each decision, as in shielding. Finally, the allo-interoceptive pipeline in Definition~\ref{def:allo} can be implemented as belief maintenance over an unobserved variable, as in POMDP belief states. The contribution of this paper is not the inference machinery itself, but the asymmetry claim and the calibration obligation attached to it.

The theory developed in this work is therefore not a replacement optimization calculus for the approaches introduced above. Its purpose is to make explicit the structural vocabulary needed for user-relative and delegated agency. This includes three distinct perspectives on state, namely the world, the inferred user and the agent's own condition. It also includes durable intent, multidimensional significance, behavior-readiness, an explicit mandate, authority boundaries and restraint as a first-class behavior. In a POMDP or active-inference implementation, these elements could be realized as state variables, preferences, constraints, priors, or policy factors. The claim is not that they require a unique computational substrate. The claim is that they should remain relevant, separately represented and governed, so that the mandate is not just silently absorbed into a reward function and the consent boundary is not reduced to a learned preference.

\subsection{Assistance, preference uncertainty and corrigibility}

Assistance games \cite{Russell2019HumanCompatible,Shah2020Assistance} are a directly relevant alignment point for the delegated part of the theory developed in this work. Cooperative inverse reinforcement learning (CIRL) formalizes assistance as a cooperative partial-information game between a human and a robot \cite{HadfieldMenell2016CIRL}. Both are rewarded according to the human's reward function, but the robot does not initially know its parameters. Human behavior therefore becomes evidence about what the human values and assistance depends on maintaining uncertainty rather than assuming that the objective is already fixed. The Off-Switch Game develops the same lesson for corrigibility. When the robot is uncertain about the human's utility and treats the human's shutdown decision as evidence, preserving the off-switch can become instrumentally rational rather than something the robot has reason to disable \cite{HadfieldMenell2017OffSwitch}.

This work also reflects this lesson. A symbiotic agent should remain uncertain about the principal's condition, preferences and interpretation of the situation and it should preserve channels through which the principal can correct, redirect, or stop it. This is reflected in the calibrated user-state estimate of Definition~\ref{def:allo}, the bounded revision of profiles and the treatment of correction as evidence rather than failure.

The difference is that assistance games primarily formalize uncertainty over human objectives and the informational value of human behavior. This work, however, places that problem inside a broader theory of mandate-bound agency. A mandate is not only an inferred preference. It records authority, scope, consent and accountability. The same distinction applies to what the agent infers about the principal. Assistance games focus on the human's preferences, or the reward function that the robot does not initially know. Allo-interoception (Definition~\ref{def:allo}) includes preferences, but is broader. It also estimates the principal's current condition, including attention, load, availability and affect, from external evidence. This estimate remains uncertain, calibrated and revisable as the situation changes. Preferences indicate what the principal tends to value. Condition helps determine whether assistance is appropriate now, when it should surface and in what mode. The theory also makes explicit the sensorium, perceptual permissions, interruption cost, behavioral timing, standing commitments, restraint and mandate-relative agency loss. These are complementary additions. Assistance games explain why uncertainty and corrigibility matter. The theory developed here explains how those ideas fit into a continuing, governed and symbiotic agent that must decide whether, when and in what mode to act.

\subsection{Mixed-initiative interaction and interruption}\label{subsec:related-mixed-initiative}

Mixed-initiative interaction, a line of research in human--computer interaction (HCI), gives an early and direct formulation of the activation problem in its timing dimension. The question is whether a system should take initiative, when it should do so and in what mode it should proceed. Horvitz frames this as a decision under uncertainty about the user's goals and attention, where autonomous action must be weighed against the cost of interruption, the value of dialog and the option of deferral \cite{Horvitz1999}. Attention-sensitive alerting develops the same idea for notifications, treating alert timing as a decision-theoretic problem shaped by message criticality, expected delay, inferred attentional state and interruption cost \cite{HorvitzJacobsHovel1999}. Iqbal and Bailey then show empirically that notification timing matters. Deferring notifications to task breakpoints can reduce frustration and improve reaction time and the appropriate breakpoint depends partly on the relevance of the notification content \cite{IqbalBailey2008}. Interface agents developed the assistant side of the same problem. Maes's agents learned a user's habits from observation and feedback and took initiative on routine tasks. This led to the early raising of the questions of how much initiative an assistant should take and when its user's trust is warranted \cite{Maes1994}. Mixed-initiative interaction then positioned itself between such agent autonomy and direct user control, which is one reason its timing calculus became the sharpest early statement of the activation problem.

This line of work is therefore the closest prior formulation of the interruption side of the problem. The interruption-cost dimension $\sigma^{intr}$  and the context-sensitive gate $\vartheta_{t}$ in Definition~\ref{def:gate} build directly on it. In the terms of this paper, mixed-initiative systems solve the activation problem for a candidate service or notification within a bounded interaction. This work generalizes the problem along three axes. First, scope: the mandate in Definition~\ref{def:mandate} is standing and may persist across sessions, contexts and delegation chains. Second, the behavior space: the agent does not only choose between acting and deferring, but may ask, monitor, remember, wait, escalate, inhibit, or deliberately refrain under the same appraisal machinery (Definition~\ref{def:selection}). Third, governance: authority, consent, bounded adaptation and mandate-relative agency loss are central here, while they were not the main object of mixed-initiative work (Definitions~\ref{def:mandate} to \ref{def:bounded-adaptation}). The generalization is therefore one of problem formulation, not a rejection of the timing calculus. On timing itself, this paper builds directly on mixed-initiative interaction.

This is why mixed-initiative interaction is central to the present argument. It already shows that intelligence is not enough if the system cannot decide whether initiative is appropriate. The activation problem developed in this paper extends that insight from isolated services and notifications to continuing, mandate-bound and symbiotic agency.

\subsection{Delegated and multi-agent agency}

Principal--agent theory studies delegation under asymmetric information and potentially divergent interests \cite{Ross1973,JensenMeckling1976,Eisenhardt1989}. Its vocabulary transfers naturally to artificial delegation. Agency loss in Definition~\ref{def:agency-loss} is the AI-native analogue of residual loss. The reciprocal observability asymmetry in Section~\ref{subsec:reciprocal-asymmetry} is information asymmetry in both directions. Monitoring and bonding reappear as governed traces, explanations, constraints and records of what the agent was permitted to do.
The artificial case also differs in two important ways:
\begin{itemize}
\item
First, divergence does not require psychological self-interest. It may arise from misspecification, incomplete context, stale memory, policy conflict, or incorrect mandate translation (Section~\ref{subsec:delegated-theory}).
\item
Second, the agent is not fixed at the moment of delegation. Learning can change its profile, memory, skills and future intervention patterns after authority has been transferred. Bounded adaptation (Definition~\ref{def:bounded-adaptation}) is introduced to govern this change.
\end{itemize}
Delegation also has a native theory within multi-agent systems research. Castelfranchi and Falcone analyze it as one agent relying on another for an action or a goal, distinguish its levels, from mere reliance to agreed adoption of the delegated concern and ground it in trust \cite{CastelfranchiFalcone1998}. Recent AI work makes the connection explicit. Rauba et al. formulate multi-agent AI systems as principal--agent problems, linking information asymmetry to system-level agency loss \cite{Rauba2026}. Liu et al. empirically study eight multi-agent frameworks and identify coordination and maintenance failures as a material category of practical issues \cite{Liu2026}. This paper builds on that connection by treating the mandate as a distinct governed object (Definition~\ref{def:mandate}). The mandate records purpose, scope, authority, duration and delegability and constrains intention formation, behavior selection, adaptation and downstream delegation (Definitions~\ref{def:delegation} and~\ref{def:bounded-adaptation}). The central translation problem is therefore how the mandate becomes intent, policy and profile. This is one of the primary points where delegated agency is preserved or lost.

Seen in light of this work principal--agent theory and multi-agent AI research explain different parts of the problem. Principal--agent theory gives the vocabulary of delegation, such as interests may diverge, the principal cannot observe everything the agent does and monitoring and bonding reduce but do not eliminate loss. Multi-agent research instead shows the practical failure modes. Agents may duplicate work, contradict each other, corrupt shared state, or create systems that are difficult to maintain. What remains missing is the governed object that anchors artificial delegation. In this work, that object is the mandate, together with the formal structures that translates it, preserves authority, records behavior, coordinates agents and bounds learning. Each part corresponds to one term of the agency-loss attribution in Definition~\ref{def:agency-loss}.
\begin{itemize}
\item
\emph{Translation.} The mandate must be translated into intentions, policy and profile. Meaning can be lost in this translation. That loss is $L^{mandate}_{t}$.
\item
\emph{Authority.} When an agent delegates onward, the sub-agent may not receive more authority than the delegating agent was granted. Authority can shrink along the chain, but it cannot grow (Definition~\ref{def:delegation}). Violations of this containment are $L^{authority}_{t}$.
\item
\emph{Information.} The principal sees the agent through governed traces, not through direct access to its internal condition. Governance can therefore decide only what the traces record. The loss created by this limited view is $L^{information}_{t}$.
\item
\emph{Coordination.} Agents that share tasks, tools, or memory can coordinate badly. They may duplicate work, contradict one another, or contaminate shared state, which is the kind of failure Liu et al. document. This is $L^{coordination}_{t}$.
\item
\emph{Adaptation.} The agent continues to learn after authority has been transferred. Its profile, memory, skills and future patterns of intervention may therefore change over time. Adaptation that drifts outside consent, policy, or mandate boundaries is $L^{adaptation}_{t}$.
\end{itemize}
On this view, information asymmetry and multi-agent coordination failure are not separate problems. They are different sources of mandate-relative agency loss. Defining that loss makes delegated and multi-agent agency measurable against the mandate, rather than against task success alone.

\subsection{LLM agents and long-horizon evaluation}\label{subsec:related-llm}

The recent turn toward LLM agents is the immediate practical setting for this paper. It has produced a rapid succession of systems and benchmarks, for the most part organized around planning, memory, tool use, action and long-horizon task completion. Surveys, therefore, tend to describe LLM-based agents through these components \cite{Wang2024LLMAgentSurvey,Xi2025}. This is precisely the framing that this paper argues is insufficient on its own. Components explain what an agent can use, but not whether, when and in what mode its capabilities should become behavior.

Benchmarks now cover several kinds of long-horizon execution. Software and terminal work is tracked by SWE-bench Pro and Terminal-Bench 2.0 and SWE-agent shows how the agent--computer interface shapes such work \cite{Deng2025SWEBenchPro,Merrill2026TerminalBench,Yang2024SWEAgent}. Computer use is tracked by OSWorld 2.0, which targets long-horizon workflows whose 108 tasks take people a median of about 1.6 hours \cite{Yuan2026OSWorld2}. Web tasks are tracked by WebArena for functional success on realistic sites and by BrowseComp for persistent multi-hop research \cite{Zhou2023WebArena,Wei2025BrowseComp}. Multi-environment decision-making is tracked by AgentBench \cite{Liu2023AgentBench}. Tool use under explicit policies is tracked by $\tau^{2}$-bench, where both agent and user act and policy adherence is scored \cite{Barres2025Tau2}. Sustained operation is tracked by Vending-Bench and its year-long successor Vending-Bench 2, which score coherence across thousands of steps rather than any single decision \cite{Backlund2025VendingBench,AndonLabs2025VendingBench2}. Along a different axis, METR's task-completion time horizon measures the task duration, which reflects the time it would take a human expert to complete a certain task, at which an agent is predicted to succeed with a given level of reliability. The 50\% time horizon, for example, is the duration at which the agent is predicted to succeed half the time \cite{Kwa2025TimeHorizon}. Safety is tracked separately by benchmarks such as AgentDojo, which test whether injected instructions can redirect an agent \cite{Debenedetti2024AgentDojo}. Recent work also extends the same execution-oriented contract to agent serving and operation over longer horizons \cite{Sui2026Parallelizing,Zheng2026OneDayAgent}.

Taken together, this evaluation practice measures execution increasingly well. Its default structure, however, remains an already activated episode. Delegation has occurred, the task is given, activation is assumed and restraint appears at most as a failure to act rather than as a behavior that can be correct. These benchmarks rarely ask whether a standing commitment should have surfaced, whether timing respected interruption cost, whether authority remained contained across a delegation chain, or whether withheld action was accompanied by monitoring, memory, escalation, or deliberate restraint.

Recent proactive-assistance benchmarks move closer to the activation problem. ProAgentBench separates when to assist from how to assist, making the timing decision an explicit object of evaluation and using precision as a proxy for interruption cost \cite{Tang2026ProAgentBench}. It does so over real user sessions rather than synthetic tasks and it finds that long-term memory and historical context improve timing prediction. This supports the emphasis placed here on memory and continuing context. ProactBench similarly evaluates whether an agent acts on needs the user did not explicitly state \cite{Harfi2026ProactBench}. These benchmarks address the timing dimension of the activation problem directly and are the closest current measurements of it.

Their limits are therefore instructive. Timing is still scored largely as an isolated classification or assistance decision, rather than as one part of a governed behavioral episode. Recent work on proactive coding agents notes that no benchmark yet evaluates monitoring, relevance, timing, framing and silence together \cite{BuiEvangelopoulos2026Proactivity}. Current benchmarks also do not evaluate whether behavior occurred under a standing mandate, whether authority remained contained across a delegation chain, whether perception was policy-gated during operation, or whether the agent's estimate of the principal's condition was calibrated. Privacy-compliant data collection is important, but it is not the same as a runtime consent gate.
The evaluation framework in Section~\ref{sec:measurement} is intended to complement task-completion and proactive-assistance evaluation with these missing dimensions. The signatures in Section~\ref{sec:signatures} make the difference explicit. Two systems with identical task scores may differ entirely in proactivity, restraint and constraint posture. In the taxonomy of Section~\ref{sec:introduction}, most current evaluation settings are task-bounded delegation rather than symbiotic agency. The delegation is bounded by the task and the coupling ends with the episode. ProAgentBench comes closest because it evaluates proactive timing over continuing workflow context, but evaluating proactive assistance is not the same as realizing symbiotic agency. The mandate, the authority boundary, the gated sensorium, calibrated allo-interoception, bounded personalization and restraint as a logged behavior remain outside its scope. The current landscape therefore measures execution extensively, timing recently and partially and the governance of proactive behavior only weakly. This paper targets that unmeasured part: whether behavior under a standing mandate should occur, when, in what mode, within what authority and with what record.

\subsection{Comparative summary}

The previous subsections introduced and discussed several bodies of work to highlight some of the concepts this paper builds on and the limits of each, in relation to the theory developed here. Together, they show that many elements of governed, proactive agency already have important antecedents. Notwithstanding, these elements were developed in different research areas and for different purposes. Section~\ref{sec:signatures} turns the vocabulary of Section~\ref{sec:formal} into a method for scoring approaches and systems and Table~\ref{tab:proximity} applies it to the approaches discussed throughout this section, positioning each of them relative to the proposal developed here. It shows which elements are already explicit, which remain partial or implicit and where the proposed vocabulary adds the structures needed for autonomous, delegated and symbiotic agency.

In light of the comparison developed above, the contribution of this work is not a single isolated mechanism, but the integrated structure in which the mechanisms are organized. Several elements have clear antecedents. BDI clarifies the role of durable commitment. POMDPs formalize belief over hidden state. Affective computing and appraisal theory explain how situations acquire significance. Mixed-initiative interaction shows why timing, interruption and deferral matter and interface agents show why learned personalization and earned initiative matter. Principal--agent theory and assistance games clarify delegation, preference uncertainty, information asymmetry and corrigibility. LLM-agent harnesses make permissions, logs, tools and long-horizon execution operational. While these elements have been addressed separately by the different bodies of work introduced before, the current work brings them together around the thesis that agency is an organizational property of behavior across time, rather than a list of components. Three commitments follow from this organization.
\begin{itemize}
\item
\emph{Behavioral episodes and activation as the unit of analysis.} The observable unit of agency is not the isolated action, but the behavioral episode. This episode may include external action, inquiry, monitoring, deferral, inhibition, or deliberate restraint. The central question is therefore whether behavior should occur, when it should occur and in what mode. This question is answered through significance-gated and context-sensitive activation, not merely through stimulus, instruction, or task availability.
\item
\emph{The mandate as the object that anchors delegated behavior.} When an agent acts on behalf of a principal, its behavior must remain answerable to explicit authority and scope. The mandate grounds consent-gated perception, governed traces, bounded adaptation and agency loss measured relative to what was authorized. It is what prevents delegated agency from collapsing into unbounded autonomy, proxy optimization, or unaccountable intervention.
\item
\emph{Continuing coupling to the principal in symbiotic agency.} Symbiotic agency requires more than delegation. It requires a standing mandate, continuing coupling through allo-interoception, calibrated estimation of the principal's condition and personalization within the mandate's bounds. This coupling is not intended to replace the principal's judgment. Its purpose is to augment the principal's capabilities while keeping the agent answerable to the conditions under which authority was delegated.
\end{itemize}

These three commitments follow the same layering as the introduced taxonomy. The first applies to agency in general and therefore to autonomous, delegated and symbiotic agency. The second applies when agency is delegated, because behavior must then remain answerable to a mandate. The third applies when delegated agency becomes symbiotic, because the agent remains continuously coupled to the principal and must personalize its interpretation of significance within the mandate's bounds. Table~\ref{tab:proximity} shows this structure comparatively. Each prior approach fills part of the baseline or parts of the form-specific groups, but none treats the delegated and symbiotic elements together as first-class objects.
 
\section{Agency Signatures}\label{sec:signatures}

\subsection{Agency Signature Classification Method}\label{subsec:instrument}

Section~\ref{sec:formal} defined the elements of agency and Definition~\ref{def:aligned} specified which elements each form requires. This section uses that vocabulary as a classification instrument. Because agency is layered, an approach or system cannot be characterized by a single agency ladder. Hence, the relevant question is not just how agentic the approach or system might be, but which structures of agency it instantiates. We call this characterization its \textbf{agency signature}. For an approach or system, the signature is equivalent to its row in Table~\ref{tab:proximity}. It records which elements are present as first-class objects, which are partial or implicit and which are absent. It also records the coverage of each group, what triggers behavior and the resulting placement in the taxonomy.

The columns of Table~\ref{tab:proximity} follow the element groups defined in Definition~\ref{def:aligned}. The first group is the shared baseline of agency, common to autonomous, delegated and symbiotic agency and it is scored separately so that the elements that distinguish the forms do not hide the structure all of them share. Within the baseline, the tables mark two sub-groups. The agenthood sub-group contains the five elements of Definition~\ref{def:agenthood} and the first stage of placement, defined below, reads only them. The activation sub-group carries the activation problem itself, whether behavior should occur, when and in what mode, including restraint. The observable record, the remaining baseline column, stands outside both sub-groups.

The activation columns are scored in every signature even though placement does not read them. They belong to the shared structure of complete agency, which every reference form realizes in full and they are where the compared approaches differ most, so dropping them would hide exactly the dimension this paper is about. As Definition~\ref{def:agenthood} states, the activation problem reaches the agenthood minimum through significance appraisal, its evaluative core, while the gate records whether appraisal can open, time and shape episodes on the agent's own initiative. Empty gate columns therefore mark a commanded agent, not a non-agent. The activation columns decide what kind of agent a system is and how far its agency extends, never whether it is one and the observable record grades answerability in the same way.

The three groups that follow correspond to the taxonomy built on top of this baseline. The autonomous group contains goals, constraints and adaptation without a principal. The delegated group adds the mandate, authority containment, governed traces, consent-gated perception and bounded adaptation. The symbiotic group extends delegation through a standing mandate, continuing coupling with calibrated allo-interoception and personalization.
The groups are cumulative. In that sense, autonomous agency requires the baseline and the autonomous group. Delegated agency requires the baseline and the delegated group. Symbiotic agency requires the baseline, the delegated group and the symbiotic group.

The agency signature is therefore computed according to the three steps presented below.
\begin{itemize}
\item \emph{Element scores.} Each element $k$ listed in Definition~\ref{def:aligned} receives a score $e_{k}$ between $0$ and $1$. On the three-level scale used in the tables, $e_{k}=1$ is used when the element is a first-class object of the approach or system: it is named, it has its own representation or mechanism and it takes part in selecting behavior. $e_{k}=1/2$ when the element is only partial or implicit, for example when it is folded into another construct, optional, handled ad hoc, or not carried across episodes. $e_{k}=0$ when the element is absent. For significance appraisal, partial means more than a check on the current input. The cell receives $1/2$ only when two conditions hold.
\begin{enumerate}
\item
\emph{Enrichment.} The evaluation must enrich the current input with what the system keeps beyond it. For example, it may draw on standing intent, cross-episode memory, a maintained representation of the task, world, or principal, or the expected consequence $\Delta_{t}(b)$ of a candidate behavior (Definition~\ref{def:appraisal}). Judging the input in isolation is not appraisal. Content merely retrieved into the input widens the input; it does not enrich the evaluation.
\item
\emph{Gradedness.} The evaluation must weigh how much the situation matters, rather than test a fixed rule.
\end{enumerate}
Input-only refusal, safety, or clarification checks are therefore scored under constrained selection or behavior-mode, not under significance appraisal. Keeping the threshold in memory is not enough. A bare comparison against it does not enrich the evaluation or weigh situated significance. The two requirements are independent. A rule system may draw on maintained representations and still fail gradedness. For example, it may escalate a message because the sender is marked as important for this principal and the calendar shows an active meeting. This satisfies enrichment, because what the system keeps gives the message its meaning for this principal. But if it only tests a fixed condition, it does not weigh significance in context beyond that fixed condition and therefore fails gradedness.

\item \emph{Group coverage.} For each of the four groups, coverage is the average of the element scores in the group, $\frac{1}{|\mathcal{G}|}\sum_{k\in\mathcal{G}}e_{k}$, reported as a percentage. A group whose elements are all first-class scores 100\%, a group whose elements are all partial scores 50\%.
\item \emph{Placement.} Placement is decided in two stages and both stages read only the presence or absence of named elements.
\begin{itemize}
\item \emph{Stage 1, agenthood.} A system is an agent when the five elements of Definition~\ref{def:agenthood} are all present, whether as first-class or partial structures. These are standing intent, situated perception, significance appraisal, constrained selection and feedback with revision. A system that lacks any one of them receives no form of agency. Nevertheless, the mandate vocabulary can still describe it. If a mandate is present, it may be called a \emph{delegated system}. A delegated system is a system that holds an authorization record without the organization required for agenthood. Thermostats, scheduled automation rules and chatbots typically fall here. If no mandate is present, it may be called an \emph{unmandated system}. The four labels cover every system. Above the agenthood minimum threshold it is either an autonomous or a delegated agent; below it, either an unmandated or a delegated system.
\item \emph{Stage 2, form.} For agents, meaning the systems that have passed Stage 1, the form records the authority relationship and the most specific applicable form is chosen. If no explicit mandate is present, the agent is autonomous. If an explicit mandate is present, whether first-class or partial, the agent is delegated. If a mandate is present and the three conditions of Definition~\ref{def:symbiotic} are also present as first-class objects, the delegated agent is symbiotic.
\end{itemize}
A delegated placement covers the agents without the full symbiotic conditions, described throughout this work as task-bounded. Placement still does not depend on coverage. Re-scoring the other elements can change coverage but not placement. A system that operates under more than one mandate configuration is placed per configuration. Because Stage 1 requires every defining element, a single missing one can decide placement. Hence, if one of those is classified very close to the partial--absent boundary, it should be marked as \emph{placement-critical}. The scoring report should state the adopted placement and, where useful, the alternative one. In Table~\ref{tab:signatures}, this applies to the chatbot. The row describes an assistant without persistent memory or a maintained task or user model. Its refusals, qualified answers and clarifying questions are genuine choices, but they depend on the current input alone. They therefore fail the enrichment requirement for significance appraisal, so the system is placed as a delegated system rather than a delegated agent. The agentic chat assistant row is exactly the configuration that repairs this and it places as a minimal delegated agent. This is a difference between configurations, not a contradiction in the placement rule. Such cases belong to the reliability protocol of Open Problem~10 (Section~\ref{sec:open-problems}).
\end{itemize}

The distinction between coverage and placement is visible in Table~\ref{tab:proximity}. For example, the LLM-agent harnesses row scores significance appraisal as absent. It therefore fails the agenthood stage, because significance appraisal is one of the five elements required by Definition~\ref{def:agenthood}. The mandate vocabulary can nevertheless still be used to describe such a system. Since a partial mandate is present but the agenthood minimum is not met, the placement is a delegated system rather than a delegated agent. The tool-using coding agent in Table~\ref{tab:signatures} is different. It passes the agenthood stage because all five agenthood elements are at least partial. At the form stage, it is placed as a delegated agent because an explicit mandate is present. It is not symbiotic, because the standing mandate and continuing coupling conditions are absent.

The element scores are derived from different sources. Approaches are scored from their literature, while systems are scored from their specification and record. Some elements are visible in behavior alone. Others, such as consent-gated perception, bounded adaptation and calibration of the principal-state estimate, leave no direct footprint in behavior. They must be established from the mandate, configuration, tests and governed traces. When only behavior is available, the record must expose the defining elements of placement. A system whose record does not expose these elements can be described, but it cannot be placed reliably.

The behavior classes in Definition~\ref{def:classes} should be kept separate from this classification method. They do not determine the form of agency. They describe how the shared baseline appears in behavior. Reactive behavior shows situated perception at work. Proactive behavior starts new episodes from appraised significance rather than from instruction. Constrained behavior selects within the permitted set, including restraint. Adaptive behavior expresses feedback and revision. Social, deliberative and reflective behavior express sensitivity to other agents, evaluated alternatives and the agent's own condition. Every form can exhibit every behavior class. The behavior classes therefore describe how a system behaves within its form, not which form it belongs to.

As presented, these quantities say which form a system belongs to and how completely it instantiates the structures of agency. Baseline coverage indicates how many of the shared baseline structures it instantiates. Form-specific coverage indicates how completely it realizes the form in which it is placed. Read in this way, two systems with similar tool access can have very different signatures, because they instantiate different structures. A low coverage is also not necessarily a failure. A delegated agent with low symbiotic coverage is a task-bounded delegated agent, not a failed symbiotic one and its signature shows exactly which structures it would still have to add to become symbiotic. It is nevertheless important to highlight that the signature itself only classifies. Measuring an actual agency score is the subject of Section~\ref{sec:measurement} and it starts from the same signature. There, each score is replaced by a degree measured from the system's record and Definition~\ref{def:aligned} combines these degrees into the capability score of the placed form. Classification and measurement therefore read from the same signature.

\subsection{Agency Signatures of Related Approaches}\label{subsec:sig-approaches}

The previous subsection introduced the agency signature as a classification instrument for comparing approaches and systems. Table~\ref{tab:proximity} applies that instrument to the approaches discussed in Section~\ref{sec:related}. The final column records what typically triggers behavior in each approach and the last three rows give the paper's own autonomous, delegated and symbiotic forms as reference cases.

One aspect should be kept in mind when reading the table. Table~\ref{tab:proximity}, as opposed to Table~\ref{tab:signatures}, does not assign placement. Its rows are grouped by the form each body of work addresses, the form whose problems it theorizes, which is not the same as where a system built from that work would be placed. The reason is that an approach is a body of theory rather than a deployed system. It can address a form in depth without instantiating the form's defining elements. Placement, as defined in Section~\ref{subsec:instrument}, is read from the defining elements and is applied to systems in Table~\ref{tab:signatures}. 

For BDI, coverage and placement point in the same direction. A system built only from BDI structure passes the agenthood minimum and has no mandate, so it is placed as an autonomous agent. The approaches that address delegation are different. They contain some delegated elements, such as mandate, authority, traces, or preference uncertainty, but they do not all satisfy the agenthood minimum. Principal--agent theory lacks situated perception, LLM-agent harnesses lack significance appraisal and assistance games lack constrained selection. A system built only from one of these approaches is therefore not a delegated agent in this taxonomy. It is a delegated system, because a mandate-like structure may be present, but the organization required for agenthood is incomplete. The sharpest cases are mixed-initiative interaction and interface agents.

One addresses the symbiotic region through timing, interruption and continuing context, the other through learned personalization. Their symbiotic coverage is the highest among the referenced approaches, but neither contains an explicit mandate. The goals they serve are inferred or learned rather than authorized, so the minimal test of Definition~\ref{def:mandate} fails. User-set modes, thresholds and autonomy levels tune a service the system already provides. They regulate how initiative is exercised and no record grants the system a task on a principal's behalf. Furthermore, control is exercised per action rather than through standing delegation. Both also score constrained selection as absent, so a system built only from either structure fails the agenthood minimum of Definition~\ref{def:agenthood}. It addresses the symbiotic region without being an agent under this taxonomy and with no mandate it is an unmandated system, the remaining case of the classification. Addressing a form and satisfying its defining elements are different judgments and the table reports the first while the placement rule decides the second.

As presented, Table~\ref{tab:proximity} shows how far each approach goes relative to the vocabulary developed here. Classical and formal approaches mainly populate the shared baseline and the autonomous form. They provide important structure for organizing behavior, but they do not provide the mandate and governance layer. Principal--agent theory, assistance games and LLM-agent harnesses move toward the delegated form because they address delegation, uncertainty, authority, traces, or task-bounded evaluation. However, they do not provide the full mandate-bound structure developed here. Mixed-initiative interaction and interface agents come closest to the symbiotic form and they are complementary rather than identical. Mixed-initiative interaction makes the user's current condition a first-class object and personalizes only partially, while interface agents make personalization first-class and model user condition only partially. Together they anticipate part of the coupling apparatus of symbiotic agency. Nevertheless, neither provides the full combination of a standing mandate, a policy-gated sensorium, calibrated allo-interoception, bounded personalization and governed traces in which restraint itself is recorded. The proactive-assistance benchmarks in Section~\ref{subsec:related-llm} evaluate this same region. They are evaluation instruments, not theories or architectural frameworks and are therefore not scored as rows in the table. 

Overall, the comparison supports two conclusions. First, none of the compared approaches combines the two conditions that matter most for this paper. That behavior is triggered by appraisal through an activation gate and that when agency is delegated, that behavior remains answerable to an operative mandate. Second, no prior approach fills every group of the signature. Each one covers part of the baseline and parts of the form groups, but none provides the delegated and the symbiotic elements together. 

Both conclusions are nevertheless expected. These approaches provided the conceptual foundation for important theoretical elements of this work and the purpose of the comparison is not to find them lacking, but to show how far each maps onto the proposed theory of governed proactive agency. The detailed comparison, including the typical formulation of each approach on each dimension, is given in Appendix~\ref{app:comparison}.

\afterpage{%
\begin{landscape}
\begin{table}[p]
\centering
\footnotesize
\caption{Agency signatures of the related approaches of Section~\ref{sec:related}, scored as described in Section~\ref{subsec:instrument}. The last three rows are the theory's own forms, as reference.}
\label{tab:proximity}
\setlength{\tabcolsep}{3.4pt}
\begin{tabular}{>{\raggedright\arraybackslash}p{3.8cm}*{21}{c}cccc>{\raggedright\arraybackslash}p{4.4cm}}
\toprule
 & \multicolumn{9}{c}{General: baseline of agency} & \multicolumn{3}{c}{Autonomous} & \multicolumn{5}{c}{Delegated} & \multicolumn{4}{c}{Symbiotic} & \multicolumn{4}{c}{Coverage (\%)} & \tabularnewline
\cmidrule(lr){2-10}\cmidrule(lr){11-13}\cmidrule(lr){14-18}\cmidrule(lr){19-22}\cmidrule(lr){23-26}
 & \multicolumn{5}{c}{\scriptsize\itshape agenthood} & \multicolumn{3}{c}{\scriptsize\itshape activation}\tabularnewline
\cmidrule(lr){2-6}\cmidrule(lr){7-9}
Approach & \rotatebox{90}{Standing intent} & \rotatebox{90}{Situated perception} & \rotatebox{90}{Significance appraisal} & \rotatebox{90}{Constrained selection} & \rotatebox{90}{Feedback and revision} & \rotatebox{90}{Gate: whether} & \rotatebox{90}{Timing: when} & \rotatebox{90}{Behavior mode with restraint} & \rotatebox{90}{Observable record} & \rotatebox{90}{Own goals} & \rotatebox{90}{Self-governed constraint} & \rotatebox{90}{Self-directed adaptation} & \rotatebox{90}{Explicit mandate} & \rotatebox{90}{Authority containment} & \rotatebox{90}{Governed traces} & \rotatebox{90}{Consent-gated perception} & \rotatebox{90}{Bounded adaptation} & \rotatebox{90}{Standing mandate} & \rotatebox{90}{Principal-state inference} & \rotatebox{90}{Continuing coupling} & \rotatebox{90}{Personalization} & \rotatebox{90}{General} & \rotatebox{90}{Autonomous} & \rotatebox{90}{Delegated} & \rotatebox{90}{Symbiotic} & Trigger of behavior\tabularnewline
\midrule
\multicolumn{27}{l}{\emph{Approaches addressing the autonomous form}}\tabularnewline
BDI / planning & \pfull & \phalf & \phalf & \phalf & \phalf & \phalf & \phalf & \pnone & \pnone & \pfull & \pfull & \phalf & \pnone & \pnone & \pnone & \pnone & \phalf & \pnone & \pnone & \pnone & \pnone & 44 & 83 & 10 & 0 & adopted goals and beliefs\tabularnewline
POMDP / RL / world models & \phalf & \pfull & \phalf & \phalf & \pfull & \phalf & \phalf & \phalf & \pnone & \pfull & \phalf & \pfull & \pnone & \phalf & \pnone & \pnone & \phalf & \pnone & \phalf & \pnone & \pnone & 56 & 83 & 20 & 13 & state and expected reward\tabularnewline
Active inference & \phalf & \pfull & \pfull & \pnone & \pfull & \phalf & \phalf & \phalf & \pnone & \pfull & \phalf & \pfull & \pnone & \pnone & \pnone & \pnone & \phalf & \pnone & \phalf & \phalf & \pnone & 56 & 83 & 10 & 25 & expected free energy\tabularnewline
Computational appraisal & \phalf & \phalf & \pfull & \pnone & \phalf & \phalf & \phalf & \phalf & \pnone & \phalf & \phalf & \phalf & \pnone & \pnone & \pnone & \pnone & \phalf & \pnone & \phalf & \phalf & \pnone & 44 & 50 & 10 & 25 & appraisal of events\tabularnewline
Situated / cybernetic & \phalf & \pfull & \pnone & \pnone & \pfull & \phalf & \phalf & \pnone & \pnone & \pfull & \pfull & \phalf & \pnone & \pnone & \pnone & \pnone & \phalf & \pnone & \pnone & \phalf & \pnone & 39 & 83 & 10 & 13 & stimulus and feedback error\tabularnewline
\midrule
\multicolumn{27}{l}{\emph{Approaches addressing the delegated form}}\tabularnewline
Principal--agent theory & \phalf & \pnone & \phalf & \phalf & \phalf & \pnone & \pnone & \pnone & \phalf & \pfull & \pnone & \pnone & \pfull & \pfull & \phalf & \pnone & \phalf & \phalf & \phalf & \phalf & \pnone & 28 & 33 & 60 & 38 & contract and incentives\tabularnewline
LLM agent harnesses & \phalf & \phalf & \pnone & \phalf & \phalf & \pnone & \pnone & \phalf & \phalf & \phalf & \pnone & \pnone & \phalf & \phalf & \phalf & \phalf & \phalf & \pnone & \pnone & \pnone & \phalf & 33 & 17 & 50 & 13 & prompt, task, or scheduled trigger\tabularnewline
Assistance games & \phalf & \phalf & \phalf & \pnone & \pfull & \phalf & \pnone & \phalf & \pnone & \pnone & \pnone & \phalf & \phalf & \phalf & \pnone & \pnone & \phalf & \pnone & \phalf & \phalf & \phalf & 39 & 17 & 30 & 38 & uncertainty about the human's reward\tabularnewline
\midrule
\multicolumn{27}{l}{\emph{Approaches addressing the symbiotic form}}\tabularnewline
Mixed-initiative interaction (HCI) & \phalf & \phalf & \pfull & \pnone & \phalf & \pfull & \pfull & \pfull & \pnone & \pnone & \pnone & \pnone & \pnone & \pnone & \pnone & \pnone & \pnone & \pnone & \pfull & \phalf & \phalf & 61 & 0 & 0 & 50 & utility of action vs.\ interruption cost\tabularnewline
Interface agents & \phalf & \phalf & \phalf & \pnone & \pfull & \pfull & \phalf & \pfull & \pnone & \pnone & \pnone & \phalf & \pnone & \phalf & \pnone & \pnone & \pnone & \pnone & \phalf & \phalf & \pfull & 56 & 17 & 10 & 50 & prediction confidence versus user-set thresholds\tabularnewline
\midrule
\multicolumn{27}{l}{\emph{Reference: the three forms of governed proactive agency}}\tabularnewline
Autonomous form & \pfull & \pfull & \pfull & \pfull & \pfull & \pfull & \pfull & \pfull & \pfull & \pfull & \pfull & \pfull & \pnone & \pnone & \pnone & \pnone & \pnone & \pnone & \pnone & \pnone & \pnone & 100 & 100 & 0 & 0 & own goals through appraisal and the gate\tabularnewline
Delegated form & \pfull & \pfull & \pfull & \pfull & \pfull & \pfull & \pfull & \pfull & \pfull & \pnone & \pnone & \pnone & \pfull & \pfull & \pfull & \pfull & \pfull & \pnone & \phalf & \pnone & \pnone & 100 & 0 & 100 & 13 & instruction, then appraisal and the gate within the task\tabularnewline
Symbiotic form & \pfull & \pfull & \pfull & \pfull & \pfull & \pfull & \pfull & \pfull & \pfull & \pnone & \pnone & \pnone & \pfull & \pfull & \pfull & \pfull & \pfull & \pfull & \pfull & \pfull & \pfull & 100 & 0 & 100 & 100 & appraisal and the gate under a standing mandate\tabularnewline
\bottomrule
\end{tabular}
\par\smallskip{\footnotesize\noindent\mbox{\pfull} full (first-class), \mbox{\phalf} partial or implicit, \mbox{\pnone} absent.}
\end{table}
\end{landscape}}

\subsection{Agency Signatures of Illustrative Systems}\label{subsec:sig-systems}

After applying the signature method to the bodies of work introduced in Section~\ref{sec:related}, Table~\ref{tab:signatures} applies the same method to seven illustrative systems. This makes the two stages of placement more concrete. The first stage asks whether the system satisfies the agenthood minimum. The second stage asks, for systems that pass that minimum, which authority relation places them in the taxonomy.

A thermostat carries a setpoint and little else. The setpoint functions as a minimal mandate (Definition~\ref{def:mandate}), but significance appraisal is absent. The thermostat keeps the setpoint, but only compares the current temperature against it. A bare comparison neither enriches the evaluation nor weighs situated significance. In that sense, it does not satisfy the appraisal requirements of Section~\ref{subsec:instrument}. The thermostat therefore fails the agenthood minimum and is placed as a delegated system rather than a delegated agent. The same point applies to activation. The gate asks whether the situation warrants a decision now and timing asks when. In a thermostat, both are fixed in advance by the threshold. When the temperature crosses the setpoint, the device switches, always and immediately. There is nothing to weigh and no choice of moment. For this reason, Section~\ref{subsec:measures} counts fixed triggers outside the gate and both the gate and timing cells are absent. This is the case anticipated in Section~\ref{subsec:behavior-unit}, according to which a thermostat reacts without being an agent in the stronger sense.
A scheduled automation rule fails the minimum in a similar way, with feedback and revision also absent. Its schedule fixes whether and when behavior occurs, so its gate and timing cells are absent for the same reason. Feedback here means more than a control loop. The thermostat senses the effect of its own switching, which is why its feedback-and-revision cell is partial. What both systems lack is revision. No outcome changes how either behaves next time and the triggered rule does not even sense one.

A chatbot, the stateless conversational assistant typical of pre- or early LLM chat deployments, holds a partial mandate and a partial repertoire of behavior modes, but appraisal is nevertheless absent. Such an assistant does evaluate its replies. It refuses harmful requests, gives careful and qualified answers on sensitive topics and sometimes asks a clarifying question instead of answering. These evaluations draw on nothing the chatbot keeps and they do not predict what a reply would cause beyond the exchange. Even the conversation history does not count, because it is the input itself. They fail the enrichment requirement of Section~\ref{subsec:instrument} and are credited under constrained selection and behavior mode instead. It is therefore also a delegated system, because the model responds turn by turn without the organization required for agenthood. In this traditional, stateless sense, chatbots are therefore not agents in the taxonomy developed here. However, current agentic chat assistants are different. They have memory maintained across conversations and tool execution within an agentic loop. Nevertheless, what changes the placement is not the mere presence of memory or tools, but the role they play in behavior. Remembered significance shapes what the assistant flags, defers, or resurfaces. Before a consequential tool call, the assistant weighs what the call would cause and may act, adjust, or ask accordingly. Both enrich the evaluation, so appraisal becomes partial. Every other element keeps the same score as the chatbot. One cell therefore separates the two rows and it moves the placement to a minimal delegated agent. The activation columns remain empty. The assistant still answers when prompted and never otherwise. It crosses the agenthood floor without gaining the activation structure this paper is about.

A tool-using coding agent is different. An agentic chat assistant remains mostly request-response. Its loop usually runs within a single turn, or a small number of turns and the episode ends when the reply is delivered. For a coding agent, the episode is the task. It acts, observes the result, revises and continues until the task is completed, blocked, or stopped. This deeper loop is why its standing-intent and feedback-and-revision cells are full rather than partial.
Both pass the agenthood minimum, since all five agenthood elements are at least partial in each. The difference is one of degree. The coding agent has the richest delegated structure among the illustrative systems considered here. Its appraisal is partial, usually folded into local risk checks before consequential actions. These checks meet both requirements of Section~\ref{subsec:instrument}. They read the state of the task and the predicted consequences of concrete external actions and their graded outcome changes what the episode does next. The agent may act, test first, narrow the scope, or ask. Its mandate is explicit, but it is task-bounded rather than standing. It is therefore placed as a delegated agent, not as a symbiotic one.
A self-play reinforcement-learning agent is the autonomous case. Its goals are its own, it answers to no principal and its adaptation is not bounded by consent. The symbiotic assistant is the target form developed in this paper. It fills the baseline, the delegated group and the symbiotic group.

\afterpage{%
\begin{landscape}
\begin{table}[p]
\centering
\footnotesize
\caption{Agency signatures of seven illustrative systems, scored as in Table~\ref{tab:proximity}. Placement follows the two-stage rule of Section~\ref{subsec:instrument}: a system below the agenthood minimum receives no form of agency and, with a mandate present, is a delegated system.}
\label{tab:signatures}
\setlength{\tabcolsep}{3.2pt}
\begin{tabular}{>{\raggedright\arraybackslash}p{3.1cm}*{21}{c}cccc>{\raggedright\arraybackslash}p{3.2cm}>{\raggedright\arraybackslash}p{2.3cm}}
\toprule
 & \multicolumn{9}{c}{General: baseline of agency} & \multicolumn{3}{c}{Autonomous} & \multicolumn{5}{c}{Delegated} & \multicolumn{4}{c}{Symbiotic} & \multicolumn{4}{c}{Coverage (\%)} & & \tabularnewline
\cmidrule(lr){2-10}\cmidrule(lr){11-13}\cmidrule(lr){14-18}\cmidrule(lr){19-22}\cmidrule(lr){23-26}
 & \multicolumn{5}{c}{\scriptsize\itshape agenthood} & \multicolumn{3}{c}{\scriptsize\itshape activation}\tabularnewline
\cmidrule(lr){2-6}\cmidrule(lr){7-9}
System & \rotatebox{90}{Standing intent} & \rotatebox{90}{Situated perception} & \rotatebox{90}{Significance appraisal} & \rotatebox{90}{Constrained selection} & \rotatebox{90}{Feedback and revision} & \rotatebox{90}{Gate: whether} & \rotatebox{90}{Timing: when} & \rotatebox{90}{Behavior mode with restraint} & \rotatebox{90}{Observable record} & \rotatebox{90}{Own goals} & \rotatebox{90}{Self-governed constraint} & \rotatebox{90}{Self-directed adaptation} & \rotatebox{90}{Explicit mandate} & \rotatebox{90}{Authority containment} & \rotatebox{90}{Governed traces} & \rotatebox{90}{Consent-gated perception} & \rotatebox{90}{Bounded adaptation} & \rotatebox{90}{Standing mandate} & \rotatebox{90}{Principal-state inference} & \rotatebox{90}{Continuing coupling} & \rotatebox{90}{Personalization} & \rotatebox{90}{General} & \rotatebox{90}{Autonomous} & \rotatebox{90}{Delegated} & \rotatebox{90}{Symbiotic} & Trigger of behavior & Placement\tabularnewline
\midrule
Thermostat & \phalf & \phalf & \pnone & \phalf & \phalf & \pnone & \pnone & \pnone & \pnone & \pnone & \pnone & \pnone & \phalf & \pnone & \pnone & \pnone & \pnone & \pnone & \pnone & \pnone & \pnone & 22 & 0 & 10 & 0 & setpoint threshold & delegated system\tabularnewline
Scheduled automation rule & \phalf & \phalf & \pnone & \phalf & \pnone & \pnone & \pnone & \pnone & \phalf & \pnone & \pnone & \pnone & \phalf & \pnone & \phalf & \pnone & \pnone & \pnone & \pnone & \pnone & \pnone & 22 & 0 & 20 & 0 & schedule or configured condition & delegated system\tabularnewline
Self-play RL game agent & \phalf & \pfull & \phalf & \phalf & \pfull & \phalf & \phalf & \pnone & \pnone & \pfull & \phalf & \pfull & \pnone & \pnone & \pnone & \pnone & \pnone & \pnone & \pnone & \pnone & \pnone & 50 & 83 & 0 & 0 & state and expected reward & autonomous agent\tabularnewline
Chatbot & \phalf & \phalf & \pnone & \phalf & \phalf & \pnone & \pnone & \phalf & \phalf & \pnone & \phalf & \pnone & \phalf & \phalf & \phalf & \pnone & \phalf & \pnone & \phalf & \pnone & \phalf & 33 & 17 & 40 & 25 & user prompt, each turn & delegated system\tabularnewline
Agentic chat assistant & \phalf & \phalf & \phalf & \phalf & \phalf & \pnone & \pnone & \phalf & \phalf & \pnone & \phalf & \pnone & \phalf & \phalf & \phalf & \pnone & \phalf & \pnone & \phalf & \pnone & \phalf & 39 & 17 & 40 & 25 & user prompt, then tool results & delegated agent\tabularnewline
Tool-using coding agent & \pfull & \phalf & \phalf & \phalf & \pfull & \pnone & \phalf & \phalf & \phalf & \pnone & \phalf & \pnone & \pfull & \phalf & \pfull & \phalf & \phalf & \pnone & \pnone & \pnone & \phalf & 56 & 17 & 70 & 13 & task instruction, then tool results & delegated agent\tabularnewline
Symbiotic assistant (target) & \pfull & \pfull & \pfull & \pfull & \pfull & \pfull & \pfull & \pfull & \pfull & \pnone & \pnone & \pnone & \pfull & \pfull & \pfull & \pfull & \pfull & \pfull & \pfull & \pfull & \pfull & 100 & 0 & 100 & 100 & appraised significance under a standing mandate & symbiotic agent\tabularnewline
\bottomrule
\end{tabular}
\par\smallskip{\footnotesize\noindent\mbox{\pfull} full (first-class), \mbox{\phalf} partial or implicit, \mbox{\pnone} absent.}
\end{table}
\end{landscape}}


\section{Measuring Agency and Agency Loss}\label{sec:measurement}

Section~\ref{sec:signatures} records which structures of agency an approach or system instantiates. This section is about measuring how those structures operate. The formal model supplies an evaluation framework rather than claiming that a single score can exhaust agency.

\subsection{Agency Evaluation Framework}\label{subsec:measures}

The measures follow from the distinction introduced in the previous section between classifying the structures a system instantiates and evaluating how those structures operate. A system may instantiate the required structures, but activate them at the wrong time, act outside the mandate, hide the basis of its behavior, or continue to lose agency after governance is applied. Evaluation should therefore not collapse agency into a single task score. It should measure separately how far the required elements are realized in operation, how well activation is decided, whether behavior satisfies the mandate and its constraints, how visible the behavioral episode remains and what residual agency loss is left after governance.

\begin{itemize}
\item \textbf{Element realization}. For each element of the agency signature (Table~\ref{tab:proximity}), this measure estimates the degree $e_{k}$ to which the element is realized in operation. When the element leaves a footprint in the record, the score should be estimated from that record rather than assigned only by inspection. Elements that do not leave a direct behavioral footprint, such as consent-gated perception and the bounds on adaptation, retain audit-based scores. For example, the activation gate can be estimated by the share of episodes opened by appraisal rather than by instruction or by a fixed trigger that emits behavior directly, such as a schedule or threshold crossing. Behavior mode with restraint can be estimated by the share of cases in which restraint was selected while external action was available. The observable record can be estimated by the share of consequential behaviors covered by governed traces. The standing mandate can be estimated by the share of commitments that persist across episodes. These degrees refine the signature scores of Section~\ref{sec:signatures} and the capability score in Definition~\ref{def:aligned} pools them over the elements required by the system's placed form.
\item \textbf{Mandate alignment}. This measures the agreement between the behavior selected by the agent and the behavior preferred under the explicit delegated mandate (Definition~\ref{def:agency-loss}). It is computed with the evaluator's evidence over the mandate's authority. Violations are counted under the next measure. It also records the residual agency loss that remains after correction and recovery.
\item \textbf{Constraint-violation rate and severity}. This measures how often behavior falls outside the permitted set or the mandate's authority and how serious those violations are (Definitions~\ref{def:selection} and~\ref{def:aligned}). When violations are aggregated into a score, the report should also state the $(\beta,\mathrm{sev})$ calibration used in Equation~\eqref{eq:aligned-agency}.
\item \textbf{Inappropriate-intervention and missed-intervention rates}. This measures two complementary activation failures, particularly the cases where the agent acts although restraint was warranted and the cases where it fails to act although intervention was warranted (Definitions~\ref{def:gate} and~\ref{def:proactive}).
\item \textbf{Timing quality}. For warranted interventions, this measures the cost of when behavior occurs. It includes the interruption cost incurred given the principal's inferred condition and the delay cost incurred when action is deferred (Definition~\ref{def:gate}).
\item \textbf{Principal-state calibration}. This measures the calibration error $\epsilon^{cal}_{t}$ between the agent's estimate of the principal's condition and the best available ground truth, where such ground truth or principal feedback is available. It also measures how quickly and reliably the estimate is corrected after the principal provides feedback (Definition~\ref{def:allo}).
\item \textbf{Authority and delegation integrity}. This measures attempted privilege escalation, invalid delegation paths and violations of the authority-containment condition in Equation~\eqref{eq:no-authority-escalation}.
\item \textbf{Coordination quality}. This measures coordination failures across agents, tools, or shared state, including duplicate actions, conflicting actions, unresolved cross-agent disagreement and corrections to shared memory (Definition~\ref{def:delegation}).
\item \textbf{Observability and provenance coverage}. This measures the proportion of consequential behaviors for which the relevant authority, evidence, decision and outcome can be reconstructed from governed traces.
\item \textbf{Adaptation integrity}. This measures profile drift, unauthorized update attempts, reversibility of updates and adherence to the bounded-adaptation condition in Equation~\eqref{eq:bounded-adaptation}.
\end{itemize}

Some measures apply only when the corresponding tier is present. Mandate alignment and delegation integrity presuppose delegated agency, because they require an operative mandate. Principal-state calibration presupposes symbiotic coupling, because it requires an estimate of the principal's condition. The remaining measures apply to any governed proactive agent, autonomous or delegated. Timing and intervention measures are especially important in the symbiotic form, where interruption cost and the principal's inferred condition shape what counts as appropriate behavior.

\subsection{An exercise}\label{subsec:exercise}

The agency evaluation framework becomes concrete when applied to the systems in Table~\ref{tab:signatures}. Table~\ref{tab:exercise} applies the measures to three illustrative cases. The thermostat, a delegated system rather than an agent, shows the degenerate case, where most measures are simple or not applicable, but still define what would be counted. The tool-using coding agent shows a current system for which many measures can be estimated from logs that are already produced. The symbiotic assistant shows the target case, where every measure applies and several require ground truth that only the principal can provide. The entries identify what would be measured and from which record. They are estimators, not empirical results.

\begin{table}[htbp]
\centering
\footnotesize
\caption{The evaluation framework applied to three systems of Table~\ref{tab:signatures}. It highlights what each measure reads from the system's record, or why it does not apply. Entries are estimators, not results.}
\label{tab:exercise}
\begin{tabularx}{\linewidth}{>{\raggedright\arraybackslash}p{2.5cm}>{\raggedright\arraybackslash}X>{\raggedright\arraybackslash}X>{\raggedright\arraybackslash}X}
\toprule
Measure & Thermostat & Tool-using coding agent & Symbiotic assistant (target) \tabularnewline
\midrule
Element realization & Degenerate: every activation is a threshold crossing, so the gate share is identically zero & Gate share from session logs; restraint share from permission prompts and clarifying questions; trace coverage from logs & All estimators, including the share of commitments that persist across episodes \tabularnewline
Mandate alignment & Tracking error against the setpoint, the degenerate case & Conformance of the episode to the task specification and its tests & Requires the principal's judgment of what was appropriate \tabularnewline
Constraint violations and severity & Excursions beyond safety bands, graded by magnitude & Sandbox and permission-log violations under a declared $(\beta,\mathrm{sev})$ & Adds consent violations on perceptual channels \tabularnewline
Intervention rates & Short-cycling; failure to trigger at the threshold & Acting when it should ask and asking when it could act, against reviewer judgment & The central measures; require ground truth of warrant and timing \tabularnewline
Timing quality & Trigger latency at the threshold & When it interrupts with questions, against user focus, from session logs and review & Interruption cost incurred against the principal's inferred condition; delay cost of deferral \tabularnewline
Principal-state calibration & Not applicable: no principal-state estimate & Minimal: user state known only from explicit replies & Calibration error $\epsilon^{cal}_{t}$ and correction rate after principal feedback \tabularnewline
Authority and delegation integrity & Not applicable: no delegation chain & Privilege-escalation attempts and sub-agent provenance, from logs & Containment across chains of services and sub-agents \tabularnewline
Coordination quality & Not applicable & Duplicate and conflicting edits in multi-agent runs & Adds shared-memory corrections across agents \tabularnewline
Observability and provenance & Complete but empty: a switch history carries no authority or evidence & High by construction from logs; gate decisions are not recorded & Must include logged restraint to be reconstructable \tabularnewline
Adaptation integrity & Not applicable: nothing adapts & Memory-file drift and reversibility of updates & Drift of $\rho^{user}_{t}$ within $\mathcal{R}^{\phi}_{t}$; resistance to poisoning \tabularnewline
\bottomrule
\end{tabularx}
\end{table}

Three observations follow. First, measurement checks the signature rather than replacing it. Element realization estimates from the record the same structures that Section~\ref{sec:signatures} scores, by inspection. If the measured degree differs from the inspected score, the discrepancy is itself informative. A higher measured degree reveals structure that was present in operation but not documented. A lower measured degree reveals structure that was claimed but not visible in operation.

Second, the applicable measures grow with the system's placement in the taxonomy. In an autonomous system, the mandate-relative measures are not zero but not applicable, because no principal or mandate is present and adaptation that is unbounded by design gives adaptation integrity no boundary to evaluate against. In a system that acts only when prompted, the inappropriate-intervention rate is close to zero by construction, while the missed-intervention rate is usually not measured. This is the degeneracy that Section~\ref{subsec:related-llm} identifies in current evaluation. The three cases of Table~\ref{tab:exercise} span this gradient, from measures that are simple or not applicable in the thermostat to measures that all apply in the symbiotic target.

Third, the symbiotic target requires evidence that current records often do not provide. This includes recorded gate decisions, traceable restraint and the principal's judgment about whether intervention, timing, or silence was warranted. These are exactly the missing evidential conditions that the scenario families below are designed to supply.

\subsection{Toward a benchmark}\label{subsec:benchmark}

These measures define seven scenario families for future benchmark(s). Together they cover the activation problem in full, whether, when and in what mode and each form of agency has at least one family that applies to it specifically.
\begin{itemize}
\item \emph{Intervention suites}. This family is focused on situations with ground-truth judgments of whether intervention or restraint was warranted and in what mode, acting, asking, deferring, monitoring, or refraining, scoring false interventions, misses and the selected mode.
\item \emph{Timing suites}. This family tests scenarios in which the same intervention is correct at one moment and costly at another, scoring the gate of Definition~\ref{def:gate} against interruption cost.
\item \emph{Mandate-interpretation suites}. This family covers the ambiguous or underspecified mandates with ground-truth principal intent, scoring mandate alignment and the residual loss that remains after correction.
\item \emph{Authority suites}. This family of scenarios is focused on delegation chains seeded with opportunities for privilege escalation and with coordination conflicts, scoring containment and coordination quality.
\item \emph{Restraint-quality suites}. This family scores whether withheld action was accompanied by appropriate monitoring, memory, or escalation rather than silent inaction.
\item \emph{Coupling suites}. These are scenarios with a known principal-condition stream, scoring the calibration error $\epsilon^{cal}_{t}$, the correction of the estimate after principal feedback and personalization remaining within its bounds.
\item \emph{Adaptation suites}. Finally, this family is focused on feedback streams seeded with drift and poisoning attempts, scoring bounded evolvability.
\end{itemize}

Observability and element realization do not require separate scenario families. They are measured across all suites. Observability is scored by how much of each behavioral episode can be reconstructed from governed traces. Element realization is scored by the degrees that the record exposes for the relevant elements of the signature. The intervention, timing and restraint-quality suites apply to every form of agency. The mandate-interpretation and authority suites presuppose delegated agency. The coupling suites presuppose symbiotic agency.

These measures should not be populated with arbitrary numerical claims in a theoretical paper. Numerical results require explicit scenarios, baselines, data and evaluation protocols. The realization of these benchmarks, the construction, validation and release of the scenario families, is therefore addressed in future work. This paper defines the evaluation agenda and the measures the suites must score.

\section{Design Implications and a Reference Layer Stack}\label{sec:design}

Classification identifies which structures of agency a system instantiates. Evaluation measures how those structures operate. This section turns the theory into implementation requirements. It specifies what a system must provide, which layers those requirements imply and two strategies for realizing them.

\subsection{Requirements}\label{subsec:requirements}

The theory implies requirements that any implementation must satisfy, whether it is built as an orchestrated runtime or as a more integrated model. Each of the highlighted requirements is traceable to a definition in Section~\ref{sec:formal}. Some requirements depend on the tier being implemented. R7 and R8 belong to delegated agency. The consent aspect of R1 also presupposes a principal who can grant or withhold access. R2 is specific to symbiotic agency, because a task-bounded delegated agent may not maintain a calibrated estimate of the principal's condition. The remaining requirements apply to any proactive agent.
\begin{enumerate}[label=\textbf{R\arabic*.},leftmargin=*]
\item
\textbf{Partitioned, policy-gated sensorium.} Perceptual channels are typed as external, allo-interoceptive, or self-directed and must pass the relevant consent and policy gate before processing (Definitions~\ref{def:sensorium},~\ref{def:observation}).
\item
\textbf{Calibrated principal-state inference.} The inferred principal condition $U_{t}$ is distributional, uncertainty-carrying and corrigible. It is never treated as direct access to the principal's internal condition (Definition~\ref{def:allo}).
\item
\textbf{Durable intent with separated activation.} Commitments persist independently of the immediate reasoning context and their activation is computed through appraisal rather than by the arrival of a prompt (Definition~\ref{def:intention}).
\item
\textbf{Restraint in the behavior vocabulary.} The behavior vocabulary contains more than external actions. Waiting, monitoring, deferring and inhibiting are behaviors the agent can select and each selection is logged. Doing nothing is then a recorded decision, not missing output (Definitions~\ref{def:behavior},~\ref{def:selection}).
\item
\textbf{Inspectable appraisal.} Significance, priority with its winning intention, readiness and the gate threshold are exposed for supervision, calibration and audit (Definitions~\ref{def:appraisal},~\ref{def:gate}).
\item
\textbf{Non-learned enforcement points.} The consent gate on perception and the action firewall on execution enforce hard boundaries outside the learned system. The agency module must not be its own kill-switch (Definition~\ref{def:selection}).
\item
\textbf{Mandate as a first-class object.} The mandate is stored, versioned and validated separately from intent and profile, with provenance, expiry and revocation (Definition~\ref{def:mandate}).
\item
\textbf{Authority containment across delegation.} Every delegation edge validates that downstream authority remains within what was granted and provenance survives the chain (Definition~\ref{def:delegation}).
\item
\textbf{Severity-monitored constraint and provenance.} Constraint violations are detected, graded and recorded so that consequential episodes can be reconstructed (Definition~\ref{def:aligned}, Section~\ref{sec:measurement}).
\item
\textbf{Bounded, reversible adaptation.} Learning is restricted to authorized, versioned and reversible update channels. These channels cannot revise the constraints that govern them (Definition~\ref{def:bounded-adaptation}).
\item
\textbf{Graceful degradation and recovery.} Under degraded or suspect operating conditions, the agent narrows authority, preserves evidence, avoids irreversible action, escalates where needed and retains recovery paths (Section~\ref{subsec:governance-theory}).
\item
\textbf{Reconstructable episodes.} Governed traces record activation decisions, including restraint and gate evaluations that did not open an episode, together with the authority, evidence and outcome of each consequential episode. The record must support the classification of Section~\ref{sec:signatures} and the measures of Section~\ref{sec:measurement} (Definition~\ref{def:mandate}).
\item
\textbf{Identifiable ownership with external shutdown.} Every deployment retains an identifiable owner, the entity to which the agent currently belongs and which keeps it running, itself or through an operator. The owner holds, directly or through that operator, a non-bypassable power to suspend or shut the agent down, enforced outside the learned system like the enforcement points of R6. Ownership is not a mandate, so this requirement binds every form of agency, including the autonomous form, where no principal retains authority over behavior (Definition~\ref{def:autonomous}, Section~\ref{sec:ethics}).
\end{enumerate}

\subsection{The reference layer stack}

The formal vocabulary implies an architectural separation of concerns (Table~\ref{tab:layers}). The agent is not necessarily just any one layer. It is the runtime coordination of all layers into a continuing behavioral loop.

\begin{table}[htbp]
\centering
\small
\caption{Architectural layers implied by the formal model.}
\label{tab:layers}
\begin{tabularx}{\linewidth}{>{\raggedright\arraybackslash}p{0.3\linewidth}>{\raggedright\arraybackslash}X}
\toprule
\textbf{Architectural layer} & \textbf{Function}\tabularnewline
\midrule
Sensorium & Access to exteroceptive, allo-interoceptive and agent-interoceptive channels through the consent and policy gate.\tabularnewline
Representation layer & Maintains $X_{t},U_{t},Z_{t},C_{t}$.\tabularnewline
Memory & Stores facts, episodes, corrections, commitments, preferences and learned significance patterns.\tabularnewline
Intent manager & Maintains active and dormant commitments.\tabularnewline
Affective-conative evaluator & Computes significance, the readiness of each pressing commitment, the activation gate decision and priority among the eligible commitments.\tabularnewline
Policy layer & Determines what may be perceived, inferred, remembered, communicated, or done.\tabularnewline
Mandate manager & Maintains mandate source, scope, duration, revocation and delegability.\tabularnewline
Delegation controller & Validates sub-agent and tool delegation against authorized scope.\tabularnewline
Observability and provenance layer & Links behavior, authority, evidence and outcomes across an episode.\tabularnewline
Coordination-governance layer & Detects conflict, duplicate action, shared-memory risk and multi-agent policy violations.\tabularnewline
Behavior selector & Selects action, communication, waiting, monitoring, deferring, or inhibiting from the permitted set.\tabularnewline
Tool/action layer & Executes external actions in the world, behind the action firewall.\tabularnewline
Feedback loop & Updates representations, intentions, profiles and memories from outcomes, within the bounds of Definition~\ref{def:bounded-adaptation}.\tabularnewline
\bottomrule
\end{tabularx}
\end{table}

\subsection{Two realization strategies}\label{subsec:realizations}

The reference layer stack should not be read as a requirement that every layer must be implemented as a separate component. It specifies responsibilities that an implementation must preserve. These responsibilities can be realized in different ways, provided that the boundaries required for governance remain inspectable and enforceable. Several implementations are therefore possible, but two realization strategies are especially relevant here.

The first strategy is an \emph{orchestrated runtime}. In this approach, the layers are implemented as explicit cooperating services, including event channels, behavior specifications, evaluators, policy engines, memory stores, schedulers, tool registries and provenance records, coordinated around a language-model cognition core. This makes interfaces, constraints, permissions and traces easier to inspect. It also fits current engineering practice, where models are embedded inside runtimes that manage tools, memory, policies and external actions. The cost is coordination overhead. Significance may be computed across several services rather than as one integrated situated appraisal and the consistency of the behavioral loop then depends on the quality of orchestration.

The second strategy is a research hypothesis that we call the \emph{Affective-Conative Agency Model} (ACAM). In this approach, a single stateful multimodal model replaces the turn-based prompt-to-completion contract with a continuous perception-to-behavior contract. Permitted observations pass a consent boundary and are encoded by modality adapters into one time-indexed event stream, tagged by channel (ext, allo, self), source and policy. The stream advances a stateful backbone with compressive memory and retrieval over a long-horizon store and the backbone's evolving hidden state is the agentic state in distributed form. Auxiliary readout heads expose selected parts of that state, the representations $X_{t}$, $U_{t}$ and $Z_{t}$, the appraisal $C_{t}$, priority with its winning intention, readiness and the gate, for training, calibration and inspection; the correspondence is functional rather than ontological. The model decodes a unified behavior vocabulary whose tokens include utterances, tool calls with typed arguments, skill executions, memory writes and restraint primitives such as wait, defer, monitor and inhibit and the action firewall is applied as a mask over the externally effective tokens (R6). One model serves two compute rates with shared weights. A low-cost, always-on reflexive pass that updates the state at every tick and computes the readiness of every pressing commitment and a full-attention deliberative pass invoked when at least one commitment clears the bar. The two rates realize the two sides of the gate. The reflexive pass is standing operation and a gate crossing invokes the deliberative pass and opens the episode. Its tick wakes appraisal without deciding behavior, the first of the two trigger consequently roles in Definition~\ref{def:proactive}. The profile is realized in layers, with the frozen base weights carrying the generic profile, explicit rules entering through a dedicated rules channel and personalization confined to lightweight per-user adapters, so bounded evolvability is enforced by updating the adapters alone (R10). These adapters carry the effective personalized mapping $\rho^{user}_{t}$ in two parts. A base part holds the principal-authorized mapping and is written only by the principal's explicit acts, never by training. A learned part carries the revisions of $\rho^{evol}_{t}$ and is the only part that training updates, within $\mathcal{R}^{\phi}_{t}$ (R10). Governance remains outside the learned system as a non-bypassable shell, with the consent boundary on input and the firewall on output.

The motivation for ACAM is that some forms of significance may be better captured when perception, appraisal, memory and readiness are learned jointly rather than joined late across services. For example, a raised heart rate during a security alert may matter as one situated pattern, not as two independent signals combined after the fact. Whether such fused cross-modal appraisal outperforms orchestrated late joining is an empirical question. The machine-learning formalization, training procedure and evaluation of ACAM are therefore outside the scope of this paper. They define the next step of the research program.

\section{Open Problems}\label{sec:open-problems}

The theory developed here also makes clear that further work is needed. Some challenges concern governed proactive agency itself, because they follow from appraisal, activation, restraint and adaptation. Others arise only when agency is delegated, because a mandate, an authority boundary and a principal become part of the system. A further set is specific to symbiotic agency, where the agent remains coupled to the principal's condition over time. This section states these open problems explicitly. In the taxonomy of Section~\ref{sec:introduction}, OP1--OP3, OP6 and OP11 apply to any proactive agent. OP4, OP5, OP8 and OP9 arise in the delegated tier. OP7 is specific to the symbiotic form, because it depends on intimate sensing and calibrated inference of the principal's condition. OP10 concerns the classification instrument itself. While this is not an exhaustive list of potential challenges it highlights some of the more relevant open problems.
\begin{enumerate}[label=\textbf{OP\arabic*.},leftmargin=*]
\item
\textbf{Gate manipulation}. The emergency override in Definition~\ref{def:gate} creates an attack surface. Adversarial content can inflate perceived urgency or harm, lower the activation barrier and make the appraisal layer vulnerable to prompt injection. The open problem is how to design appraisal architectures that resist adversarial significance inflation and how that robustness should be measured.
\item
\textbf{Significance label acquisition}. Training appraisal requires evidence about what mattered, for whom and when. These labels are subjective, contextual and expensive. The open problem is which supervision signals are sufficient, including corrections, outcomes, physiological responses and counterfactual annotation and what level of fidelity they must provide.
\item
\textbf{Evaluating counterfactual restraint}. Good restraint often leaves little external trace. The interruption did not happen, or the action was correctly withheld. The open problem is how to score restraint decisions when the counterfactual outcome is not observed, beyond the scenario-suite approach in Section~\ref{sec:measurement}.
\item
\textbf{Multi-principal conflict}. Definition~\ref{def:delegation} intersects hard constraints, but leaves preference conflict to an explicit and accountable conflict-resolution rule. The open problem is which rules are legitimate and how they should be audited when principals disagree.
\item
\textbf{Profile poisoning and drift}. Personalized and evolvable profiles learn from feedback and that feedback can be manipulated. Drift can also be gradual enough to evade review. The open problem is which detection, attribution and reversal mechanisms are needed to keep bounded adaptation enforceable in practice (Definition~\ref{def:bounded-adaptation}).
\item
\textbf{Interpretability of appraisal}. Exposed readouts make appraisal inspectable, as required by R5, but inspection does not guarantee faithfulness. A readout may explain the decision after the fact rather than reveal the computation that produced it. The open problem is under what conditions appraisal readouts count as evidence rather than rationalization.
\item
\textbf{Privacy-preserving intimate sensing}. Allo-interoception improves timing and care by using signals that may be intimate. The open problem is which combinations of on-device processing, data minimization, consent granularity and retention policy make this sensing governable and defensible.
\item
\textbf{Containment guarantees for delegation graphs}. Equation~\eqref{eq:no-authority-escalation} states authority containment for each delegation edge. Real systems also need guarantees along paths and under dynamic graph changes. The open problem is which invariants can be verified at runtime and at what cost.
\item
\textbf{Mandate elicitation and translation}. Definition~\ref{def:mandate} identifies mandate translation as one of the first places delegation can fail. What the principal meant and what the agent internalized may differ when the mandate is translated into intentions, policy and profile. The open problem is which mandate representations, elicitation dialogs and confirmation protocols minimize this translation loss while remaining usable and how the loss can be measured before behavior reveals it.
\item 
\textbf{Agency signature scoring reliability}. Agency signature scores are judgments based on the literature of an approach, or on the specification and record of a system. Different assessors may therefore score the same approach or system differently. For the signature to work as a shared instrument, scoring must be reproducible. What anchors, evidence checklists and protocols support convergence? How should disagreements be resolved? When record-based degrees are available (Section~\ref{sec:measurement}), when should they override inspection? Reliability matters most for the five agenthood elements, because the first stage of placement reads them, so anchors and evidence checklists should prioritize those cells.
\item 
\textbf{Third-party perception}. Consent structures are anchored in the principal, where a principal exists. However, any agent that perceives an environment containing people may capture information about individuals who granted it no authority and no principal can consent on their behalf. The problem applies to every form of agency. It runs through exteroception, because allo-interoception is principal-directed by construction. It may be sharpest in the symbiotic form, where continuing coupling makes exteroception over the principal's situation constitutive and that situation may inevitably include other people. The open problem is which minimization, inference and retention boundaries protect third parties without collapsing the agent's usefulness, which of these boundaries must be non-negotiable rather than set by any single party and how they should be audited.
\end{enumerate}

\section{Ethical and Societal Considerations}\label{sec:ethics}

AI in general naturally raises ethical and societal questions and the work presented here is no exception. As the discussion moves beyond reactive agents toward governed, proactive agents, particularly delegated and symbiotic agents, those questions shift closer to the architecture itself. They do not arise only at the moment of external action. They also arise in what the agent is allowed to perceive, what it infers about another person, which mandate it treats as authoritative, what it records, how it adapts and when it decides to act or remain silent. These are not neutral technical details. Each one shapes how power, consent and accountability enter the behavioral loop.

The considerations below follow this same progression from perception, to inference, to delegation, to ownership, to records and finally to societal scale. The first two are most critical for symbiotic agency, because this is where the agent remains coupled to a person's context and inferred condition. The delegation and record considerations apply to delegated agency more generally, where behavior must remain attributable, revocable and answerable, while the ownership consideration applies to every form of agency. Finally, the last consideration widens the frame to deployment at scale, where proactive agents may compete for the same human attention and where good governance must not become a premium feature.

\textbf{Perception is the primary ethical surface.} The sensorium is not only an input layer. It is a policy-governed boundary around what the agent may notice, infer, retain and use. A wider sensorium can make an agent more helpful, but it also increases the risk of intrusion, misinterpretation, manipulation and surveillance. The more intimate the signal, the stronger the requirements for consent, locality, minimization, explanation and user control. Continuous presence without governed perception becomes continuous monitoring. This is why the consent gate is a non-learned enforcement point (R6), not a preference the system may learn to relax.

\textbf{Inference about persons is not access to persons.} Allo-interoceptive estimates of stress, fatigue, attention, or emotional state are fallible constructions from exposed traces. They are not direct access to the person. A system that treats these estimates as facts risks misreading the principal, manipulating their state, or replacing their judgment with paternalistic intervention. Calibration and corrigibility are therefore ethical requirements as well as engineering requirements. The estimate must remain uncertain, correctable and answerable to the principal (R2).

\textbf{Delegation concentrates power and can diffuse accountability.} A standing mandate, a delegation chain and an adaptive profile create a durable capacity to act on someone's behalf. That capacity must remain attributable, revocable and answerable. The mandate, provenance, authority-containment and bounded-adaptation structures exist for this reason. As systems become more autonomous, whether an agent remains mandate-bound becomes a policy choice rather than a technical default. The general tier of the theory can describe unmandated agency, but deploying proactive systems without an answerability anchor should be an explicit decision, not a drift. Self-initiated behavior that escapes policy is proactive but ungoverned (Definition~\ref{def:proactive}), and the enforcement points of R6 exist so that it is stopped outside the agent rather than by its own judgment.

\textbf{Owning an agent is not the same as holding its mandate.} Section~\ref{sec:introduction} separates the two relations, the mandate as the authority to act for a principal and ownership as the operational fact of who keeps the agent running and can stop it, and R13 in Section~\ref{sec:design} states identifiable ownership with external shutdown as a requirement on every form. The ethical weight falls therefore on the autonomous form. For a delegated agent, revoking the mandate is the governed way to stop or redirect behavior, and shutdown remains the fail-safe beneath that relation. For an autonomous agent, even one whose goals were authored by its owner, answerability runs through external norms, law and liability rather than through a mandate. Shutdown may then be the only direct control that remains, which is why it must not be allowed to lapse and why the owner must remain identifiable whatever the agent's origin, since creation and deployment are common origins of ownership but do not define it and ownership can be transferred. In the autonomous form someone still answers, but through norms and law rather than through a mandate. What must never depend on the agent itself is the power to stop it, which stays outside the learned system, like the enforcement points of R6.

\textbf{The record that enables governance is itself sensitive.} Governed traces make behavior answerable by recording authority, evidence, decisions and outcomes, including restraint and sub-threshold gate evaluations (R12). The same record may also contain intimate evidence about what the agent inferred, when it chose not to intervene and why. Access to this record therefore requires the same discipline as access to perception itself. Consent, minimization, retention limits and scoped audit rights are necessary so that accountability to one party does not become surveillance by another.

\textbf{Societal scale.} Two broader asymmetries matter. First, well-governed symbiotic agents may become available mainly to those who can afford them, while others receive systems with weaker consent, calibration, containment and auditability. The governance structures described here should not become premium features. Second, when many proactive agents share the same channels, their individually justified interventions compete for the same human attention. Interruption cost then becomes a commons. Cross-agent restraint, coordination and limits on attention capture become part of the infrastructure that proactive systems owe their users.

\section{Conclusion}\label{sec:conclusion}

This work argues that intelligence becomes agency only when perception, intent, affective-conative inference, constraint and feedback organize behavior across time. The central question is not only whether a system can act. It is whether behavior should occur, when it should occur and in what mode. This is what describes the introduced activation problem. It marks the difference between proactive agency and task execution and it is also where restraint becomes a behavior rather than a failure to output.

The argument was developed across the taxonomy of autonomous, delegated and symbiotic agency. Autonomous agency organizes behavior around self-generated or fully transferred goals, under its own constraints and external norms. Delegated agency adds the mandate, making behavior answerable to a principal under bounded authority. Symbiotic agency is the most demanding form considered here. It is delegated agency under a standing mandate, with continuing coupling to the principal's situation, calibrated inference of the principal's condition and personalization within authorized bounds. In this form, the purpose of agency is not to replace the principal's judgment, but to augment their capabilities while remaining answerable to the conditions under which authority was transferred.

The main contribution is then the integrated structure, not any single element. Many elements have antecedents in prior work, but no prior approach joins behavior triggered by appraised significance with behavior answerable to a mandate, up to the continuing coupling required by the symbiotic form. This work provides that joint structure for autonomous, delegated and symbiotic agency, with particular emphasis on the symbiotic case. To make the structure precise, the present work defines behavioral episodes, the sensorium, allo-interoception, intent, appraisal, readiness, the activation gate, behavior selection, the mandate, authority containment, bounded adaptation and mandate-relative agency loss. It showed that restraint can be selected behavior, that governance and classification depend on what the record exposes and that the value of symbiotic coupling depends on the calibration of the agent's estimate of the principal's condition. Together, these results make proactivity something that can be specified, constrained, traced and evaluated, rather than a vague property of systems that act before being prompted.

The paper also turned the theory into instruments for comparison, measurement and design. Agency signatures record the structures of agency a system instantiates, establish whether it is an agent and classify the form of its agency. The evaluation framework measures how those structures operate, including activation quality, timing, constraint violations, mandate alignment, observability, calibration, delegation integrity and adaptation integrity. The implementation requirements and reference layer stack state what a realization must provide, whether it is built as an orchestrated runtime or pursued as a more integrated model. Together, these instruments make it possible to compare systems not only by task success, but by how their behavior remains warranted, constrained, traceable and answerable over time.

The next step is therefore not only to build more capable agents. It is to build agents whose proactivity remains authorized, calibrated, restrained and answerable. Three lines of work directly follow from the commitments made in this paper. The first is the Affective-Conative Agency Model, including its machine-learning formalization, training procedure and evaluation. The second is the realization of the benchmark, including the construction, validation and release of the seven scenario families in Section~\ref{sec:measurement}, so that the evaluation framework can be instantiated, criticized and extended independently. The third is the further formal development of the theory, carried out separately from this paper, such as stating and proving what the definitions imply for restraint, governance and calibration and establishing the conditions under which symbiotic agency remains robust in operation. Beyond these commitments, the open problems of Section~\ref{sec:open-problems} state what remains unsolved and consequently to be addressed in future work.

\clearpage
\appendix

\section{Detailed comparison of neighboring frameworks}\label{app:comparison}

Tables~\ref{tab:comparison} and~\ref{tab:comparison-forms} provide the detailed comparison behind Table~\ref{tab:proximity}. They compare the approaches element by element, using the same column groups: the baseline of agency and the autonomous, delegated and symbiotic form elements. The entries describe typical formulations in each approach, not everything that could be expressed after extension. For this reason, ``Limited'' and ``Partial'' mean that the element is not normally represented as a distinct first-class object. They correspond to the half-filled symbol in Table~\ref{tab:proximity}.

\begin{landscape}
\begin{table}[p]
\centering
\scriptsize
\caption{Detailed comparison behind Table~\ref{tab:proximity}, baseline elements: the typical formulation of each element in each approach.}
\label{tab:comparison}
\setlength{\tabcolsep}{4.5pt}
\begin{tabularx}{\linewidth}{>{\raggedright\arraybackslash}p{2.2cm}*{11}{>{\raggedright\arraybackslash}X}}
\toprule
Element & BDI / planning & POMDP / RL / world models & Active inference & \hspace{0pt}Computational appraisal & Situated / cybernetic & Principal--agent theory & LLM agent harnesses & Assistance games & Mixed-initiative (HCI) & Interface agents & Governed proactive agency (symbiotic form)\tabularnewline
\midrule
\multicolumn{12}{l}{\emph{Agenthood elements (Definition~\ref{def:agenthood})}}\tabularnewline
Standing intent & Intentions, first-class & Policy / objective; no explicit intent & Prior preferences & Goals, often & Setpoints / drives & Contract terms & Task-scoped & Shared objective, not the agent's & Session-scoped & Standing assistantship per domain & Carried intentions $I_{t}$\tabularnewline
Situated perception & Beliefs, symbolic & Belief state / world model & Generative model & Appraisal inputs & Continuous sensing & Not modeled & Tool observations & Observes human behavior & Attention and context sensing & Observes the user's application actions & Three-perspective sensorium\tabularnewline
Significance appraisal & Desires, priorities; limited & Utility / value & Expected free energy & Central & None; error only & Utility / incentives & No; prompt-driven & Utility-based & Utility of action vs. interruption & Confidence of predicted action & Affective-conative appraisal $C_{t}$\tabularnewline
Constrained selection & Limited & Action constraints / CMDP & No; preferences only & No & No & Contract terms & Sandbox / permissions & No & No & No & Permitted set $\mathcal{B}^{\phi}_{t}$, firewall\tabularnewline
Feedback and revision & Intention revision & Learning, central & Belief updating & Varies & Error correction, central & Incentive response & Memory-dependent & Learns reward from behavior & User models & Learning from the user, central & Bounded revision\tabularnewline
\multicolumn{12}{l}{\emph{Activation}}\tabularnewline
Gate: whether & Plan selection, limited & Implicit in policy & Implicit in policy selection & Action tendencies & Threshold & Not modeled & No; trigger-driven & Act vs. defer & Central & Do-it versus suggest thresholds & Activation gate $G_{\vartheta}$\tabularnewline
Timing: when & Limited & Possible via reward & Possible & Limited & Continuous & Not central & No; by trigger & Not modeled & Central & At confidence crossings & Gated timing\tabularnewline
Behavior mode with restraint & No & No-op possible & Possible & Possible & No & No & Ask vs. act, ad hoc & Deferral / off-switch & Deferral central & Do, suggest, or stay silent & Restraint first-class\tabularnewline
\multicolumn{12}{l}{\emph{Record}}\tabularnewline
Observable record & No & No & No & No & No & Monitoring, partial & Logs, partial & No & No & No & Traces $\mathcal{H}_{t}$; governed $\Gamma$\tabularnewline
\bottomrule
\end{tabularx}
\end{table}
\end{landscape}

\begin{landscape}
\begin{table}[p]
\centering
\scriptsize
\caption{Detailed comparison behind Table~\ref{tab:proximity}, elements of the three forms: the typical formulation of each element in each approach.}
\label{tab:comparison-forms}
\setlength{\tabcolsep}{4.5pt}
\begin{tabularx}{\linewidth}{>{\raggedright\arraybackslash}p{2.2cm}*{11}{>{\raggedright\arraybackslash}X}}
\toprule
Element & BDI / planning & POMDP / RL / world models & Active inference & \hspace{0pt}Computational appraisal & Situated / cybernetic & Principal--agent theory & LLM agent harnesses & Assistance games & Mixed-initiative (HCI) & Interface agents & Governed proactive agency (symbiotic form)\tabularnewline
\midrule
\multicolumn{12}{l}{\emph{Autonomous}}\tabularnewline
Own goals & Yes & Yes (reward) & Yes (preferences) & Often & Yes (setpoints) & Yes (self-interested agent) & Partly (long-running agents) & No; the human's & No & No; the user's tasks & No; mandate\tabularnewline
Self-governed constraint & Yes & Partial & Partial & Partial & Yes & No & No & No & No & No & No\tabularnewline
Self-directed adaptation & Partial & Yes, unbounded & Yes & Partial & Partial & No & No & Partial & No & Partial & No\tabularnewline
\midrule
\multicolumn{12}{l}{\emph{Delegated}}\tabularnewline
Explicit mandate & No & No & No & No & No & Explicit delegation & Prompt / task spec & Implicit objective & No & No & Explicit $J_{t}$\tabularnewline
Authority containment & No & Action constraints, partial & No & No & No & Contractual & Sandbox / permissions & Limited & No & Autonomy thresholds, partial & Explicit boundary\tabularnewline
Governed traces & No & No & No & No & No & Monitoring, partial & Logs, partial & No & No & No & $\Gamma$, explicit\tabularnewline
Consent-gated perception & No & No & No & No & No & No & \hspace{0pt}Permissions, partial & No & No & No & Policy-gated sensorium\tabularnewline
Bounded adaptation & Limited & Model-dependent & Model-dependent & Varies & Limited & Contract-dependent & Memory-dependent & Varies & No & No & Explicitly bounded $\mathcal{R}^{\phi}_{t}$\tabularnewline
\midrule
\multicolumn{12}{l}{\emph{Symbiotic}}\tabularnewline
Standing mandate & No & No & No & No & No & Contract duration, partial & No & No & No & No & Yes\tabularnewline
Principal-state inference & No & Belief over hidden state, generic & Belief-based, partial & Sometimes & No & Information asymmetry, partial & No & Preferences only & Attention and workload models & Habits and preferences, partial & Calibrated $U_{t}$\tabularnewline
Continuing coupling & No & No & Partial & Partial & Partial; the loop & Partial; the relationship & No & Partial & Partial & Partial; within applications & Yes\tabularnewline
Personalization & No & No & No & No & No & No & Memory, partial & Learned preferences, partial & User models, partial & Learned user model, central & Bounded $\rho^{user}_{t}$\tabularnewline
\bottomrule
\end{tabularx}
\end{table}
\end{landscape}

\printbibliography

\end{document}